\documentclass[12pt]{report}
\usepackage[letterpaper,top=1in,bottom=1in,right=1in,left=1in]{geometry}

\usepackage{graphicx}
\usepackage{amsmath}
\usepackage{cite}
\usepackage{siunitx}
\usepackage[hidelinks]{hyperref}
\usepackage{enumerate}
\usepackage{listings}
\usepackage{xcolor}
\usepackage{setspace}

\usepackage[nottoc]{tocbibind}

\usepackage{titlesec}
\titleformat{\chapter}[display]{\normalfont\bfseries}{\centering CHAPTER \thechapter}{0pt}{\centering}
\titleformat{\section}[block]{\normalfont\bfseries}{\thesection}{12pt}{}
\titleformat{\subsection}[block]{\normalfont}{\thesubsection}{12pt}{}

\usepackage{natbib} % has a nice set of citation styles and commands

\usepackage{mathtools} % amsmath with fixes and additions
\usepackage{booktabs} % commands to create good-looking tables
\usepackage{algorithm}
\usepackage{algorithmic}
\usepackage{amsmath}
\usepackage{amsthm}
\usepackage{amssymb}
\usepackage[capitalize,noabbrev]{cleveref}
\usepackage{nicefrac}
\usepackage{booktabs}
\usepackage{tikz}
\usetikzlibrary{arrows,automata}

\theoremstyle{definition}
\newtheorem{definition}{Definition}[section]
\theoremstyle{plain}
\newtheorem{theorem}{Theorem}[section]
\newtheorem{lemma}[theorem]{Lemma}

\hypersetup{colorlinks=true,citecolor=darkgray,linkcolor=black,urlcolor=black}
\crefname{equation}{Eq.}{Eqs.}

\theoremstyle{plain}
\theoremstyle{remark}

\newtheorem{corollary}[theorem]{Corollary}
\newtheorem{remark}[theorem]{Remark}

\newcommand{\piwsu}{\hat{\pi}}
\newcommand{\Real}{\mathbb{R}}
\newcommand{\opt}{^{\star}}
\newcommand{\PiR}{\Pi_{\mathrm{R}}}
\newcommand{\PiH}{\Pi_{\mathrm{H}}}
\renewcommand{\P}{\mathbb{P}}

\renewcommand{\cite}[1]{\citep{#1}}

\DeclareMathOperator*{\argmax}{argmax}
\DeclareMathOperator*{\argmin}{argmin}
\usetikzlibrary{arrows.meta,automata}
\usepackage{subcaption}

\usepackage{xcolor}
\usepackage{bm}
\usepackage{natbib}
\usepackage{amsmath,amssymb,amsthm,graphicx,url}
\usepackage{mathtools}
\usepackage{nicefrac}
\usepackage{booktabs}
\usepackage{tabularray}
\usepackage{microtype}
\usepackage{bbding}
\usepackage{eucal}
\usepackage[capitalize,noabbrev]{cleveref}

\renewcommand{\cite}{\citep}

\usepackage{tabularray}
\UseTblrLibrary{booktabs}

\newcommand{\E}{\mathbb{E}}

\newcommand{\states}{\mathcal{S}}
\newcommand{\actions}{\mathcal{A}}
\renewcommand{\exp}[1]{\operatorname{exp}\left( #1\right) }

\newcommand{\tr}{^{\top}}

\renewcommand{\exp}[1]{\operatorname{exp}\left(#1\right)}

\DeclareMathOperator{\ess}{ess}

\DeclareMathOperator{\ext}{ext}

\usepackage{tikz}
\usetikzlibrary{arrows.meta,automata}
\DeclareMathOperator{\cvaro}{CVaR}
\DeclareMathOperator{\evaro}{EVaR}
\DeclareMathOperator{\ermo}{ERM}

\newcommand{\PiHR}{\Pi_{\mathrm{HR}}}
\newcommand{\PiMR}{\Pi_{\mathrm{MR}}}
\newcommand{\PiMD}{\Pi_{\mathrm{MD}}}
\newcommand{\PiSR}{\Pi_{\mathrm{SR}}}
\newcommand{\PiSD}{\Pi_{\mathrm{SD}}}

\newcommand{\Nats}{\mathbb{N}}

\renewcommand{\exp}[1]{\operatorname{exp}\left( #1\right) }
\newcommand{\probs}[1]{\Delta_{#1}}  % probability space

\newcommand{\BellI}{T}
\newcommand{\BellIE}{L}

\renewcommand{\P}{\mathbb{P}}

\newcommand{\Ex}[1]{\mathbb{E}\left[ #1 \right]}

\newcommand{\erm}[2]{\ermo_{#1}\left[#2\right]}

\newcommand{\evar}[2]{\evaro_{#1} \left[#2\right]}
\newcommand{\ermp}[3]{\ermo_{#1}^{#2}\left[#3\right]}

\theoremstyle{plain}
\newtheorem{proposition}[theorem]{Proposition}
\theoremstyle{definition}
\newtheorem{assumption}[theorem]{Assumption}
\theoremstyle{remark}

\begin{document}

% front matter pages (go to corresponding files and replace dummies with your info)
% comment out pages that are not used
\doublespacing
\pagenumbering{roman}

\chapter*{Efficient Algorithms for Mitigating Uncertainty and Risk in
Reinforcement Learning}
\thispagestyle{empty}

\begin{center}

BY

\vfill

Xihong Su

%BS/MS, My University, Year
PhD in Computer Architecture, Harbin Institute of Technology, 

Harbin, China, 2012

BS in Computer Engineering, University of New Hampshire, 

Durham, USA, 2018

\vfill

DISSERTATION

\vfill

Submitted to the University of New Hampshire

in Partial Fulfillment of

the Requirements for the Degree of

\vfill

%Doctor of Philosophy/Master of *
Doctor of Philosophy

in

Computer Science

\vfill

September, 2025

\end{center}

\pagebreak
\null

\vfill

\begin{center}

ALL RIGHTS RESERVED

\copyright 2025

Xihong Su

\end{center}

\pagebreak

\begin{figure}[htp]
    \centering
    \includegraphics[width=0.99\linewidth]{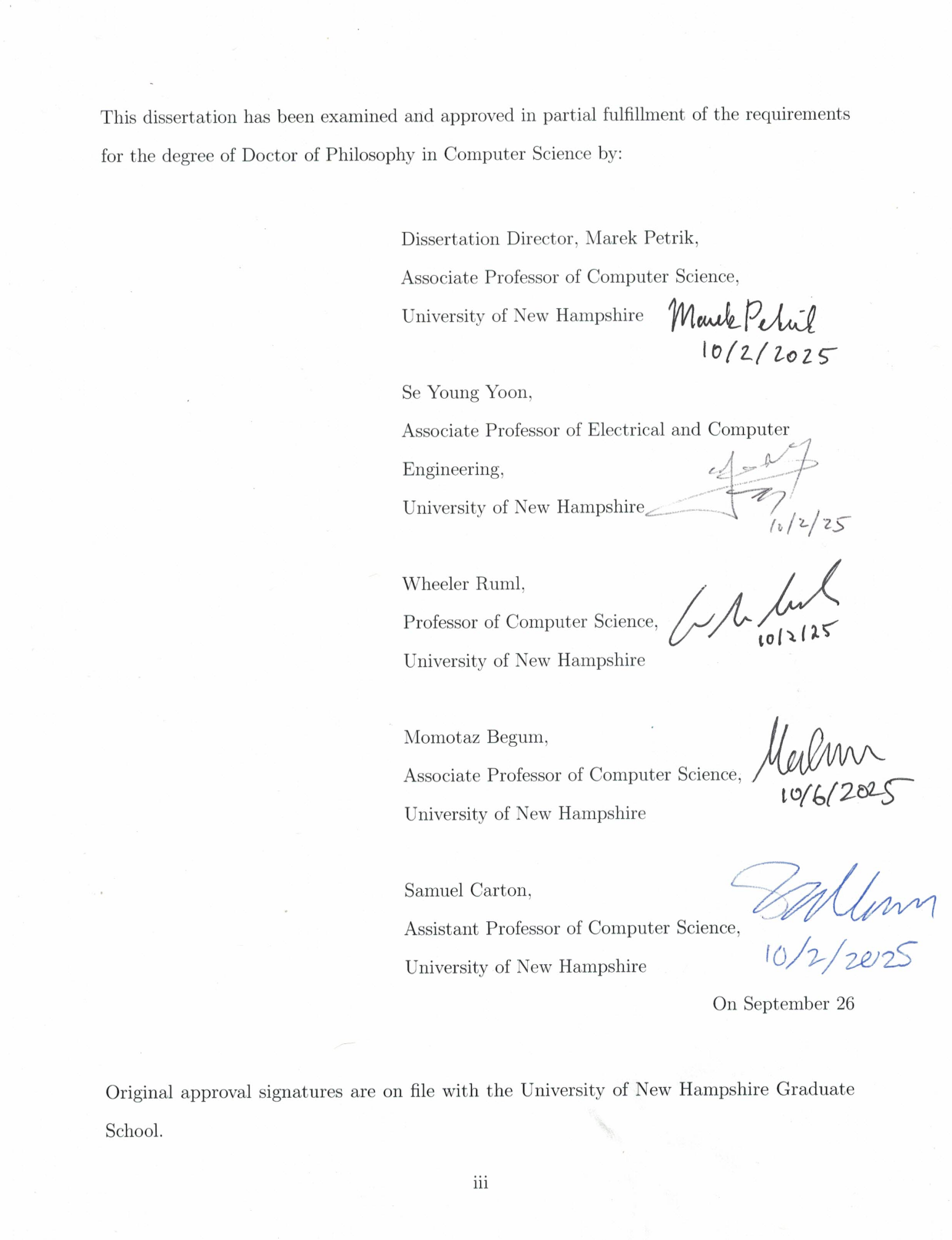}
   
\end{figure}

% \noindent This dissertation has been examined and approved in partial fulfillment of the requirements for the degree of Doctor of Philosophy in Computer Science by:

% \vspace{0.5in}

% \hfill
% \begin{minipage}{0.6\textwidth}

% Dissertation Director, Marek Petrik, \\ Associate Professor of Computer Science, \\University of New Hampshire \\

% Se Young Yoon, \\
% Associate Professor of Electrical and Computer \\ Engineering, \\ University of New Hampshire \\

% Wheeler Ruml, \\Professor of Computer Science, \\ University of New Hampshire\\

% Momotaz Begum, \\Associate Professor of Computer Science, \\ University of New Hampshire\\

% Samuel Carton,\\ Assistant Professor of Computer Science, \\ University of New Hampshire

% \hfill On September 26

% \end{minipage}

% \vspace{0.5in}

% \noindent Original approval signatures are on file with the University of New Hampshire Graduate School.

% \pagebreak
\chapter*{DEDICATION}
\addcontentsline{toc}{chapter}{DEDICATION}

\vfill

\begin{center}

To my father, Wanren, and my mother, Shurong, who taught me the value of hard work and perseverance.

To my husband, Tao, who supported me through the ups and downs of this dissertation. Your belief in me, your willingness to listen to my struggles, and your constant encouragement were invaluable. I couldn't have done it without you.

To my son, Joseph, and my daughter, Grace, for your endless love and for reminding me what truly matters.

\end{center}

\vfill

\pagebreak
\chapter*{ACKNOWLEDGEMENTS}
\addcontentsline{toc}{chapter}{ACKNOWLEDGEMENTS}

I would like to thank the people who made this truly joyful journey possible. 

First and foremost, I am deeply grateful to my advisor, Prof. Marek Petrik, for his invaluable guidance and support throughout the research projects since I began working with him in September 2021. His thoughtful advice
greatly influenced both my thinking and research and was essential in forming and refining the ideas described in this work. His insightful feedback and guidance on the papers improved my writing skills. Prof. Petrik is mathematically gifted, and collaborating with him on mathematical proofs deepens my understanding of challenging problems and enhances my research skills.  

This dissertation would not have been possible without the support of my committee. I would like to express my sincere thanks to Prof. Wheeler Ruml, Prof. Momotaz Begum, Prof. Samuel Carton, and Prof. Se Young Yoon for their time, expertise, and thoughtful feedback. Their guidance has been invaluable in helping me complete this dissertation. Coauthoring papers with Julien Grand-Cl\'ement, Jia Lin Hau, Gersi Doko, and Kishan Panaganti was a very stimulating learning experience. 
This work was supported by the generous National Science Foundation(NSF grants 2144601 and 221806).

I am very grateful to all group members of the UNH RLR lab for the insightful discussions and the sharing of diverse perspectives in reinforcement learning, which greatly enriched my understanding and research.
The precious memories of group hiking, skiing, and kayaking adventures are special and will always hold a place in my heart. I truly hope that the friendships formed here will last a lifetime.

Many thanks as well to the Department of Computer Science at UNH for providing a supportive and intellectually stimulating environment for my doctoral studies. By serving as a TA for some courses, I gained valuable experience in teaching and interacting with students. In addition, thank you to Prof. Elizabeth Varki and Prof. Momotaz Begum, for their crucial and essential support during the first phase of my PhD journey. This journey might not have continued without their support.

I would like to express my sincere thanks to my friends outside of my academic circle for their constant support. Sincere thanks are extended
to Sue Li for her unconditional kindness and invaluable support during my difficult time. I hope that one day, I will be able to become a person like her. Sincere thanks are extended to Yunzhi Qian for exploring world history with me during our leisure time. She opened up my eyes to the world around me. Thanks to Bill and Carol for their kindness in volunteering to help with my English. Their efforts are truly making a positive impact on me and have inspired me to ``pay it forward" in the future.

Finally, I am very grateful to Prof. Radim Bartos, Prof. Marek Petrik, and Stephen Wissow for encouraging me to get outdoors and engage in activities like running, biking, and swimming, which have helped me relax and feel refreshed. I don't have time or space to mention everyone who has touched me or helped me, but thank you all!

\pagebreak
\tableofcontents
\listoftables
\listoffigures
\chapter*{ABSTRACT}
\addcontentsline{toc}{chapter}{ABSTRACT}

\begin{center}

Efficient Algorithms for Mitigating Uncertainty and Risk in Reinforcement
Learning

\vspace{0.2in}

by

\vspace{0.2in}

Xihong Su

University of New Hampshire, September, 2025

\end{center}

\vspace{0.2in}

\noindent Reinforcement learning~(RL) studies the methodologies of improving the performance of sequential decision-making for autonomous agents and has achieved remarkable success in domains such as games, robotics, autonomous systems, finance, and healthcare. The Markov Decision Process~(MDP) is a mathematical framework for modeling agent-environment interactions in sequential decision-making problems. There are two primary sources of uncertainty in RL: epistemic uncertainty and aleatoric uncertainty. In RL, risk refers to the potential for an agent's policy to lead to undesirable outcomes, especially when the environment is uncertain.  In many domains, researchers seek policies that maximize the objectives while mitigating the uncertainty and risk in policymaking.

This dissertation makes three main contributions. First, the Multi-Model Markov Decision Process (MMDP) is a model that captures epistemic uncertainty, and solving MMDPs optimally is NP-hard. Previous work proposed a dynamic programming algorithm to compute the optimal policy approximately; however, it lacks an optimality guarantee for the computed policy. We identify a new connection between policy gradient and dynamic programming in MMDPs and propose the Coordinate Ascent Dynamic Programming (CADP) algorithm to compute a Markov policy that maximizes the discounted return averaged over the uncertain models. CADP adjusts model weights iteratively to guarantee monotone policy improvements to a local maximum.

Second, risk-averse policies are employed in RL to handle uncertainty and risk. In a discounted setting, the discount factor plays a crucial role in ensuring that the Bellman operator (used in dynamic programming ) satisfies the contraction property, and the value functions are always bounded. However, there are situations where the agent needs to optimize for long-term goals, and then ignoring discounting can be a suitable approach. The challenge is that the Bellman operator may not be a contraction if there is no discounting of future rewards. We study the risk-averse objectives in the total reward criterion(TRC), which means no discounting of future rewards.
We establish sufficient and necessary conditions for the exponential ERM Bellman operator to be a contraction and prove the existence of stationary deterministic optimal policies for ERM-TRC and EVaR-TRC. We also propose exponential value iteration, policy iteration, and linear programming algorithms for computing optimal stationary policies for ERM-TRC and EVaR-TRC.

Third, a major challenge in deriving practical RL algorithms is that the model of the environment is often unknown. However, traditional definitions of risk measures assume a known discounted or transient MDP model. We propose model-free Q-learning algorithms for computing policies with risk-averse objectives: ERM-TRC and EVaR-TRC. The challenge is that Q-learning ERM Bellman may not be a contraction. Instead, we use the monotonicity of Q-learning ERM Bellman operators to derive a rigorous proof that the ERM-TRC and the EVaR-TRC Q-learning algorithms converge to the optimal risk-averse value functions. The proposed Q-learning algorithms compute the optimal stationary policy for ERM-TRC and EVaR-TRC.

\pagebreak

% chapters
\pagenumbering{arabic}

%\input{chapters/basics}
% add more chapters here using \input{}

\chapter{Introduction}

Reinforcement learning~(RL) studies the methodologies for improving the performance of sequential decision-making for autonomous agents and has achieved remarkable success in domains such as games, robotics, autonomous systems, finance, and healthcare~\cite{Sutton2018, kushwaha2025survey,garcia2015comprehensive}. The Markov Decision Process~(MDP) is a mathematical framework for modeling agent-environment interactions in sequential decision-making under uncertainty~\cite{Sutton2018}, taking into account both the short-term outcomes of current decisions and opportunities for making decisions in the future~\cite{Kallenberg2021markov}. An MDP models states, actions, transitions, and associated rewards. MDPs can be classified based on how they handle rewards over time, specifically into discounted,
undiscounted, and average reward MDPs. The most common goal in an MDP is to find a policy that maximizes the expected total discounted reward over time~\cite{barto2021reinforcement}.

There are two main sources of uncertainty in RL: the epistemic (or model) uncertainty due to imperfect knowledge about the underlying MDP model and aleatoric uncertainty(inherent noise or randomness ) due to the inherent stochasticity of the MDP model~\cite{rigter2021risk}. As an illustrative example, consider a navigation system for an automated taxi service. The system attempts to minimize travel duration subject to uncertainty due to the traffic conditions, traffic lights, and pedestrian crossings. For each new day of operation, the traffic conditions are initially unknown. This corresponds to epistemic uncertainty over the MDP model. The epistemic uncertainty is reduced as the agent collects experience and learns which roads are busy or not busy. Traffic lights and pedestrian crossings cause stochasticity in the transition durations corresponding to aleatoric uncertainty. The aleatoric uncertainty remains even as the agent learns about the environment~\cite{rigter2021risk}.
This dissertation proposes algorithms to mitigate the effects of epistemic and aleatoric uncertainty.

\section{Mitigating the Effects of Epistemic Uncertainty}
\label{sec:epistemic-uncertainty}

In RL, epistemic uncertainty refers to the uncertainty caused by a lack of knowledge or incomplete information about the environment, including its transition dynamics and reward function. This type of uncertainty can, in principle, be reduced with more  information~\cite{hullermeier2021aleatoric}. For example, an RL agent exploring a complex environment, such as an autonomous robot in a new house, will have high epistemic uncertainty in areas it has not yet visited because it lacks sufficient data about those specific state-action spaces. This uncertainty can diminish as the agent gathers more data about the environment. When the transition dynamics and reward functions are known with certainty, standard dynamic programming methods can be used to find an optimal policy, or set of decisions, that maximizes the expected rewards over the planning horizon~\cite{puterman2014markov}. 

A significant obstacle in RL is the limited interaction between the agent and the environment, often due to factors such as data-collection costs or safety constraints. To overcome this, estimating an MDP's parameters must rely on observational data or multiple external sources, which inevitably leads to model errors. Model errors pose a significant challenge in many practical applications. Even minor errors can accumulate, and policies that perform well in the estimated model can fail catastrophically when deployed~\cite{steimle2018multi,behzadian2021optimizing,petrik2019beyond,nilim2005robust}. The general problem of multiple conflicting models in medical decision-making has been recognized~\cite{bertsimas2018optimal, steimle2017markov}. In a report from the Cancer Intervention and Surveillance Modeling Network regarding a comparative modeling effort for breast cancer, the authors note that ``the challenge for reporting multimodel results to policymakers is to keep it (nearly) as simple as reporting one-model results, but with the understanding that it is more informative and more credible. We have not yet met this challenge "~\cite{habbema2006chapter}. 

A stream of research on epistemic uncertainty in RL has taken the approach of multiple models~\cite{ahmed2017sampling,buchholz2019computation,merakli2020risk,Petrik2016,ahluwalia2021policy}. The most closely related work to this dissertation is that of Steimle et 
 al.~\cite {steimle2018multi}, which proposes a multi-model Markov decision process (MMDP) that generalizes an MDP to allow for multiple models of the transition probabilities and rewards, each defined on a common state space and action space. The exact model, including transition probabilities and rewards, is unknown, and instead, the agent possesses a \emph{distribution} over MDP models. The model weights (distribution) may be determined by expert judgment, or estimated from empirical distributions using Bayesian inference, or treated as uniform priors~\cite{steimle2021multi}. 
 The objective is to compute a policy that maximizes the average return across these uncertain models~\cite{buchholz2019computation,steimle2018multi}.
 The optimal policies in these models may need to be history-dependent. Prior work treats model weights as uniform priors. Once the distribution(model weights) over multiple MDP models is specified, it will remain the same~\cite{steimle2021multi,bertsimas2018optimal}.

 The concept of multiple models is also common in the stochastic programming literature; however, stochastic programming typically involves sequential decision-making under uncertainty across multiple stages. In stochastic programming, each model corresponds to a ``scenario" representing a discrete realization of the uncertain parameters within a problem, and the agent has varying levels of information about the model in different stages~\cite{birge2011introduction}.  For example, in a two-stage stochastic programming
 , the agent selects initial actions during the first stage before knowing which of the multiple scenarios will occur. After some uncertainty in the problem has been revealed, the agent then takes recourse actions in the second stage~\cite{steimle2021multi}. In contrast, in the
MMDP, the agent needs to specify all actions before the model parameters are realized.

 The MMDP work model also resembles the contextual Markov Decision Processes (CMDP)~\cite{Hallak2015a}, which also involves multiple MDPs. In CMDPs, the agent neither has any information about the parameters of any of the MDPs nor does it know which MDP it is interacting with. In contrast, MMDP assumes that the agent has the complete characterization of each of the MDPs, but it does not know which MDP it is interacting with~\cite{steimle2018multi}. The MMDP also differs from multi-task RL in that the MMDP is motivated by problems with a shorter horizon. In contrast, multi-task learning is suitable for problems where the planning is sufficiently long to observe convergence of estimates to their true parameters through a dynamic learning process~\cite{steimle2018multi}.

The MMDP model differs from robust dynamic programming, which represents parameter ambiguity through an ambiguity set formulation. The standard robust dynamic programming is a ``max-min" approach in which the agent seeks to find a policy that maximizes the worst-case performance when the transition probabilities are allowed to vary within an ambiguity set that satisfies rectangularity~\cite{iyengar2005robust,nilim2005robust}. Unfortunately, rectangular MDP formulations
tend to lead to overly conservative policies that achieve poor returns when model errors are minor. Robust MDPs define a set of plausible MDP parameters and optimize decisions with respect to the expected reward under the worst-case parameters. Much of the research in robust dynamic programming has focused on ways to mitigate the effects of parameter ambiguity while avoiding overly conservative policies by either using non-rectangular ambiguity sets or optimizing different objective functions~\cite{Delage2009, Xu2012,wiesemann2013robust,mannor2016robust,petrik2019beyond}.

\section{Mitigating the Effects of Aleatoric Uncertainty and Risk}
\label{sec:aleatoric-uncertainty}
Aleatoric uncertainty refers to the inherent randomness and noise present in data, systems, or experiments. In RL, aleatoric arises from the inherent stochasticity of the environment, such as unpredictable state transitions or reward generation processes. It is considered irreducible even with more data. Risk arises from the uncertainty associated with future events, and is often quantified by the potential for loss, and is inevitable since the consequences of actions are uncertain at the time a decision is made~\cite{shen2014risk}.

In standard RL, the common goal is to find a policy that maximizes the expected discounted cumulative reward over a given time horizon~\cite{li2020quantile,rigter2021risk}. However, this policy willingly accepts arbitrarily large risks to achieve the maximum increase in expected profit. This may be undesirable, especially when the law of large numbers is not applicable and expectation alone provides little insight into the actual dynamics~\cite{meggendorfer2022risk}, such as in autonomous navigation systems and treatment plans. An autonomous car is expected to drive safely always, and not ``on average"~\cite{bisi2021algorithms}. For a disease treatment plan, the drug responses to patients are stochastic due to variations in patients' physiology or genetic profiles. It is desirable to select a set of treatments with high effectiveness and minimize adverse effects~\cite{lam2022risk}.

Since risk is inevitable, an agent may be interested in optimizing a specific quantile of the cumulative reward, rather than its expectation. For example, a physician may want to determine the optimal drug regimen for a risk-averse patient to maximize the 0.10 quantile of the cumulative reward; this is the cumulative improvement in health that is expected to occur with at least 90$\%$ % 
probability for the patient. A cloud computing company, such as Amazon, might want its service to be optimized at the 0.01 quantile, meaning that the company strives to provide a service that satisfies its customers 99$\%$ of the time~\cite{li2020quantile}.

Risk-averse policies, which involve choosing a more certain but possibly lower expected total reward, are also considered economically rational. For example, a risk-averse investor might decide to put money into a bank account with a low but guaranteed interest rate rather than into a stock with possibly high expected returns but also a chance of high losses~\cite{shen2014risk}. Various approaches have been developed to model risk aversion. The finance literature has focused on developing functionals for directly measuring risk, such as coherent risk measures~\cite{shen2014risk}. The economics literature has explored how to express this attitude through a utility function that maps uncertain outcomes to a certain equivalent. To incorporate risk,  all existing literature applies a nonlinear transformation to either the experienced reward values or the transition probabilities~\cite{shen2014risk}. Finally, control literature has instead explored worst-case optimization or robust control, which can be seen as an extreme degree of risk aversion, in which one wants to protect oneself from the worst possible outcome~\cite{bisi2021algorithms}.

RL approaches are organized into two main classes: model-based and model-free. Model-based approaches aim to estimate either the transition model or the reward model, and then solve the MDP using exact or approximate solutions. On the other hand, model-free techniques work without the necessity of estimating a complete model, but they have as the object of their learning task the value function or the policy itself~\cite{bisi2021algorithms}.

\subsection{Model-based Risk-averse Policy}

To compute a risk-averse policy, a natural approach is to directly learn a risk-averse RL policy that minimizes the risk of having a small expected total reward~\cite{howard1972risk}. For quantifying such a risk, one can use monetary risk measures like entropic risk measure(ERM)~\cite{follmer2011entropic}, value-at-risk (VaR)~\cite{dempster2002risk}, conditional value-at-risk (CVaR)~\cite{rockafellar2000optimization}, and entropic value-at-risk (EVaR)~\cite{ahmadi2012entropic}. These risk measures capture the
total reward volatility and quantify the possibility of rare but catastrophic events.
ERM is the only law-invariant risk measure that satisfies the tower property, which is essential in constructing dynamic programs~\cite{hau2023entropic}. 
% ERM can also be viewed as a mean-variance criterion, where the risk is expressed as the variance of
% total reward~\cite{fei2021risk}. 
Alternatively, VaR, CVaR, and EVaR use quantile criteria and account for the return distribution~\cite{hong2020latent}. For example, the expectation may favor low probability outcomes yielding a high
return, whereas the $\alpha$-level-VaR guarantees the gain is greater than or equal to that value with high probability 1-$\alpha$~\cite{hau2024q}.

% But the VaR is not coherent~\cite{ahmadi2021risk}.

A coherent risk measure is a mathematical function used in financial economics to quantify risk that satisfies four properties: monotonicity, sub-additivity, positive homogeneity, and translation invariance~\cite{ahmadi2012entropic}. These properties ensure the risk measure behaves in a way that aligns with intuitive notions of risk and consistent financial principles. ERM and VaR are not coherent risk measures. CVaR is the best-known coherent risk measure and the most commonly used to model risk aversion in MDPs. CVaR is popular because of its intuitive interpretation as the expectation of the undesirable tail of the return random variable~\cite{greenberg2022efficient}.

Solving the CVaR objective under the discounted criterion requires a history-dependent optimal policy~\cite{hau2023entropic, bauerle2011markov,bauerle2022markov,pflug2016time,li2022quantile}. One can only formulate a dynamic program to compute the CVaR value function when the state space is augmented with an additional continuous random variable~\cite{bauerle2011markov,li2022quantile}, which significantly complicates the computation of the value function and the implementation of the policy. A popular remedy for the complexity of CVaR-MDPs is to use nested, also known as iterated or Markov, CVaR risk measure~\cite{bauerle2011markov,osogami2012iterated}. MDPs with a nested CVaR objective admit a value function that can be solved efficiently using dynamic programming. Unfortunately, the nested CVaR approximation of CVaR is inaccurate and difficult to interpret. 

EVaR is a coherent risk measure and has emerged as a promising alternative to the widely used CVaR in financial risk management. EVaR closely approximates CVaR and VaR at the same risk level~\cite{Ahmadi-Javid2012, ahmadi2017analytical}. Also, EVaR reduces to an optimization over ERM, preserving most of the computational advantages of ERM, which admits dynamic programming decompositions~\cite{patek1999stochastic, deFreitas2020,smith2023exponential, denardo1979optimal, hau2023entropic, Hau2023a}. Recent work derives a new dynamic programming formulation for ERM and uses it to compute the optimal Markov policies for EVaR discounted objectives~\cite{hau2023entropic}.

We consider the total reward criterion~(TRC) with both positive and negative rewards.  TRC also assumes an infinite horizon but does not discount future rewards. When the rewards are non-positive, the total reward criterion is equivalent to the stochastic shortest path problem, and when they are non-negative, it is equivalent to the stochastic longest path~\cite{Dann2023}. Prior work on TRC imposes constraints on the sign of the instantaneous rewards, such as all positive rewards~\cite{blackwell1967positive} or all negative rewards~\cite{bertsekas1991analysis, freire2016extreme, carpin2016risk, de2020risk, fei2021exponential, fei2021risk,  ahmadi2021risk, cohen2021minimax, meggendorfer2022risk}. The restriction on signs of rewards excludes certain shortest path problems with zero-length cycles~\cite{bertsekas2013stochastic}. 
Although nested CVaR under the total reward criterion may admit stationary policies that can be computed using dynamic programming~\cite{ ahmadi2021risk, meggendorfer2022risk, gavriel2012risk, deFreitas2020}, our work shows that the TRC with nested CVaR can be unbounded~\cite{su2024stationary} because the Bellman operator in the TRC with the risk-averse objectives may not be a contraction. This is significantly different from the discounted criterion, under which the discount factor ensures that the value functions are bounded.

For many robotic applications, TRC can be a better alternative to the more commonly used discounted, infinite-horizon MDPs or finite-horizon MDPs~\cite{carpin2016risk}. First, most robotic tasks have a finite duration, which is usually not known in advance, so one cannot use a finite horizon. Second, a discounted infinite-horizon reward is not suitable because the rewards received later are no less important~\cite{carpin2016risk}. Our work also indicates TRC may be preferable to the discounted criterion in a broad range of risk-averse reinforcement learning
domains.

%  Some works address the problem of optimizing the CVaR of the undiscounted total
% reward in a Bayes-Adaptive MDP (BAMDP)~\cite{sharma2019robust, rigter2021risk}. In model-based Bayesian RL, a belief distribution over the underlying MDP is maintained. The Bayesian RL problem can be reformulated as a planning problem in a Bayes-Adaptive MDP (BAMDP) with an augmented state space composed of the belief over the underlying MDP and the state in the underlying MDP~\cite{duff2002optimal}. These works reformulate the problem as a two-player stochastic game. This game is played between an agent computing optimal history-dependent policies with respect to a belief over models, and an adversary perturbs this belief. The problem can be solved efficiently only
% when the number of horizons is small, since the number of augmented states, i.e., possible posterior
% distributions, increases exponentially as the time horizon increases~\cite{wang2024bayesian}.

\subsection{Model-free Risk-averse Policy}

Traditional definitions of risk measures assume a known  discounted~\cite{hau2023entropic, Kastner2023, Marthe2023a, lam2022risk, li2022quantile, bauerle2022markov, Hau2023a}  or transient MDP model~\cite{su2025risk} and use machine learning to approximate a global value or policy function~\cite{moerland2023model}. Unfortunately, the dynamic model of the environment is often unknown. Model-free Q-learning serves as the foundation of many modern reinforcement learning algorithms. However, generalizing it to risk-averse objectives is not trivial because computing the risk of random return involves evaluating the full distribution of the return rather than just its expectation~\cite{hau2024q}.

There have been a few successful attempts~\cite{shen2014risk,hau2024q,wang2024bayesian,keramati2020being, lim2022distributional} with discounted rewards. Some recent works have presented model-free methods to find optimal CVaR policies~\cite{keramati2020being, lim2022distributional} or optimal VaR policies~\cite{hau2024q}, where the optimal policy may be either Markovian or history-dependent. The work in~\cite{wang2024bayesian}  formulates the infinite-horizon Bayesian risk MDP (BRMDP) with discounted rewards and finds the optimal stationary policy. The state space only contains the physical states. Then they develop a Q-learning algorithm to learn this optimal stationary policy. The work in ~\cite{shen2014risk} derives a risk-sensitive Q-learning algorithm, which applies a nonlinear utility function to the temporal difference~(TD) residual.

With the total reward criterion, recent work proposes algorithms to compute the optimal stationary
policy for the EVaR-TRC objective~\cite{su2025risk}. To develop a model-free risk-averse Q-learning algorithm, we need to investigate elicitable risk measures. Elicitability is important for regression analysis, model comparison, and model prediction~\cite{ince2025constructing,shen2014risk}. 
Elicitability refers to the fact that elicitable functionals admit a representation as the $\argmin$ of an expected loss function. Classical examples of elicitable risk measures are the expected value, which is the minimizer of the squared loss, and quantiles, which are the $\argmin$ of the expected pinball loss~\cite{ince2025constructing}. For monetary risk measures, VaR~\cite{ziegel2016coherence} is elicitable under some conditions, and CVaR is elicitable conditionally on the VaR~\cite{coache2023conditionally}. The ERM risk measure
can be formulated as a utility-based shortfall risk measure~\cite{Follmer2016stochastic}, which is elicitable~\cite{shen2014risk}. 

Following the line of work in~\cite{su2025risk}, we focus on ERM-TRC and EVaR-TRC objectives. The reason for studying the ERM is that the optimal policy can be shown to be stationary, which significantly simplifies the dynamic programming equations. The main technical challenge in developing the Q-learning risk-averse algorithms is that the Bellman operator in the TRC with the ERM objective may not be a contraction. Therefore, the risk-averse Q-learning algorithm will need to utilize other properties, such as the monotonicity of the Bellman operator, to identify the iteration that does not converge to a bounded value function.

\section{Overview of Dissertation}

This dissertation is organized as follows. \cref{chp:mmdp} describes a new DP algorithm for optimizing a risk-neutral objective in MMDPs. Unlike previous work that uses uniform and fixed model weights, we employ the policy gradient to adjust the model weights at each time step, and the proposed algorithm is guaranteed to reach the local maximum.  \cref{chp:model-based} studies the total reward criterion under risk-averse measures: ERM and EVaR. The main goal is to establish sufficient and necessary conditions for the existence of optimal stationary policies. We propose a linear programming algorithm to compute the optimal stationary policies for the ERM and EVaR. The experiments show that the TRC criterion may be preferable to a discounted criterion in a broad range of risk-averse RL domains. Based on the work in  \cref{chp:model-based}, we leverage the elicitability property of the ERM and propose Q-learning algorithms for solving ERM-TRC and EVaR-TRC objectives in  \cref{chp:model-free}. We prove the convergence of the proposed risk-averse Q-learning algorithms and show that they converge to the optimal value functions. Finally, \cref{chp:conclusion} concludes by discussing potential future work.

\chapter{Solving Multi-Model MDPs by Coordinate Ascent and Dynamic Programming} \label{chp:mmdp}

Multi-model Markov decision process~(MMDP) is a promising framework for computing policies that are robust to parameter uncertainty in MDPs. MMDPs aim to find a policy that maximizes the expected return over a \emph{distribution} of MDP models. Because MMDPs are NP-hard to solve, most methods resort to approximations. In this chapter, we derive the policy gradient of MMDPs and propose CADP, which combines a coordinate ascent method and a dynamic programming algorithm for solving MMDPs. The main innovation of CADP compared with earlier algorithms is to take the coordinate ascent perspective to adjust model weights iteratively to guarantee monotone policy improvements toward a local maximum. A theoretical analysis of CADP proves that it never performs worse than previous dynamic programming algorithms like WSU. Our numerical results indicate that CADP substantially outperforms existing methods on several benchmark problems.

This work appeared in the $39^{th} $ Conference on Uncertainty in Artificial Intelligence
(Xihong Su, Marek Petrik. Solving multi-model MDPs by coordinate ascent and dynamic programming. UAI 2023~\cite{su2023solving}.) 

\section{Introduction}

Markov Decision Processes~(MDPs) are commonly used to model sequential decision-making in uncertain environments, including reinforcement learning, inventory control, finance, healthcare, and medicine~\cite{puterman2014markov,boucherie2017markov,Sutton2018}. In most applications, like reinforcement learning~(RL), parameters of an MDP must be usually estimated from observational data, which inevitably leads to model errors. Model errors pose a significant challenge in many practical applications. Even small errors can accumulate, and policies that perform well in the estimated model can fail catastrophically when deployed~\cite{steimle2021multi,behzadian2021optimizing,petrik2019beyond,nilim2005robust}. Therefore, it is important to develop algorithms that can compute policies that are reliable even when the MDP parameters, such as transition probabilities and rewards, are not known exactly.

Our goal in this chapter is to solve finite-horizon \emph{multi-model MDPs}~(MMDPs), which were recently proposed as a viable model for computing reliable policies for sequential decision-making problems~\cite{buchholz2019computation,steimle2021multi,ahluwalia2021policy,Hallak2015a}. MMDPs assume that the exact model, including transition probabilities and rewards, is unknown, and instead, one possesses a \emph{distribution} over MDP models. Given the model distribution, the objective is to compute a \emph{Markov} (history-independent) policy that maximizes the return averaged over the uncertain models. MMDPs arise naturally in multiple contexts because they can be used to minimize the expected Bayes regret in \emph{offline reinforcement learning}~\cite{steimle2021multi}.

Since solving MMDPs optimally is NP-hard~\cite{steimle2021multi,buchholz2019computation}, most algorithms compute approximately optimal policies. One line of work has formulated the MMDP objective as a \emph{mixed integer linear program}~(MILP)~\cite{buchholz2019computation,lobo2020soft,steimle2021multi}. MILP formulations can solve small MMDPs optimally when given sufficient time, but they are hard to scale to large problems~\cite{ahluwalia2021policy}. Another line of work has sought to develop \emph{dynamic programming} algorithms for MMDPs~\cite{steimle2021multi,lobo2020soft,buchholz2019computation,steimle2018multi}. Dynamic programming formulations lack optimality guarantees, but they exhibit good empirical behavior and can serve as the basis for scalable MMDP algorithms that leverage reinforcement learning, value functions, or policy approximations.

The main contribution of this chapter is that we identify a new connection between policy gradient and dynamic programming in MMDPs and use it to introduce \emph{Coordinate Ascent Dynamic Programming} (CADP) algorithm. CADP improves over both dynamic programming and policy gradient algorithms for MMDPs. In fact, CADP closely resembles the prior state-of-the-art dynamic programming algorithm, Weight-Select-Update~(WSU)~\cite{steimle2021multi}, but uses adjustable model weights to improve its theoretical properties and empirical performance. Compared with generic policy gradient MMDP algorithms, CADP reduces the computational complexity, provides better theoretical guarantees and better empirical performance. 

Although we focus on tabular MMDPs in this work, algorithms that combine dynamic programming with policy gradient, akin to actor-critic algorithms, have an impressive track record in solving large and complex MDPs in reinforcement learning. Similarly to actor-critic algorithms, the ideas underlying CADP generalize readily to large problems; however, the empirical and theoretical analysis of such approaches is beyond the scope of this paper. It is also important to note that one cannot expect the policy gradient in MMDP to have the same properties as in ordinary MDPs. For example, recent work shows that policy gradient converges to the optimal policy in tabular MDPs~\cite{bhandari2021linear,agarwal2021theory}, but one cannot expect the same behavior in tabular MMDPs because this objective is NP-hard.

Finally, our CADP algorithm can serve as a foundation for new robust reinforcement learning algorithms. Most popular robust reinforcement learning algorithms rely on robust MDPs in some capacity. Robust MDPs maximize returns for the \emph{worst} plausible model error~(e.g.,~\cite{iyengar2005robust,ho2021a,goyal2022robust}) but are known to generally compute policies that are overly conservative. This is a widely recognized problem, and several recent frameworks attempt to mitigate it, such as \emph{percentile-criterion}~\cite{Delage2009,behzadian2021optimizing}, \emph{light-robustness}~\cite{Buchholz2019}, \emph{soft-robustness}~\cite{Derman2018,lobo2020soft}, and \emph{distributional robustness}~\cite{Xu2012}. \emph{Multi-model MDPs} can be seen as a special case of light-robustness and soft-robustness, and CADP can be used, as we show below, to improve some of the existing algorithms proposed for these general objectives.  

The remainder of the chapter is organized as follows. We discuss related work in \cref{sec:related-work}. The multi-model MDP framework is defined in \cref{sec:framework}. In \cref{sec:solution-methods}, we derive the policy gradient of MMDPs and present the CADP algorithm, which we then analyze theoretically in \cref{sec: theoretical-analysis}. Finally, \cref{sec:numer-exper} evaluates CADP empirically.

\section{Related Work} \label{sec:related-work}

Numerous research areas have considered formulations or goals closely related to the MMDP model and objectives. In this section, we briefly review the relationship of MMDPs with other models and objectives; given the breadth and scope of these connections, it is inevitable that we omit some notable but only tangentially relevant work.

\paragraph{Robust and soft-robust MDPs}

Robust optimization is an optimization methodology that aims to reduce solution sensitivity to model errors by optimizing for the worst-case model error~\cite{Ben-Tal2009}. Robust MDPs use the robust optimization principle to compute policies to MDPs with uncertain transition probabilities~\cite{nilim2005robust,iyengar2005robust,wiesemann2013robust}. However, the max-min approach to model uncertainty MDPs has been proposed several times before under various names~\cite{Satia1973,Givan2000}. Robust MDPs are tractable under an independence assumption popularly known as \emph{rectangularity}~\cite{iyengar2005robust,wiesemann2013robust,petrik2019beyond,goyal2022robust,mannor2016robust}. Unfortunately, rectangular MDP formulations tend to give rise to overly conservative policies that achieve poor returns when model errors are small. \emph{Soft-robust}, \emph{light-robust}, and \emph{distributionally-robust} objectives assume some underlying Bayesian probability distribution over uncertain models and use risk measures to balance the average and the worst returns better~\cite{Xu2012,lobo2020soft,Derman2018,Delage2009,Satia1973}. Unfortunately, virtually all of these formulations give rise to intractable optimization problems. 

\paragraph{Multi-model MDPs}

MMDP is a recent model that can be cast as a special case of soft-robust optimization~\cite{steimle2021multi,buchholz2019computation}. The MMDP formulation also assumes a Bayesian distribution over models and seeks to compute a policy that maximizes the average return across these uncertain models~\cite{buchholz2019computation}. Even though optimal policies in these models may need to be history-dependent, the goal is to compute \emph{Markov policies}. Markov policies depend only on the current state and time step and can be more practical because they are easier to understand, analyze, and implement~\cite{Petrik2016}. Existing MMDP algorithms either formulate and solve the problem as a mixed integer linear program~\cite{buchholz2019computation,steimle2021decomposition,ahluwalia2021policy}, or solve it approximately using a dynamic programming method, like WSU~\cite{steimle2021multi}.

\paragraph{POMDPs}

One can formulate an MMDP model as a \emph{partially observable MDP}~(POMDP) in which the hidden states represent the uncertain model-state pairs ~\cite{kaelbling1998planning,steimle2021multi,buchholz2019computation}. Most POMDP algorithms compute history-dependent policies~\cite{kochenderfer2022algorithms}, and therefore, are not suitable for computing Markov policies for MMDPs~\cite{steimle2021multi}. On the other hand, algorithms for computing finite-state controllers in POMDPs~\cite{Vlassis2012} or implementable policies~\cite{Petrik2016,ferrer2020solving} compute stationary policies. Stationary policies are inappropriate for the finite-horizon objectives that we target.

\paragraph{Bayesian Multi-armed Bandits}

MMDPs are also related to Bayesian exploration and multi-armed bandits. Similarly to MMDPs, \emph{Bayesian exploration} seeks to minimize the Bayesian regret, which is computed as the average regret over the unknown MDP model~\cite{lattimore2020bandit}. Most research in this area has focused on the bandit setting, which corresponds to an MDP with a single state. \emph{MixTS} is a recent algorithm that generalizes Thompson sampling to the full MDP case~\cite{hong2022thompson}. MixTS achieves a sublinear regret bound but computes policies that are history dependent and, therefore, are not Markov. We include MixTS in our empirical comparison and show that Markov policies cannot achieve sublinear Bayes regret. 

\paragraph{Policy Gradient}

As discussed in the introduction, CADP combines policy gradient methods with dynamic programming. Policy gradient methods are widespread in reinforcement learning and take a first-order optimization to policy improvement. Many policy gradient methods are known to be adaptations of general unconstrained or constrained first-order optimization algorithms---such as Frank-Wolfe, projected gradient descent, mirror descent, and natural gradient descent---to the return maximization problem~\cite{bhandari2021linear}. CADP builds on these methods but uses dynamic programming to perform more efficient gradient updates reusing much more prior information than the generic techniques.

\section{Framework: Multi-model MDPs}\label{sec:framework}

In this section, we formally describe the MMDP framework and show how it arises naturally in Bayesian regret minimization. We also summarize WSU, a state-of-the-art dynamic programming algorithm, to illustrate the connections between CADP and prior dynamic programming algorithms. 

\paragraph{MMDPs}
A \emph{finite-horizon MMDP} comprises the horizon $\mathcal{T}$, states $\mathcal{S}$, actions $\mathcal{A}$, models $\mathcal{M}$, transition function $p$, rewards $r$, initial distribution $\mu$, model distribution $\lambda$~\cite{steimle2021multi}. The symbol $\mathcal{T} = \{1, \dots , T\}$ is the set of decision epochs, $\mathcal{S} = \{1, \dots , S\}$ is the set of states, $\mathcal{A} = \{ 1, \dots , A \}$ is the set of actions, and $\mathcal{M} = \{ 1, \dots , M \}$ is the set of possible  models. The function $p^m\colon \mathcal{S} \times \mathcal{A} \to \Delta^{\mathcal{S}},\, m \in \mathcal{M}$ is the transition probability function, which assigns a distribution from the $S$-dimensional probability simplex $\Delta^{S}$ to each combination of a state, an action, and a model $m$ from the finite set of models $\mathcal{M}$. The functions $r^m_t\colon \mathcal{S} \times \mathcal{A} \to \Real, m\in \mathcal{M}, t\in  \mathcal{T}$ represent the reward functions. $ \mu $ is the initial distribution over states. Finally, $\lambda = (\lambda_{1}, \dots, \lambda_{M} )$ represents the set of initial model probabilities (or weights) with $\lambda_{m} \in (0,1)$ and $ \sum_{m \in \mathcal{M}} \lambda_{m} = 1$.

Note that the definition of an MMDP does not include a discount factor. However, one could easily adapt the framework to incorporate the discount factor $\gamma \in [0,1]$. It is sufficient to define a new time-dependent reward function $\hat{r}^{m}_{t} = \gamma^{t-1} r^{m}_{t}$ and solve the MMDP with this new reward function. 

Before describing the basic concepts necessary to solve MMDPs, we briefly discuss how one may construct the models $\mathcal{M}$ and their weights $\lambda_m$ in a practical application. The model weights $\lambda_{m} \in (0,1)$  may be determined by expert judgment, estimated from empirical distributions using Bayesian inference, or treated as uniform priors~\cite{steimle2021multi}. Prior work assumes that the decision maker has accurate estimates of model weights $\lambda_{m}$ or treats model weights as uniform priors~\cite{steimle2021multi,bertsimas2018optimal}. Once $\lambda_{m}, m \in \mathcal{M}$ is specified, the value of $\lambda_{m}$ does not change.

The solution to an MMDP is a \emph{deterministic Markov policy} $\pi_t \colon \mathcal{S} \to  \mathcal{A} ,\, t\in \mathcal{T}$ from the set of deterministic Markov policies $\Pi$. A policy $\pi_t(s)$ prescribes which action to take in time step $t$ and state $s$. It is important to note that the action is non-stationary---it depends on $t$---but it is independent of the model and history. The policies for MMDPs, therefore, mirror the optimal policies in standard finite-horizon MDPs. To derive the policy gradient, we will also need randomized policies $\PiR$ defined as $\pi_t\colon \mathcal{S} \to \Delta^{\mathcal{A}}, t\in \mathcal{T}$.

The \emph{return} $\rho\colon \Pi \to \Real$ for each policy $\pi\in \Pi$ is defined as the \emph{mean} return  across the uncertain true models:\begin{equation}\label{eq:return}
  \rho(\pi) =
  % \sum_{m \in\mathcal{M}} \lambda_{m} \cdot \mathbb{E}^{\pi,p^{m},\mu} \left[ \sum_{t=1}^{T}  r_{t}^{m}(S_{t},A_{t})\right] ~.
\mathbb{E}^{\lambda} \left[  \mathbb{E}^{\pi,p^{\tilde{m}},\mu} \left[ \sum_{t=1}^{T}  r_{t}^{\tilde{m}}(\tilde{s}_{t},\tilde{a}_{t}) \mid \tilde{m} \right] \right]~.
\end{equation}
$\tilde{m}, \tilde{s_t}$ and $\tilde{a_t}$ are random variables. 

The decision maker seeks a policy that maximizes the return:
\begin{equation}\label{eq:objective-return}
  \rho\opt \; =\;  \max_{\pi \in \Pi} \rho(\pi)~.
\end{equation}
As with prior work on MMDPs, we restrict our attention to deterministic Markov policies because they are easier to compute,
analyze, and deploy. While history-dependent policies can, in principle, achieve better returns than Markov policies, our numerical results show that existing state-of-the-art algorithms compute history-dependent policies that are inferior to the Markov policies computed by CADP.

Next, we introduce some quantities that are needed to describe our algorithm. Because an MMDP for a fixed model $m \in \mathcal{M}$ is in ordinary MDP, we can define the value function $v_{t,m}^{\pi}\colon \mathcal{S}\to \Real$ for each $\pi\in \Pi$, $t\in \mathcal{T}$, and $s\in \mathcal{S}$ as \cite{puterman2014markov}
\begin{equation}\label{eq:v-t-m-s}
v_{t,m}^{\pi}(s) = \mathbb{E}\left[\sum_{t' =t}^{T}r_{t'}^m(\tilde{s}_{t'},\tilde{a}_{t'}) \mid \tilde{s}_t = s, \tilde{m} = m\right].
\end{equation}
The value function also satisfies the \emph{Bellman equation} 
\begin{equation}\label{eq:backup}
v_{t,m}^{\pi}(s) = \sum_{a\in \mathcal{A}}  \pi_t(s,a)\cdot q_{t,m}^{\pi}(s,a) \,,
\end{equation}  
where state-action value function $q^{\pi}_{t,m}\colon \mathcal{S}\times \mathcal{A} \to \Real$ is defined as
\begin{equation}\label{eq:value-0f-(state-action)}
q_{t,m}^\pi (s, a) = r_t^{m}(s,a)  + \sum_{s' \in \mathcal{S}} p_t^m(s' \mid  s,a) \cdot v_{t+1,m}^{\pi}(s').
\end{equation}
The optimal value function $v\opt_{t,m}\colon \mathcal{S} \to \Real$ is the value function of the optimal policy $\pi\opt_m \in \Pi$ and satisfies that
\[
 v\opt_{t,m}(s) = \max_{a\in \mathcal{A}} q\opt_{t,m}(s,a), \quad  s\in \mathcal{S},
\]
where $q\opt_{t,m} = q_{t,m}^{\pi\opt_m}$.

Unlike in an MDP, the value function does \emph{not} represent the value of being in a state because the model $m$ is unknown.

\paragraph{Bayesian regret minimization}
The multiple models in MMDPs can originate from various sources~\cite{steimle2021multi}. We now briefly describe how these models can arise in offline RL because this is the setting we focus on in the experimental evaluation. In offline RL, the decision maker needs to compute a policy $\pi \in \Pi$ using a logged dataset of state transitions $\mathcal{D} = (t, s_i, a_i, s_i')$. In Bayesian offline RL, the decision maker is equipped with a prior distribution $\kappa \in \Delta^{\mathcal{M}}$ over the (possibly infinite) set of models $\mathcal{M}$ and uses the data $\mathcal{D}$ to compute a posterior distribution $\lambda \in \Delta^{\mathcal{M}}$. The goal is then to find a policy $\hat{\pi}\in \Pi$ that minimizes the Bayes regret
\begin{equation}\label{eq:regret}
 \hat{\pi} \in \arg \min_{\pi \in \Pi}  
\mathbb{E}^{\lambda} \left[\rho^{\tilde{m}}(\pi\opt_{\tilde{m}}) - \rho^{\tilde{m}}(\pi) \right]\,,
\end{equation}
where $\rho^m\colon \Pi \to \Real$ for $m\in \mathcal{M}$ is the return for the model $m$ and each policy. Note that $\tilde{m}$ is the random variable that represents the model. A policy is optimal in ~\eqref{eq:regret} if and only if it is optimal in \eqref{eq:objective-return} because the expectation operator is linear. The minimum regret policy can then be approximated by an MMDP using a finite approximation of the posterior distribution $\lambda$.

\paragraph{Dynamic Programming Algorithms}

The simplest dynamic program algorithm for an MMDP is known as Mean Value Problem~(MVP)~\cite{steimle2021multi}. MVP first computes an average transition probability function $\bar{p}_t\colon  \mathcal{S} \times \mathcal{A} \to  \Delta^{\mathcal{S}}, t\in \mathcal{T}$ for each $s,s'\in \mathcal{S}$, $a\in \mathcal{A}$ as
\[
 \bar{p}_t(s' \mid s,a) = \sum_{m\in \mathcal{M}} \lambda_m \cdot p_t^m(s'\mid s,a)~.
\]
and the average reward function $\bar{r}_t \colon \mathcal{S} \times \mathcal{A} \to \Real, t\in \mathcal{T}$ as
\[
 \bar{r}_t( s,a) = \sum_{m\in \mathcal{M}} \lambda_m\cdot r_t^m(s,a)~.
\]
One can then compute a policy $\bar{\pi}\in \Pi$ by solving the MDP with the transition function $\bar{p}$ and a reward function $\bar{r}$ using standard algorithms~\cite{puterman2014markov}.

A more sophisticated dynamic programming algorithm, WSU, significantly improves over MVP~\cite{steimle2021multi}. WSU resembles value iteration and updates the policy $\piwsu_t\colon \mathcal{S} \to \mathcal{A}$ and state-action value function $v^{\piwsu}_{t,m}$ for each model $m\in \mathcal{M}$ backward in time. After initializing the value function $v_{t,m}^{\piwsu}$ to $0$ at time $t = T$ for each $m\in \mathcal{M}$, it computes $q_{t,m}^{\piwsu}$ from \cref{eq:value-0f-(state-action)} at time $t = T-1$. The policy $\piwsu_t$ at time $t = T-1$ is computed by solving the following optimization problem:
\begin{equation}\label{eq:wsu-optimal-action}
  \piwsu_t(s_t) \in \arg\max_{a \in \mathcal{A}} \sum_{m \in \mathcal{M}} \lambda_m \cdot q_{t,m}^{\piwsu}(s_t, a),  \quad \forall s_t\in \mathcal{S}.
\end{equation}
The optimization in \cref{eq:wsu-optimal-action} chooses an action that maximizes the weighted sum of the values of the individual models~\cite{steimle2021multi}. After computing $\piwsu_t$ for $t=T-1$, WSU repeats the procedure for $T-2, T-3, \dots, 1$.

It is essential to discuss the limitations of WSU that CADP improves on. At each time step $t\in T, T-1, \dots , 1$, WSU computes the policy $\piwsu_t$ that maximizes the sum of values of models weighted by the \emph{initial} weights $\lambda_1, \dots , \lambda_M$. But using the initial weights $\lambda_1, \dots , \lambda_M$ here is not necessarily the correct choice. Recall that each model $m$ has potentially different transition probabilities. Simply being a state $s_t$ reveals some information about which models are more likely. One should not use the prior distribution $\lambda_1, \dots , \lambda_M$, but instead the \emph{posterior} distribution conditional on being in state $s_t$. This is what CADP does, and we describe it in the next section. 

\section{CADP Algorithm}
\label{sec:solution-methods}

We now describe CADP, our new algorithm that combines coordinate ascent with dynamic programming to solve MMDPs. CADP differs from WSU in that it appropriately adjusts model weights in the dynamic program.

In the remainder of the section, we first describe adjustable model weights in \cref{sec:model-weights}. These weights are needed in deriving the MMDP policy gradient in \cref{sec:policy-grad-mmdps}. Finally, we describe the CADP algorithm and its relationship to coordinate ascent in \cref{sec:algorithm}.

\subsection{Model Weights} \label{sec:model-weights}

We now give the formal definition of model weights. Informally, a model weight $b_{t,m}^{\pi}(s)$ represents the \emph{joint} probability of $m$ being the true model and the state at time $t$ being $s$ when the agent follows a policy $\pi$. The value $b_{t,m}^{\pi}(s)$ is useful in expressing the gradient of $\rho(\pi)$.

\begin{definition} \label{def-belief-state}
An \emph{adjustable weight} for each model $m\in \mathcal{M}$, policy $\pi\in \Pi$, time step $t\in \mathcal{T}$ , and state $s \in \mathcal{S}$ is
\begin{equation} \label{eq:belief-state}
  b_{t,m}^{\pi}(s) \; =\;  \P\left[\tilde{m} = m, \tilde{s}_t= s\right]\,,
\end{equation}
where $S_0 \sim \mu$, $\tilde{m} \sim \lambda$, and $\tilde{s}_1, \dots , \tilde{s}_T$ are distributed according to $p^{\tilde{m}}$ of policy $\pi$.
\end{definition}

Although the model weight $b_{t,m}^{\pi}(s)$ resembles the belief state in the POMDP formulation of MMDPs, it is different from it in several crucial ways. First, the model weight represents the \emph{joint} probability of a model and a state rather than a conditional probability of the model given a state. Recall that in a POMDP formulation of an MMDP, the latent states are $\mathcal{M} \times \mathcal{S}$, and the observations are $\mathcal{S}$. Therefore, the POMDP belief state is a distribution over $\mathcal{M} \times \mathcal{S}$. Second, model weights are \emph{Markov} while belief states are history-dependent. This is important because we can use the Markov property to compute model weights efficiently.

Computing the model weights $b$ directly from \cref{def-belief-state} is time-consuming. Instead, we propose a simple linear-time algorithm. At the initial time step $t=1$, we have that
\begin{equation}\label{eq:belief-initial}
  b_{1,m}^{\pi}(s) =   \lambda_{m} \cdot \mu(s), \quad \forall m \in \mathcal{M}, s \in \mathcal{S}, \pi \in \Pi.
\end{equation}
The weights for any $t' = t + 1$ for any $t = 1, \dots, T-1$ and any $s'\in \mathcal{S}$ can than be computed as
\begin{equation}\label{eq:belief-update}
  b_{t',m}^{\pi}(s') = \sum_{s_{t},a \in \mathcal{S} \times \mathcal{A}}
  p^{m}(s' |  s_{t}, a)  \pi_{t}(s_{t}, a)  b_{t,m}^{\pi}(s_{t})\,.
\end{equation}
Intuitively, the update in \cref{eq:belief-update} computes the marginal probability of each state at $t+1$ given the probabilities at time $t$. Note that this update can be performed for each model $m$ independently because the model does not change during the execution of the policy. 

Note that the adjustable model weights $b_{t,m}^{\pi}(s')$ are Markov because we only consider Markov policies $\Pi$. As discussed in the introduction, we do not consider history-dependent policies because they can be much more difficult to implement, deploy, analyze, and compute.

\subsection{MMDP Policy Gradient} \label{sec:policy-grad-mmdps}

Equipped with the definition of model weights, we are now ready to state the gradient of the return with respect to the set of randomized policies.
\begin{theorem}\label{thm:policy gradient}
The gradient of $\rho$ defined in \cref{eq:return} for each $t \in \mathcal{T}$, $\hat{s}\in \mathcal{S}$, $\hat{a} \in \mathcal{A}$, and $\pi \in \PiR$ satisfies that
\begin{equation}\label{eq: gradient-deter}
  \frac{\partial \rho(\pi)}{\partial\pi_t(\hat{s},\hat{a})}\; =\;
  \sum_{m \in \mathcal{M}} b_{t,m}^{\pi}(\hat{s})\cdot q_{t,m}^{\pi}(\hat{s},\hat{a})\,,
\end{equation}
where $q$ and $b$ are defined in \cref{eq:value-0f-(state-action)} and \cref{eq:belief-update} respectively. 
\end{theorem}
\begin{proof}
     For any time step $\hat{t} \in \mathcal{T}$, we can express the return as 
\begin{align*}
 \rho(\pi) 
&= \mathbb{E}^{\lambda,\pi,p^{\Tilde{m}},\mu} \left[ \sum_{t=1}^{T}  r_{t}^{\Tilde{m}}(\Tilde{s}_{t},\Tilde{a}_{t}) \right] \\
  &= \mathbb{E}^{\lambda,\pi,p^{\Tilde{m}},\mu} \left[ \sum_{t=1}^{\hat{t}-1}  r_{t}^{\Tilde{m}}(\Tilde{s}_{t},\Tilde{a}_{t}) \right]  + \mathbb{E}^{\lambda,\pi,p^{\Tilde{m}},\mu} \left[ \sum_{t=\hat{t}}^{T}  r_{t}^{\Tilde{m}}(\Tilde{s}_{t},\Tilde{a}_{t}) \right] \\
  &\overset{\text{(a)}}{=}  C + \mathbb{E}^{\lambda,\pi,p^{\Tilde{m}},\mu}\left[ \mathbb{E} \left[ \sum_{t=\hat{t}}^{T}  r_{t}^{\Tilde{m}}(\Tilde{s}_{t},\Tilde{a}_{t}) \mid  \tilde{s}_{\hat{t}} , \tilde{m} \right] \right]\\
  &\overset{\text{(b)}}{=}  C + \sum_{m\in \mathcal{M},s_{\hat{t}}\in \mathcal{S},a_{\hat{t}}\in \mathcal{A}}   \mathbb{P}\left[\tilde{m} = m, \tilde{s}_{\hat{t}}= s_{\hat{t}}\right] \pi_{\hat{t}}(s_{\hat{t}}, a_{\hat{t}}) \cdot  \mathbb{E}^{\lambda,\pi,p^{\Tilde{m}},\mu}\left[ \sum_{t=\hat{t}}^{T}  r_{t}^{\Tilde{m}}(\Tilde{s}_{t},\Tilde{a}_{t}) \mid  \tilde{s}_{\hat{t}} = s_{\hat{t}}, \tilde{a}_{\hat{t}} = a_{\hat{t}},  \tilde{m} = m  \right]\\
  &\overset{\text{(c)}}{=}  C + \sum_{m\in \mathcal{M},s_{\hat{t}}\in \mathcal{S},a_{\hat{t}}\in \mathcal{A}}   b_{\hat{t},m}^{\pi}(s_{\hat{t}}) \cdot  \pi_{\hat{t}}(s_{\hat{t}}, a_{\hat{t}}) \cdot  q_{\hat{t},m}^{\pi}(s_{\hat{t}}, a_{\hat{t}})\,.
\end{align*}
Here, we use $C = \mathbb{E}^{\lambda,\pi,p^{\Tilde{m}},\mu} \left[ \sum_{t=1}^{\hat{t}-1}  r_{t}^{\Tilde{m}}(\Tilde{s}_{t},\Tilde{a}_{t}) \right]$ for brevity. The step (a) follows from the law of total expectation, the step (b) follows from the definition of conditional expectation, and the step (c) holds from the definitions of $b$ and $q$ in~\eqref{eq:v-t-m-s},~\eqref{eq:backup}, and~\eqref{eq:belief-state}.

Using the expression above, we can differentiate the return for each $s\in \mathcal{S}$ and $a\in \mathcal{A}$ as
\[
  \frac{\partial \rho(\pi)}{\partial \pi_{\hat{t}}(s,a)}
  \quad =\quad 
  b_{\hat{t},m}^{\pi}(s)  \cdot  q_{\hat{t},m}^{\pi}(s, a)\,,
\]
which uses the fact that $C$, $b_{\hat{t},m}^{\pi}$, and $q_{\hat{t},m}^{\pi}$  are constant with respect to $\pi_{\hat{t}}$. The desired result then holds by substituting $t$ for $\hat{t}$, $\hat{s}$ for $s$, and $\hat{a}$ for $a$.
\end{proof}

\subsection{Algorithm} \label{sec:algorithm}

To formalize the CADP algorithm, we take a coordinate ascent perspective to reformulate the objective function $\rho(\pi)$. In addition to establishing a connection between optimization and dynamic programming, this perspective is particularly useful in simplifying the theoretical analysis of CADP in \cref{sec: theoretical-analysis}.

The return function $\rho(\pi)$ can be seen as a multivariate function with the policy at each time step seen as a parameter:
 \[
  \rho(\pi) = \rho(\pi_1, \dots,\pi_t, \dots, \pi_T)
 \]
where $\pi_t = \displaystyle [\pi_t(s_1,a_1), \dots,\pi_t(s_S,a_A)]$ for each $t \in \mathcal{T}$ with $s_i \in \mathcal{S}$ and $ a_j \in \mathcal{A}$. 

The coordinate ascent (or descent) algorithm maximizes $\rho(\pi)$ by iteratively optimizing it along a subset of coordinates at a time~\cite{Bertsekas2016nonlinear}. The algorithm is useful when optimizing complex functions that simplify when a subset of the parameters is fixed. \Cref{thm:policy gradient} shows that the return $\rho$ function has exactly this property. In particular, while $\rho$ is non-linear and non-convex in general, the following result states that the function is linear for each specific subset of parameters. 
\begin{corollary} \label{cor:ret-linear}
For any policy $\bar{\pi} \in \Pi$ and $t\in \mathcal{T}$, the function $\pi_t \mapsto \rho(\bar{\pi}_1, \dots , \pi_t, \dots , \bar{\pi}_T)$ is \emph{linear}.  
\end{corollary}
\begin{proof}
The result follows immediately from~\eqref{eq: gradient-deter}, which shows that $\partial \rho /\partial\pi_t(s,a) = \sum_{m \in \mathcal{M}} b_{t,m}^{\pi}(s)\cdot q_{t,m}^{\pi}(s,a)$ which is constant in $\pi_t(s,a)$ for each $s\in \mathcal{S}$, $a\in \mathcal{A}$, and $t\in \mathcal{T}$. Therefore, we have that $\partial^2 \rho /\partial\pi_t(s,a)^2 = 0$ and the function $\pi_t \mapsto \rho(\bar{\pi}_1, \dots , \pi_t, \dots , \bar{\pi}_T)$ is linear by the multivariate Taylor's theorem.
\end{proof}

Ordinary coordinate ascent applied to $\rho(\pi)$ proceeds as follows. It starts with an initial policy $\pi^0 = (\pi_1^0, \dots, \pi_T^0)$. Then, at each iteration $n = 1, \dots $, it computes $\pi^{n}$ from $\pi^{n-1}$ by iteratively solving the following optimization problem for each $t\in \mathcal{T}$:
\begin{equation}\label{eq:coordinate-descent}
    \pi_t^n \in \argmax_{\hat{\pi}_t \in \Real^{\mathcal{S} \times\mathcal{A}}}\rho(\pi_1^{n-1}, \dots, \hat{\pi}_t, \dots, \pi_T^{n}) 
\end{equation}
From \cref{cor:ret-linear}, this is a linear optimization problem constrained to a simplex for each state individually. Therefore, using the standard optimality criteria over a simplex~(e.g,~Ex.~3.1.2 in \cite{Bertsekas2016nonlinear}) we have that the optimal solution in \eqref{eq:coordinate-descent} for each $s\in \mathcal{S}$ satisfies that
\begin{equation}\label{eq:swsu-optimal-action}
 \pi_t^n(s)  \in \argmax_{a \in \mathcal{A}} \sum_{m \in \mathcal{M}} b_{t,m}^{\pi^{n-1}}(s)  \cdot q_{t,m}^{\pi^n} (s,a) .
\end{equation}
$\pi^n$ can be solved by enumerating over the finite set of actions. This construction ensures that we get a sequence of policies $\pi^0, \pi^1,\pi^2, \dots$ with non-decreasing returns: $\rho(\pi^0) \leq \rho(\pi^1) \leq \rho(\pi^2) \leq \dots$. 

\begin{algorithm}[tb]
\caption{OptimizePolicy} 
\label{alg:optimize-policy}
\textbf{Input}: MMDPs,  Model weights $b^{\pi^{n-1}}$ \\
\textbf{Output}: $\pi^n = (\pi_1^n,\ldots,\pi_T^n)$
\begin{algorithmic}[1]
   \STATE Initialize $v_{T+1,m}^{\pi^n}(s_{T+1}) = 0, 
     \forall m \in \mathcal{M} $ \\
    \STATE Initialize $\pi^n \gets \pi^{n-1}$ \\
   \FOR{$t = T, T-1, \dots , 1$}
   %\COMMENT{Solve \eqref{eq:coordinate-descent} for $t$} \\
   \FOR{Every state $s_t \in \mathcal{S}$}
      \STATE Update $\pi^n_t(s_t)$ according to \cref{eq:swsu-optimal-action} with $b^{\pi^{n-1}}_t(s_t)$ \label{line:wsu-equal}
        \STATE Update $v_{t,m}^{\pi^n}(s_t)$ according to \cref{eq:backup} for each $m \in \mathcal{M}$  \\
    \ENDFOR  
   \ENDFOR
\STATE \textbf{return} $\pi^n$
\end{algorithmic}
\end{algorithm}

While the coordinate ascent scheme outlined above is simple and theoretically appealing, it is computationally inefficient. The computational inefficiency arises because computing the weights $b$ and value functions $q$ necessary in~\eqref{eq:swsu-optimal-action} requires updating the entire dynamic program. The coordinate ascent algorithms must perform this time-consuming update in every iteration. To mitigate this computational issue, CADP interleaves the dynamic program with the coordinate ascent steps so that we reduce the updates of $b$ and $v$ to a minimum.

Conceptually, CADP is composed of two main components. \Cref{alg:optimize-policy} is the inner component that uses dynamic programming to compute a policy ${\pi}^{n}$ for some given adjustable model weights $b^{\pi^{n-1}}$. This algorithm uses the value function of visiting a state from \cref{eq:swsu-optimal-action} to choose an action in each state that maximizes the expected value of the action for each $s_t\in \mathcal{S}$.

At any time step $t$, the maximization attempts to improve the action to take at time step $t$. A curious feature of the update in \cref{eq:swsu-optimal-action} is that it relies on two different policies, $\pi^{n-1}$ and $\pi^n$. This is because it assumes that the weights $b_{t'}$ are computed for $t' \le t$ using policy, $\pi^{n-1}$, which is the policy the decision maker follows up to time step $t$. The policy for $t > t'$ would have been updated using the dynamic program and, therefore, is denoted as $\pi^n$.

\begin{algorithm}[tb]
   \caption{CADP: Coordinate Ascent Dynamic Programming}
   \label{alg:swsu}
\textbf{Input}: MMDP, $\pi^0$\\
\textbf{Output}: Policy $\pi\in \Pi $
\begin{algorithmic}[1]
   \STATE $n \gets 0$ 
   \REPEAT
   \STATE $ n \gets n+1 $
   \STATE $ b^{\pi^{n-1}}\gets $ model weights from \cref{eq:belief-update} using $\pi^{n-1}$
   \STATE $\pi^n \gets $ OptimizePolicy (MMDP, $ b^{\pi^{n-1}}) $
   \UNTIL{$ \rho (\pi^n) = \rho (\pi^{n-1}) $}
\STATE \textbf{return} $\pi^n$
\end{algorithmic}
\end{algorithm}

The second component of CADP is described in \cref{alg:swsu}. This algorithm begins with an arbitrary policy $\pi^0$ and then alternates between updating the adjustable model weights and refining the policy using \cref{alg:optimize-policy}. The initial policy $\pi^0$ can be arbitrary and computed using an algorithm such as MVP, WSU, or a randomized policy.

A single iteration of CADP has the time complexity of $\mathcal{O}(T S^2 A M)$, similar to running value iteration for each one of the models. The number of iterations could be quite large. In the worst case, the algorithm may run an exponential number of iterations, visiting a significant fraction of all deterministic policies. We show, however, in the following section that the algorithm cannot loop and that each iteration either terminates or computes a better policy. In contrast, the complexity of a plain coordinate ascent iteration over all parameters would be $\mathcal{O}(T^2 S^3 A^2 M)$.

It is also interesting to contrast CADP with WSU. Recall that the limitation of WSU stems from the fact that \cref{eq:wsu-optimal-action} relies on the \emph{initial} model weights that do not depend on the current state and time. We propose to use instead \emph{adjustable} model weights, which represent the joint probability of the current state and the model at each time step $t$. The following sections show that using these adjustable weights enables CADP's favorable theoretical properties and improves empirical solution quality.

\section{Error Bounds}\label{sec: theoretical-analysis}
In this section, we analyze the theoretical properties of CADP. In particular, we show that CADP will never decrease the return of the current policy. As a result, CADP can never cycle and terminates in a finite time. We also contrast MMDPs with Bayesian multi-armed bandits and show that one cannot expect an algorithm that computes Markov policies to achieve sublinear regret. 

The following theorem shows that the overall return does not decrease when the local value function does not decrease. 
\begin{theorem}\label{thm:improvement}
Suppose that \cref{alg:swsu} generates a policy $\pi^n=(\pi_t^n)^T_{t=1}$ at an iteration $n$, then \( \rho(\pi^{n}) \geq \rho(\pi^{n-1}).
\)
\end{theorem}

\begin{proof}
    Assume some iteration $n$. The proof then follows directly from the construction of the policy $\pi^n$ from $\pi^{n-1}$. By the construction in \cref{eq:swsu-optimal-action}, we have that:
  \[
    \rho(\pi_1^{n-1}, \dots\pi_{t-1}^{n-1}, \pi_t^n, \pi_{t+1}^{n}\dots, \pi_T^{n})
    \; \ge\;  
    \rho(\pi_1^{n-1}, \dots\pi_{t-1}^{n-1}, \pi_t^{n-1}, \pi_{t+1}^{n}\dots, \pi_T^{n})\,.
  \]
Note that the optimal form of the policy in \cref{eq:swsu-optimal-action} follows immediately from the standard first-order optimality criteria over a simplex~(e.g,~Ex.~3.1.2 in ~\cite{Bertsekas2016nonlinear}) and the fact that the function optimized in \cref{eq:swsu-optimal-action} is linear (\cref{cor:ret-linear}). In particular, we have that
\[
  \pi_t^n \in \argmax_{\hat{\pi}_t \in \Real^{\mathcal{S} \times\mathcal{A}}}\rho(\pi_1^{n-1}, \dots, \hat{\pi}_t, \dots, \pi_T^{n})
\]
if and only if for each $s\in \mathcal{S}$ and $a\in \mathcal{A}$
\[
  \frac{\partial \rho(\pi_1^{n-1}, \dots, \pi^n_t, \dots, \pi_T^{n})}{\partial\pi_t(s,\pi_t^n(s))}
 \;  \ge  \; 
  \frac{\partial \rho(\pi_1^{n-1}, \dots, \pi^n_t, \dots, \pi_T^{n})}{\partial\pi_t(s,a)}\,.
\]
Intuitively, this means that the optimal policy $\pi^n_t$ must choose actions that have the \emph{maximum} gradient for each state. The optimization in \cref{eq:swsu-optimal-action} then follows by algebraic manipulation from \cref{thm:policy gradient}.

\end{proof}

\Cref{thm:improvement} implies that \cref{alg:swsu} must terminate in finite time. This is because the algorithm either terminates or generates policies with monotonically increasing returns. With a finite number of policies, the algorithm must eventually terminate.
\begin{corollary}
\cref{alg:swsu} terminates in a finite number of iterations. 
\end{corollary}

Given that the iterations of CADP only improve the return of the policy, one may ask why the algorithm may fail to find the optimal policy. The reason is that the algorithm makes local improvements to each state. Finding the globally optimal solution may require changing the policy in two or more states simultaneously. The policy update in each one of the states may not improve the return, but the simultaneous update does. This property is different from the situation in MDPs, where the best action at time $t$ is independent of the actions at times $t' < t$. 

It is also important to acknowledge the limitations of our analysis. One could ensure that iterations of CADP do not decrease the policy's return by accepting only improved policy changes. CADP does better than this. It finds the improving changes and converges to a type of local maximum. The computed policy is a local maximum in the sense that no single-state updates can improve its return. 

Given the connection between MMDPs and Bayesian bandits, one may ask whether it is possible to give regret bounds on the policy computed by CADP. The main difference between CADP and multi-armed bandit literature is that we aim to compute Markov policies, whereas algorithms such as Thompson sampling compute history-dependent policies. We show next that it is impossible to achieve guaranteed sublinear regret with Markov policies.

The regret of a policy $\pi$ is defined as the average performance loss with respect to the best possible policy:
\[
  R_T(\pi) = \max_{\bar\pi \in \PiH}\rho_T(\bar\pi) - \rho_T(\pi),
\]
where $\PiH$ is the set of all history-dependent randomized policies and $\rho_T$ is the return for the horizon of length $T$. 

The following theorem shows that it is impossible to achieve sublinear regret with Markov policies. 
\begin{theorem}\label{thm:no-sublinear-regret}
There exists an MMDP for which no Markov policy achieves sub-linear regret. That is, there exists no $\pi\in \Pi$, $c>0$, and $t'>0$ such that
\[
R_t(\pi) \le c\cdot t \quad \text{for all} \quad  t \ge t'\,.
\]
\end{theorem}
\begin{proof}
    Consider the MMDP illustrated in \cref{fig:simple-mmdp-1}. 

\begin{figure}
\centering
\begin{minipage}{.4\textwidth}
\begin{tikzpicture}[->,-Latex,>=Latex,font=\large,node distance=35mm,el/.style = {inner sep=2pt, align=left, sloped},
every label/.append style = {font=\tiny}]
\node[state] (2) {$2$};
\node[state, below of=2] (3) {$3$};
\node[state, below right of=2] (4) {$4$};
\node[state, below left of=2] (1) {$1$};
\draw  (2) edge[left] node[el,above]{$a=1,2$} (1)
  (1) edge[bend right, below] node[el,below,pos=0.7]{$a=1$} (2)
  (1) edge[bend left,below] node[el,above,pos=0.7]{$a=2$} (3)
  (3) edge[bend left,below] node[el,below]{$a=1,2$} (1);
\draw (4) edge[loop,above] node[el,above]{$a=1,2$} (4);
\end{tikzpicture}
\end{minipage}
\begin{minipage}{.4\textwidth}
\begin{tikzpicture}[->,-Latex,>=stealth,font=\large,node distance=35mm,el/.style = {inner sep=2pt, align=left, sloped},
every label/.append style = {font=\tiny}]
\node[state] (2) {$2$};
\node[state, below left of=2] (1) {$1$};
\node[state, below of=2] (4) {$4$};
\node[state, below right of=2] (3) {$3$};
\draw  (2) edge[left]node[el,above]{$a=1,2$} (1)
  (1) edge[bend right, above] node[el,below,pos=0.7]{$a=1$} (2)
  (1) edge[bend left, below] node[el,above,pos=0.7]{$a=2$} (4)
  (3) edge[loop,above] node[el,above]{$a=1,2$} (3)
  (4) edge[bend left, below] node[el,below]{$a=1,2$} (1);
\end{tikzpicture}
\end{minipage}
\caption{Left: model $m_{1}$, right: model $m_{2}$}
\label{fig:simple-mmdp-1}
\end{figure}
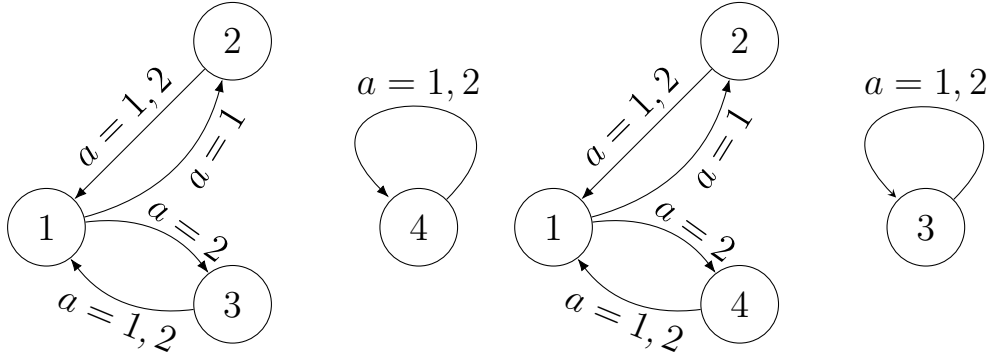

First, we describe the time steps, states, rewards, and actions for this MMDP. This MMDP has three time steps, four states $\mathcal{S} = \{1,2,3,4 \}$, two actions $\mathcal{A} = \{1,2\}$, and two models $\mathcal{M} = \{1,2\}$. The model weight for $m_{1}$ is $\lambda$, then the model weight for $m_{2}$ is $1-\lambda$.  State $1$ is the only initial state. In model $m_{1}$, the only non-zero reward 2 is received upon reaching state $2$. The agent takes action $1$, which leads to a transition to state $2$ with a probability of 1. The agent takes action $2$, which leads to a transition to state $3$ with probability 1.  The agent takes action $1$ or $2$ in state $2$, which leads to a transition to state $1$ with probability 1. The agent takes action $1$ or $2$ in state $3$, which leads to a transition to state $1$ with probability 1. The agent takes action $1$ or $2$ in state $4$, which leads to a transition to state $4$ with probability 1. 

In model $m_{2}$, the agent receives rewards 3 upon reaching state $4$ and receives rewards 2 upon reaching state $2$. The agent takes action $1$, which leads to a transition to state $2$ with probability 1. The agent takes action $2$, which leads to a transition to state $4$ with probability 1. The agent takes action $1$ or $2$ in state $2$, which leads to a transition to state $1$ with probability 1.  The agent takes action $1$ or $2$ in state $4$, which leads to a transition to state $1$ with probability 1.  The agent takes action $1$ or $2$ in state $3$, which leads to a transition to state $3$ with probability 1.

Now, let us analyze the regret of this MMDP. The optimal policy of the above example is a history-dependent policy. That is, to take action $2$ at time step $1$. At time step $2$, the agent takes action $1$ or $2$, which leads to a transaction back to state $1$. From time step $3$, if the agent is in model $m_{1}$, then take action $1$; if the agent is in model $m_{2}$, then take action $2$. 

Next, let us analyze the regret of a Markov policy for the MMDP. $S_{t}$ represents a state at time step $t$. State $1$ has two options: select action $1$ or select action $2$. If action $1$ is selected, this will give a regret value of 0 in model $1$ and a regret value of $1$ in model $2$. If action $2$ is selected, this will give a regret value of $2$ in model $1$ and a regret value of $0$ in model $2$. Therefore, at time step $1$, the total regret is 2$\lambda$ or 1$(1-\lambda$). At time step $2$, the agent takes action $1$ or $2$ in state $S_{2}$(1,3 or 4), which leads to a transition back to state $1$, and gets zero rewards and zero regrets. Then repeat the procedure. At time step $3$, the agent can take action $1$ or action $2$ in state $1$ again. For $T=3$, the trajectory of a Markov policy can be $(S_{1} =1, A_{1} = 1, S_{2},A_{2}, S_{3} =1,A_{3} =1)$, $(S_{1} =1, A_{1} = 1, S_{2},A_{2}, S_{3} =1,A_{3} =0)$, $(S_{1} =1, A_{1} = 0, S_{2},A_{2}, S_{3} =1,A_{3} =1)$, or $(S_{1} =1, A_{1} = 0, S_{2},A_{2}, S_{3} =1,A_{3} =0)$. The accumulated regret can be 2$\lambda$ +1$(1-\lambda$), 2$\lambda$ + 2$\lambda$, 1$(1-\lambda$) + 1$(1-\lambda$). That is, the regret is increased by 1$(1-\lambda$) or 2$\lambda$ for every two time steps. 
\[
R_t(\pi) \geq \frac{\min \{2\lambda, 1-\lambda\}}{2} \cdot t 
\]
Let $c = \frac{\min \{2\lambda, 1-\lambda\}}{2} $, $t' \geq 2$, then we always have 
\[
R_t(\pi) \geq c \cdot t \quad \text{for all} \quad  t \geq t'\
\]
No matter which Markovian policy the agent follows, the accumulated regret will be linear with respect to $t$. Therefore, for this MMDP, there exists no Markovian policy that achieves sub-linear regret.
\end{proof}

\section{Numerical Experiments}\label{sec:numer-exper}

\paragraph{Algorithms}In this section, we compare the expected return and runtime of CADP to several other algorithms designed for MMDPs as well as baseline policy gradient algorithms. We also compare CADP to related algorithms proposed for solving Bayesian multi-armed bandits and methods that reformulate MMDPs as POMDPs.

Our evaluation scenario is motivated by the application of MMDPs in Bayesian offline RL as described in \cref{sec:framework}. That is, we compute a posterior distribution over possible models $m\in \mathcal{M}$ using a dataset and a prior distribution. Then, we construct the MMDP by sampling \emph{training} MDP models from the posterior distribution. We evaluate the computed policy using a separate \emph{test} sample from the same posterior distribution. 

\paragraph{Domains}
The CSV files of all domains are available at \href{https://github.com/suxh2019/CADP}{https://github.com/suxh2019/CADP}. Let us describe each domain in detail. \emph{Riverswim (RS)}: This is a larger variation~\cite{behzadian2021optimizing} of an eponymous domain proposed to test exploration in MDPs~\cite{strehl2008analysis}. The MMDP consists of 20 states, 2 actions, 100 training models, and 700 test models. The training models are used to compute the policy, and test models are used to evaluate its performance. As in machine learning, this helps prevent overfitting. The discount factor is 0.9. 

\emph{Population (POP)}: The population domain was proposed for evaluating robust MDP algorithms~\cite{petrik2019beyond}. It represents a pest control problem inspired by the types of problems found in agriculture. The MMDP consists of 51 states, 5 actions, 1000 training models, and 1000 test models. The discount factor is 0.9. \emph{Population-small (POPS)} is a variation of the same domain that comprises a limited set of 100 training models and 100 test models.  

\emph{HIV}: Variations of the HIV management domains have been widely used in RL literature and proposed to evaluate MMDP algorithms~\cite{steimle2021multi}. The parameter values are adapted from Chen et al.~\cite{chen2017sensitivity}, and the rewards are based on Bala et al.~\cite{bala2006optimal}. In this case study, the objective is to find the sequence that maximizes the expected total net monetary benefit~(NMB). The MMDP consists of 4 states, 3 actions, 50 training models, and 50 test models. The discount factor is 0.9.

\emph{Inventory (INV)}: This model represents a basic inventory management model in which the decision makers must optimize stocking models at each time step~\cite{ho2021a}. The MMDP consists of 20 states, 11 actions, 100 training models, and 200 test models. The discount factor is 0.95.

Our CADP implementation initializes the policy $\pi^0$ to the WSU solution, sets the weights $\lambda_m, m\in \mathcal{M}$ to be uniform, and has no additional hyperparameters. We compare CADP with two prior MMDP algorithms: WSU, MVP~\cite{steimle2021multi} described in \cref{sec:framework}. We also compare CADP with two new gradient-based MMDP methods: mirror descent and natural gradient descent~\cite{bhandari2021linear}, which use the gradient derived in \cref{thm:policy gradient}.

We also compare CADP with applicable algorithms designed for models other than MMDPs. A natural algorithm for solving MMDPs is to reduce them to POMDPs. Therefore, we compare CADP with QMDP approximate solver~\cite{littman1995learning} and BasicPOMCP solver~\cite{silver2010monte} for POMDP planning. Recall that POMDP algorithms compute history-dependent policies, which are more complex but could, in principle, outperform Markov policies. Another method for solving MMDPs is to treat them as Bayesian exploration problems. We also compare CADP with MixTS~\cite{hong2022thompson}, which utilizes Thompson sampling to compute history-dependent randomized policies. The original MixTS algorithm assumes that one does \emph{not} observe the current state and only observes the rewards; we adapt it to our setting shown in \cref{alg:MixTS}. 
\begin{algorithm}
   \caption{Adapted MixTS}
   \label{alg:MixTS}
   \textbf{Input}: The prior of MDPs $P_0$ 
\begin{algorithmic}[1]
   \STATE Initialize $P_1 \longleftarrow P_0$\\
   \FOR{episodes t =1, $\cdots$ ,$\mathcal{N}$ }
   \STATE Sample $ M_{t}\sim P_{t}$ \\ 
   \STATE Compute $\pi_{t}$ = $\pi^{M_{t}}$ \\
   \FOR{timesteps h = 1, $\cdots$, H }
   \STATE Select $A_{h} \gets \pi_{t}(S_{h})$  \\
   \STATE Observe reward $Y_{h}$
  \STATE Update $P_{t+1}(m) 	\propto P_{t}(m)P(Y_{h} \mid A_{h};m), \forall m \in \mathcal{M}$
   \ENDFOR
    \ENDFOR
  \end{algorithmic}
\end{algorithm}
In \cref{alg:MixTS}, $P_0$ is the prior of MDPs and follows the uniform distribution. At the beginning of episode $t$, sample an MDP $M_t$ from the posterior $P_t$ and compute a policy $\pi_t$ that maximizes the value of $M_t$. Then at each time step $h$, take the action $A_h$ based on the policy $\pi_t$ and obtain the reward $Y_h$. For each MDP $m \in \mathcal{M}$, update its posterior based on the received rewards.
All algorithms were implemented in Julia 1.7, and the source code is available at \href{https://github.com/suxh2019/CADP}{https://github.com/suxh2019/CADP}.

\begin{figure}
  \centering
\includegraphics[width=0.7\linewidth]{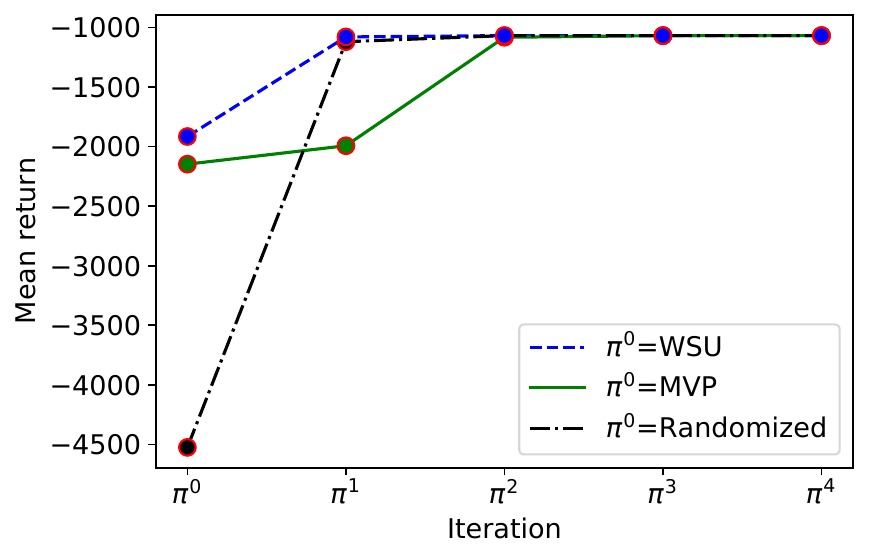}
\caption{Mean Returns of CADP with Different Initial Policies at Different Iterations}
\label{fig:returns}
\end{figure}

\paragraph{Return} First, \cref{fig:returns} compares the mean returns attained in the CADP computation, initialized with WSU, MVP, and a randomized policy change with different iterations on domain $POPS$. From the 3rd iteration on, the mean returns of CADP with the three initial policies are essentially identical. The only difference is that CADP initialized with WSU terminates one iteration earlier.  

\begin{table} 
\centering
 \caption{Mean return $\rho(\pi)$ on the Test Set of Policies $\pi$ Computed by Each Algorithm. HIV Values are in 1000s.} 
\begin{tabular}{lrrrrr}
\toprule
  \textbf{Algorithm}  & \textbf{RS} & \textbf{POP} & \textbf{POPS} & \textbf{INV} & \textbf{HIV}\\
  \midrule
  \textbf{CADP}    & \textbf{204}   &\textbf{-361} &\textbf{-1067} &323  &\textbf{42}\\
 
  WSU              &203    &-542   &-1915 & 323  &42 \\ 
  MVP              & 201   &-704   &-2147 & 323  &42 \\ 
 \midrule
  Mirror           & 181    &-1650  &-3676 &314  &\textbf{42} \\
  Gradient         & 203    &-542   &-1915 &323  &42 \\
 \midrule
  MixTS            & 167    &-1761 &-2857 & \textbf{327} & -1 \\
  QMDP             & 190    & --     & --     &  --   &40 \\
  POMCP            & 58     & --     & --     & --    &30 \\
  \midrule
  Oracle           & 210   &-168  &-882  &332  &53\\
  \bottomrule
\end{tabular}
\label{tab:returns}
\end{table}

\begin{table*}
  \centering
  \caption{Mean Returns $\rho(\pi)$ on the Test Set of Policies $\pi$ Computed by Each Algorithm.}\label{app:returns}
  \resizebox{\columnwidth}{!}{%
\begin{tabular}{lrrrrrrrrrr}
\toprule
  \textbf{Algorithm}  \bfseries  & \multicolumn{2}{c}{\textbf{RS}} \bfseries & \multicolumn{2}{c}{\textbf{POP}}  \bfseries & \multicolumn{2}{c}{\textbf{POPS}}\bfseries & \multicolumn{2}{c}{\textbf{INV}} & \multicolumn{2}{c}{\textbf{HIV}} \\
 \bfseries   &  T = 100  \bfseries & T =150  &  T = 100  \bfseries & T =150  &  T = 100  \bfseries & T =150  &  T = 100  \bfseries & T =150  &  T = 5  \bfseries & T =20\\
  \midrule
 \textbf{CADP}   &\textbf{207}  & \textbf{207}  & \textbf{-368} &\textbf{-368} &\textbf{-1082} &\textbf{-1082} &348 &\textbf{350} &\textbf{33348} &\textbf{42566}\\

   WSU               &206   &206  &-551     &-551   &-1934 &-1932 &347 &349   &\textbf{33348} &42564\\
 % MVP               &204   &204  &-717     &-717   &-2178 &-2179 &348 &\textbf{350}  &\textbf{33348} &42564 \\ 
  MVP               &204   &204  &-717     &-717   &-2178 &-2179 &348 &\textbf{350}  &\textbf{33348} &42564 \\ 
\midrule
 Mirror            &183   &183  &-1601    &-1600  &-3810 &-3800 &343 &345   &\textbf{33348} &\textbf{42566}\\
 Gradient          &206   &206  &-551     &-551   &-1934 &-1932 &347 &349   &\textbf{33348} &42564\\
\midrule
 MixTS             &172   &176  &-1961    &-1711  &-3042 &-3016 &\textbf{350} &\textbf{350}  &293 &-1026\\
QMDP               &201   &183  & -       &-      &-     &-     &-   &-     &30705  &39626\\ 
POMCP              &54      & 64    & -       &-      &-     &-     &-   &-     &25794  &30910\\
  \midrule
 Oracle     &213  & 213 &-172     &-172   &-894  &-894  &358 &360   &40159  &53856\\
  \bottomrule
\end{tabular}
}
\end{table*}
Second, \cref{tab:returns} summarizes the returns, or solution quality, of the algorithms on
HIV domain for horizon length $T=15$ and other domains for horizon length $T = 50$ evaluated on the \emph{test} set. The mean returns for time horizons $T=100$ and $T=150$ are in \cref{app:returns}. For \cref{tab:returns} and \cref{app:returns}, the algorithm ``Oracle'' describes an algorithm that knows that the true model and its returns are the means of each model's optimal values. Note that Oracle's return may be a loose upper bound on the best possible return. If the runtime of a method exceeds 900 minutes, the method is considered to have failed to find a solution and is marked with `--'. BasicPOMCP and QMDP approximate solvers fail to find solutions on domains POP, POPS, and INV.
\begin{table}
\centering
\caption{Standard deviation of returns of algorithms. HIV values are in 1000s.}
\label{standard-deviation}
\begin{tabular}{lrrrrr}
  \toprule
  \textbf{Algorithm}  & \textbf{RS} & \textbf{POP} & \textbf{POPS} & \textbf{INV} & \textbf{HIV}\\
 \midrule
\textbf{CADP}        &96    &\textbf{1081}  &\textbf{1986}  &\textbf{47}   & \textbf{11} \\ 
WSU                  &98    &1346    &3119  &49   &11   \\ 
MVP                  &89    &2018    &3577  &48   &11    \\ 
\midrule
Mirror              &\textbf{69}    &2150    &4494  &52   &11   \\ 
Gradient            &98    &1346    &3119  &49   &11   \\
 \midrule
MixTS              &223   &4398    &5454  & 58  &26   \\ 
QMDP                &213   &-       &-     &-    &64   \\
 POMCP              &72    &-       & -    & -   &55   \\ 
   \midrule
Oracle      &93   &1032    &1868  &47   &14   \\
  \bottomrule
\end{tabular}
\end{table}

\begin{table*}
  \centering
  \caption{Standard Deviation of Returns of Algorithms on Five Domains.}\label{app:standard-deviation}
   \resizebox{\columnwidth}{!}{%
\begin{tabular}{lrrrrrrrrrr}
\toprule
  \textbf{Algorithm}  \bfseries  & \multicolumn{2}{c}{\textbf{RS}} \bfseries & \multicolumn{2}{c}{\textbf{POP}}  \bfseries & \multicolumn{2}{c}{\textbf{POPS}}\bfseries & \multicolumn{2}{c}{\textbf{INV}} & \multicolumn{2}{c}{\textbf{HIV}} \\
 \bfseries   &  T = 100  \bfseries & T =150  &  T = 100  \bfseries & T =150  &  T = 100  \bfseries & T =150  &  T = 100  \bfseries & T =150  &  T = 5  \bfseries & T =20\\
\midrule
\textbf{CADP}     &98   &98   &\textbf{1095}   & \textbf{1095}     &\textbf{2007} &\textbf{2007}  &\textbf{51}  &\textbf{51} & 9342 &\textbf{11309} \\
 MVP                &90   &90   &2046   &2046  &3619   &3620  &52  &52 &\textbf{7729} &12234 \\
 WSU                &100  &100  &1364   &1364  &3147   &3146  &53  &53 &\textbf{7729}  & 12234\\
\midrule
 Mirror             &\textbf{70}  &\textbf{70}   &2081   &2081  &4534   &4530  &57  &58 &\textbf{7729} &12237 \\
 Gradient           &100  &100  &1364   &1364  &3147   &3146  &53  &53 &\textbf{7729}  & 12234\\
  \midrule
MixTS             &226  &231   &4436  &4187   &5507   &5542  &58  &58 &23689  &27792 \\
QMDP               &193  &204   & -     &-     &-      &-     &-   &-  &42987  &61596 \\
POMCP              &66   &118   &-      &-     &-      &-     &-   &-  &42208  &57772 \\
  \midrule
 Oracle    &95   &95   &1045   &1045  &1889   &1889  &51  &51 &9029  &14796 \\
  \bottomrule
\end{tabular}
}
\end{table*}

\Cref{standard-deviation} summarizes the standard deviation of returns of the algorithms on five domains for time horizon $T =50$. The results for time horizons $T=100$ and $T=150$ are in \cref{app:standard-deviation}.

\paragraph{Run-time}
\Cref{tab:runtimes} summarizes the runtime of several algorithms on the domains for horizon length $T = 50$. \cref{app:runtimes} summarizes the runtime of several algorithms on the domains for horizon length $T = 100$ and $T = 150$.
All algorithms were executed on a Ubuntu 20.04 with 3.0 GHz Intel processor and 32 GB of RAM. BasicPOMCP and QMDP approximate solvers fail to compute a policy on domains POP, POPS, and INV in a reasonable time. MVP runs fastest. MixTS runs slower than MVP but faster than other algorithms. The CADP method requires several iterations to obtain a policy that achieves the local maximum. As we expected, the time taken by CADP to solve these instances is several times that of WSU. The rate of convergence of the CADP depends on the length of the time horizon and the number of iterations performed. 

\begin{table}
    \centering
\caption{Run-times to compute a policy $\pi$ in minutes.}
  \begin{tabular}{lrrrrr}
  \toprule
 \textbf{Algorithm}   & \textbf{RS} & \textbf{POP} & \textbf{POPS} & \textbf{INV}  & \textbf{HIV}\\
  \midrule
 \textbf{CADP}  & 0.29   &69.66  &5.39  & 0.88  & 0.0053\\  
   WSU          & 0.08   & 33.65 &0.95  & 0.45  & 0.0016 \\ 
   MVP          & \textbf{0.05}   &\textbf{27.60}  &\textbf{0.36}  &\textbf{0.22}  &\textbf{0.0002}\\ 
  \midrule
   Mirror       &1.06    & 67.88  & 4.44 & 2.88   & 0.0111 \\ 
   Gradient     & 0.46   & 40.78  & 1.52 & 0.79   & 0.0042\\ 
  \midrule
   MixTS        & 0.07  &28.06   & 0.59  & 0.34   & 0.0018\\ 
   QMDP         & 712   & --     & --    & --     & 0.7100\\ 
   POMCP        & 68    & --     & --    & --     & 0.2066\\
  \bottomrule
\end{tabular}
\label{tab:runtimes}
\end{table}

\begin{table*}
  \centering
  \caption{Run-times of Algorithms on Five Domains in Minutes.}\label{app:runtimes}
   \resizebox{\columnwidth}{!}{%
\begin{tabular}{lrrrrrrrrrr}
\toprule
  \textbf{Algorithm}  \bfseries  & \multicolumn{2}{c}{\textbf{RS}} \bfseries & \multicolumn{2}{c}{\textbf{POP}}  \bfseries & \multicolumn{2}{c}{\textbf{POPS}}\bfseries & \multicolumn{2}{c}{\textbf{INV}} & \multicolumn{2}{c}{\textbf{HIV}} \\
 \bfseries   &  T = 100  \bfseries & T =150  &  T = 100  \bfseries & T =150  &  T = 100  \bfseries & T =150  &  T = 100  \bfseries & T =150  &  T = 50  \bfseries & T =100\\
\midrule
 MVP               &\textbf{0.05}  &\textbf{0.05}  &\textbf{27.68}   &\textbf{27.51}   &\textbf{0.36}   &\textbf{0.36}   &\textbf{0.22}  &\textbf{0.22}   &\textbf{0.0003}  &\textbf{0.0003}\\
 WSU               &0.12  &0.14  &40.02   &45.39   &1.53   &2.37   &0.67  &0.89   &0.0033  &0.0048\\
\textbf{CADP}              &0.52  &1.13  &124.39  &173.04  &12.12  &16.21  &1.53  &2.22   &0.0109  &0.0164\\
\midrule
 Mirror            &1.86  &3.11  &113.08  &158.06  &8.08   &11.90  &35.90 &53.6   &0.0221  &0.0330\\
 Gradient          &0.51  &0.74  &56.82   &69.32   &2.97   &4.31   &1.12  &1.44   &0.0083  &0.0123\\
  \midrule
MixTS             &0.09  &0.12  &32.08   &35.36   &0.80   &1.03   &0.47  &0.59   &0.0033  &0.0047\\
QMDP               &712   & 712  & -      &-       &-      &-      &-     & -     &0.7071  &0.7071\\
POMCP             &68    &68    & -      &-       &-      &-      &-     &-      &0.2066  &0.2066\\
  \bottomrule
\end{tabular}
}
\end{table*}

\paragraph{Discussion}
Our results show that CADP consistently achieves the best or near-best return in all domains and time horizons. This is remarkable because it only looks for Markov policies, whereas several other algorithms consider the richer space of history-dependent policies. The computational penalty that CADP incurs compared to just solving the average model, as done by MVP, is only a factor of 3-10. In comparison, state-of-the-art POMDP solvers were unable to solve most of the domains within a factor of 100 of CADP's runtime.

Let us take a closer look at the performance of the algorithms for a horizon $50$ on domain POPS. The runtime of WSU is $0.95$ minutes, but the return that WSU obtains is $-1915$. The runtime taken by CADP is $5.6$ times as much as WSU, but the return for CADP is -1067, which is significantly greater than what WSU achieves. MixTS is a sampling-based algorithm that is guaranteed to perform well over long horizons but has no guarantees for short horizons. However, this assumption may not hold in this domain, which could result in MixTS performing poorly. The mirror descent algorithm and CADP yield the same return on the domain HIV, but CADP outperforms the mirror descent algorithm significantly on other domains. Therefore, CADP outperforms these approaches with a slight runtime penalty. 

\section{Conclusions and Future Work}\label{sec:concl-future-work}
This paper proposes a new efficient algorithm, CADP, which combines a coordinate ascent method and dynamic programming to solve MMDPs. CADP incorporates adjustable weights into the MMDP and adjusts the model weights each iteration to optimize the deterministic Markov policy to the local maximum. Our experimental results and theoretical analysis show that CADP outperforms existing approximation algorithms on several benchmark problems. The only drawback of CADP is that it needs several iterations to obtain a converged policy and increases the computational complexity. In terms of future work, it would be worthwhile to scale up CADP to value function approximation and consider richer soft-robust objectives. It would also be worthwhile to design algorithms that incorporate limited memory into the policy to compute simple, history-dependent policies.

\chapter{Risk-averse Total-reward MDPs with ERM and EVaR}
\label{chp:model-based}

This chapter focuses on mitigating the effects of aleatoric uncertainty and risk due to the inherent stochasticity of the environment, such as an unpredictable state transition process. The proposed algorithms in this chapter belong to model-based RL approaches.

Optimizing risk-averse objectives in discounted MDPs is challenging because most models do not admit direct dynamic programming equations and require complex history-dependent policies. In this chapter, we show that the risk-averse {\em total reward criterion}, under the Entropic Risk Measure (ERM) and Entropic Value at Risk (EVaR) risk measures, can be optimized by a stationary policy, making it simple to analyze, interpret, and deploy. We propose exponential value iteration, policy iteration, and linear programming to compute optimal policies. Compared with prior work, our results only require the relatively mild condition of transient MDPs and allow for {\em both} positive and negative rewards. Our results indicate that the total reward criterion may be preferable to the discounted criterion in a broad range of risk-averse reinforcement learning domains. 

This work appeared in the $39^{th} $ Annual AAAI Conference on Artificial Intelligence (Xihong Su, Marek Petrik, Julien Grand-Cl\'ement. Risk-averse Total-reward MDPs with ERM and EVaR. AAAI2025~\cite{su2025risk,su2024stationary}), Seventeenth European Workshop on Reinforcement Learning (Xihong Su, Marek Petrik, Julien Grand-Cl\'ement. EVaR Optimization in MDPs with Total Reward Criterion. EWRL2024~\cite{suevar}), and ICML 2024 Workshop on Foundations of Reinforcement Learning and Control (Xihong Su, Marek Petrik, Julien Grand-Cl\'ement.Optimality of Stationary Policies in Risk-averse Total-reward MDPs with EVaR~\cite{su2024optimality}).

.

\section{Introduction}
\label{sec:intro} 

Risk-averse Markov decision processes~(MDP)~\cite{puterman205markov} that use \emph{monetary risk measures} as their objective have been gaining in popularity in recent years~\cite{Kastner2023, Marthe2023a, lam2022risk, li2022quantile, bauerle2022markov, hau2023entropic, Hau2023a,suevar,su2024optimality}. Risk-averse objectives, such as Value at Risk~(VaR), Conditional Value at Risk~(CVaR), Entropic Risk Measure~(ERM), or Entropic Value at Risk~(EVaR), penalize the variability of returns~\cite{Follmer2016stochastic}. As a result, these risk measures yield policies with stronger guarantees on the probability of catastrophic losses, which is important in domains like healthcare or finance. 

In this chapter, we target the \emph{total reward criterion} (TRC)~\cite{Kallenberg2021markov,puterman205markov} instead of the common discounted criterion. TRC also assumes an infinite horizon but does not discount future rewards. To control for infinite returns, we assume that the MDP is {\em transient}, i.e. that there is a positive probability that the process terminates after a finite number of steps, an assumption commonly used in the TRC literature~\cite{filar2012competitive}. We consider the TRC with both positive and negative rewards. When the rewards are non-positive, the TRC is equivalent to the \emph{stochastic shortest path} problem, and when they are non-negative, it is equivalent to the \emph{stochastic longest path}~\cite{Dann2023}.

Two reasons motivate our departure from discounted objectives in risk-averse MDPs. First, considering risk significantly affects discounted objectives. It is common to use discounted objectives because they admit optimal stationary policies and value functions that can be computed using dynamic programming. However, most risk-averse discount objectives, such as VaR, CVaR, or EVaR, require that optimal policies are \emph{history-dependent}~\cite{bauerle2011markov,Hau2023a,hau2023entropic} and do not admit standard dynamic programming optimality equations.

Second, TRC captures the concept of stochastic termination, a common phenomenon in reinforcement learning~\cite{Sutton2018}. In risk-neutral objectives, discounting can serve well to model the probability of termination because it guarantees the same optimal policies~\cite{puterman205markov,su2023solving}. However, as we show in this work, no such correspondence exists with risk-averse objectives, and the difference between them may be arbitrarily significant. Modeling stochastic termination using a discount factor in {\em risk-averse} objectives is inappropriate and leads to dramatically different optimal policies. 

As our main contribution, in this chapter, we show that the risk-averse TRC with ERM and EVaR risk measures admits optimal stationary policies and optimal value functions in transient MDPs. We also show that the optimal value function satisfies dynamic programming equations and can be computed using exponential value iteration, policy iteration, or linear programming algorithms. These algorithms are simple and closely resemble the algorithms for solving MDPs.

Our results indicate that EVaR is a particularly interesting risk measure in the context of reinforcement learning. ERM and the closely related exponential utility functions have been popular in sequential decision-making problems because they admit dynamic programming decompositions~\cite{patek1999stochastic, deFreitas2020,smith2023exponential, denardo1979optimal, hau2023entropic, Hau2023a}. Unfortunately, ERM is difficult to interpret, as it is scale-dependent and incomparable with popular risk measures such as VaR and CVaR. Because EVaR reduces to an optimization over ERM, it preserves most of the computational advantages of ERM, and since EVaR closely approximates CVaR and VaR at the same risk level, its value is also much easier to interpret. Finally, EVaR is also a coherent risk measure, unlike ERM~\cite{Ahmadi-Javid2012, ahmadi2017analytical}. 

\begin{table}
\centering
\begin{tabular}{lcccc} 
\toprule
& \multicolumn{2}{c}{Risk properties} & \multicolumn{2}{c}{Optimal policy} \\
\cmidrule(rl){2-3}  \cmidrule(rl){4-5} 
Risk measure & Coherent & Law inv. & Disc. & TRC \\  
\midrule
$\E$   & yes & yes & S & S\\
\midrule
EVaR   & yes & yes & M & \textcolor{blue}{S} \\
ERM    & no & yes & M & \textcolor{blue}{S} \\
NCVaR  & yes & no & S & S \\
VaR    & yes & yes & H & H \\
CVaR   & yes & yes & H & H \\
\bottomrule
\end{tabular}
\caption{Structure of optimal policies in
  risk-averse MDPs: ``S'', ``M'' and ``H'' refer to Stationary, Markov and History-dependent policies respectively.}
\label{fig:policy-comparison}
\end{table}
   
\Cref{fig:policy-comparison} puts our contribution in the context of other work on risk-averse MDP objectives. Optimal policies for VaR and CVaR are known to be history-dependent in the discounted objective~\cite{bauerle2011markov,Hau2023a} and must be history-dependent in TRC because TRC generalizes the finite-horizon objective. The TRC with Nested risk measures, such as Nested CVaR~(NCVaR), applies the risk measure in each level of the dynamic program independently and preserves most of the favorable computational properties of risk-neutral MDPs~\cite{ahmadi2021risk}. Unfortunately, nested risk measures are difficult to interpret; their value depends on the sequence in which the rewards are obtained in a complex and unpredictable way~\cite{Kupper2006} and may be unbounded even if MDPs are transient. 

While we are unaware of prior work on the TRC objective with ERM or EVaR risk-aversion {\em allowing both positive and negative rewards}, the ERM risk measure is closely related to exponential utility functions. Prior work on TRC with exponential utility functions also imposes constraints on the sign of the instantaneous rewards, such as all positive rewards~\cite{blackwell1967positive} or all negative rewards~\cite{bertsekas1991analysis, freire2016extreme, carpin2016risk, de2020risk, fei2021exponential, fei2021risk,  ahmadi2021risk, cohen2021minimax, meggendorfer2022risk}. Disallowing a mix of positive and negative rewards limits the modeling power of prior work because it requires that either all states are more desirable or all states are less desirable than the terminal state. Allowing rewards with mixed signs raises some technical challenges, which we address by employing a squeeze argument that leverages the transience of MDPs. 

\textbf{Notation.} We use a tilde to mark random variables, e.g. $\tilde{x}$. Bold lower-case letters represent vectors, and upper-case bold letters represent matrices. Sets are either calligraphic or upper-case Greek letters. The symbol $\mathbb{X}$ represents the space of real-valued random variables. When a function is defined over an index set, such as $z\colon \{1,2, \dots, N \} \to \Real$, we also treat it interchangeably as a vector $\bm{z}\in \Real^n$ such that $z_i = z(i), \forall i = 1, \dots , n$. Finally, $\Real, \Real_+, \Real_{++}$ denote real, non-negative real, and positive real numbers, respectively. $\bar{\Real} = \Real \cup \{-\infty, \infty\}$. Given a finite set $\mathcal{Y}$, the probability simplex is $\Delta_{\mathcal{Y}}:=\{x \in \Real_+^{\mathcal{Y}} \mid  \bm{1}^{\mathrm{T}}x=1\}$.

\section{Background on Risk-averse MDPs}
\label{sec:preliminaries}

The MDP below includes a sink state, which is different from the MDP defined in \cref{chp:mmdp}. Once an agent reaches the sink state, it will remain in that state and receive a reward of zero.

\paragraph{Markov Decision Processes}
We focus on solving Markov decision processes~(MDPs)~\cite{puterman205markov}, modeled by a tuple $(\bar{\states}, \actions, \bar{p}, \bar{r}, \bm{\bar{\mu}})$, where $\bar{\states} = \{1,2, \dots , S, S+1\}$ is the finite set of states and $\actions=\{1,2,\ldots, A\}$ is the finite set of actions. The transition function $\bar{p}\colon \bar{\states} \times \actions \to \probs{\bar{\states}}$ represents the probability $\bar{p}(s, a, s')$ of transitioning to $s'\in \bar{\mathcal{S}}$ after taking $a\in \mathcal{A}$ in $s\in \bar{\mathcal{S}}$ and $\bm{\bar{p}}_{sa} \in \probs{\bar{\states}}$ is such that $(\bm{\bar{p}}_{sa})_{s'} = \bar{p}(s,a,s')$. The function $\bar{r}\colon \bar{\states} \times  \actions \times  \bar{\states} \to \Real$ represents the reward $\bar{r}(s,a,s') \in \Real$ associated with transitioning from $s\in \bar{\states}$ and $a\in \mathcal{A}$ to $s'\in \bar{\states}$. The vector $\bm{\bar{\mu}} \in \probs{\bar{\states}}$ is the initial state distribution. 

We designate the state $e := S + 1$ as a \emph{sink state} and use $\mathcal{S} = \left\{ 1, \dots , S \right\}$ to denote the set of all non-sink states. The sink state $e$ must satisfy that $\bar{p}(e,a,e) = 1$ and $\bar{r}(e,a,e) = 0$ for each $a\in \mathcal{A}$, and $\bar{\mu}_e = 0$. Throughout the paper, we use a bar to indicate whether the quantity involves the sink state $e$. Note that the sink state can indicate a goal when all rewards are negative and an undesirable terminal state when all rewards are positive. 

The following technical assumption is needed to simplify the derivation. 
\begin{assumption} \label{asm:positive-initial}
  The initial distribution $\bm{\mu}$ satisfies that
  \[
   \bm{\mu} > \bm{0}.  
  \]
\end{assumption}

To lift \cref{asm:positive-initial}, one needs to carefully account for infinite values, which adds complexity to the results and distracts from the main ideas.

The solution to an MDP is a {\em policy}. Given a horizon $t \in \mathbb{N}$, a history-dependent policy in the set $\PiHR^t$  maps the history of states and actions to a distribution over actions. A \emph{Markov policy} $\pi \in \PiMR^t$ is a sequence of decision rules $\pi = (\bm{d}_0, \bm{d}_1, \dots , \bm{d}_{t-1})$ with $\bm{d}_k\colon  \states \to  \probs{\actions}$ the decision rule for taking actions at time $k$. The set of all \emph{randomized decision rules} is $\mathcal{D} = (\probs{\actions})^\states$. \emph{Stationary policies} $\PiSR$ are Markov policies with $\pi := (\bm{d})_{\infty} := (\bm{d}, \bm{d}, \dots)$ with the identical decision rule in every timestep. We treat decision rules and stationary policies interchangeably. The sets of \emph{deterministic} Markov and stationary policies are denoted by $\PiMD^t$ and $\PiSD$. Finally, we omit the superscript $t$ to indicate infinite horizon definitions of policies.  

The risk-neutral Total Reward Criterion~(TRC) objective is:
\begin{equation} \label{eq:objective-expected}
  \sup_{\pi\in \PiHR} \liminf_{t\to \infty}
  \E^{\pi,\bm{\mu}} \left[ \sum_{k = 0}^{t-1} r(\tilde{s}_k, \tilde{a}_k, \tilde{s}_{k+1} ) \right],
\end{equation}
where the random variables are denoted by a tilde and $\tilde{s}_k$ and $\tilde{a}_k$ represent the state from $\bar{\states}$ and action at time $k$.  The superscript $\pi$ denotes the policy that governs the actions $\tilde{a}_k$ when visiting $\tilde{s}_{k}$, and $\bm{\mu}$ denotes the initial distribution.
Finally, note that $\liminf$ gives a conservative estimate of a policy's return since the limit does not necessarily exist for non-stationary policies.

Unlike the discounted criterion, the risk-neutral TRC may be unbounded, optimal policies may not exist, or may be non-stationary~\cite{bertsekas2013stochastic,james2006analysis}. To circumvent these issues, we assume that all policies have a positive probability of eventually transitioning to the sink state. 
\begin{assumption} \label{def:transient-mdp}
  The MDP is \emph{transient}. That is, for each $\pi\in \Pi_{\mathrm{SD}}$:
  \begin{equation} \label{eq:transient-condition}
    \sum_{t = 0}^{\infty} \P^{\pi,s}\left[\tilde{s}_t = s'\right] \;<\;  \infty, \qquad
    \forall s,s'\in \mathcal{S} .  
\end{equation}
\end{assumption}

\Cref{def:transient-mdp} underlies most of our results. Transient MDPs are important because their optimal policies exist and can be chosen to be stationary deterministic~\citep[theorem~4.12]{Kallenberg2021markov}. Transient MDPs are also common in stochastic games~\cite{filar2012competitive} and generalize the stochastic shortest path problem~\cite{bertsekas2013stochastic}. 

An important tool in their analysis is the \emph{spectral radius} $\rho\colon \Real^{n \times  n} \to  \Real$ which is defined for each $\bm{A}\in \Real^{n \times  n}$ as the maximum absolute eigenvalue: $\rho(\bm{A}) := \max_{i=1, \dots , n} |\lambda_i|$ where $\lambda_i$ is the $i$-th eigenvalue~\cite{Horn2013}.
\begin{lemma}[Theorem 4.8~in~\cite{Kallenberg2021markov}] \label{lem:transient-spectral-radius}
An MDP is transient if and only if $\rho(\bm{P}^{\pi}) < 1$ for all $\pi\in \PiSR$.
\end{lemma}

Now, we discuss the differences between a discounted MDP and a transient MDP, which are useful in demonstrating the behavior of risk-averse objectives. The discounted infinite-horizon ERM objective for a discount factor $\gamma\in (0,1)$, a policy $\pi\in \PiSD$, and $\beta > 0$ is 
\begin{equation} \label{eq:discounted-objective}
  \rho_{\gamma}(\pi, \beta) := \ermo^{\pi,\bm{\mu}} \left[\sum_{t=0}^{\infty} \gamma^t \cdot
    r(\tilde{s}_t, \tilde{a}_t, \tilde{s}_{t+1})\right].
\end{equation}

In the risk-neutral setting, it is well-known that the discounted objective can be interpreted as TRC~\citep[Section 1.10]{Altman1998}. That is, every discounted MDP can be converted to a transient MDP. Given an MDP $\mathcal{M} = (\states, \actions, p, r, \mu)$, we construct a transient MDP $\bar{\mathcal{M}}_{\gamma} = (\bar{\states}, \actions, \bar{p}, \bar{r}, \bar{\mu})$ such that $\bar{\mathcal{S}} = \mathcal{S} \cup \left\{ e \right\}$, and $\bar{\mu}(s) = \mu(s), \forall s\in \mathcal{S}$ and $\mu(e) = 0$. The transition function $\bar{p}$ is defined as
\begin{equation}
\label{eq:convert-p}
  \bar{p}(s,a,s') \;=\; 
  \begin{cases}
    \gamma \cdot  p(s,a,s') \quad&\text{if} \quad s,s'\in \mathcal{S}, a\in \mathcal{A}, \\
    1-\gamma \quad&\text{if} \quad s\in \mathcal{S}, s' = e, a\in \mathcal{A}, \\
    1 \quad&\text{if} \quad s = s' = e, a\in \mathcal{A}, \\
    0 \quad&\text{otherwise}.
  \end{cases}
\end{equation}
When the rewards $r\colon \mathcal{S} \times \mathcal{A} \to \Real$ are independent of the next state, then $\bar{r}\colon \bar{\mathcal{S}} \times \mathcal{A} \to \Real$ are defined as
\begin{equation}
\label{eq:convert-r}
      \bar{r}(s,a) =
  \begin{cases}
    r(s,a) &\text{if } s\in \mathcal{S}, \\
    0 &\text{otherwise}.
  \end{cases}
\end{equation}

The model can be readily extended to account for the target state dependence by constructing an $\bar{\mathcal{M}}_{\gamma}$ with the reward function $\bar{r}\colon \bar{\mathcal{S}} \times  \mathcal{A} \times \bar{\mathcal{S}} \to  \Real$ as 
\[
  \bar{r}(s,a,s') =
  \begin{cases}
    \gamma^{-1} \cdot r(s,a,s') &\text{if } s, s'\in \mathcal{S}, \\
    0 &\text{otherwise}.
  \end{cases}
\]

Recall that when $\beta = 0$, we have $\lim_{\beta \rightarrow 0^+} \ermo[\tilde{x}] = \E[\tilde{x}]$.

\cref{prop:discount-ssp} shows that the expected total returns are identical in a discounted MDP $\mathcal{M}$ and in a transient MDP $\bar{\mathcal{M}}_{\gamma}$.
\begin{proposition} \label{prop:discount-ssp}
For each $\gamma\in [0,1)$ and MDPs $\mathcal{M}, \bar{\mathcal{M}}_{\gamma}$ constructed in \cref{sec:preliminaries}, we have that
\[
\rho_{\gamma}(\pi, 0) \; =\;
g_{\infty}( \pi, 0),
\quad \pi\in \PiSD.
\]
where $\rho_{\gamma}(\pi, 0)$ is defined in~\eqref{eq:discounted-objective} and $g_{\infty}( \pi, 0)$ is defined in~\eqref{eq:g-definitions}. \end{proposition}
\begin{proof}
The proposition follows from the construction above by algebraic manipulation~\citep[section 1.10]{Altman1998}. 
\end{proof}

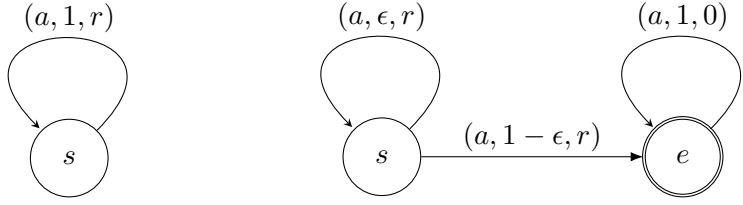
\begin{figure}
  \centering
\begin{tikzpicture}[->,-Latex,>=stealth,font=\small,node distance=40mm,el/.style = {inner sep=2pt, align=left, sloped}]
\node[ state] (1) {$s$};
\draw  
  (1) edge[loop,above] node[el,above]{$(a,1,r)$} (1);
\end{tikzpicture}
\begin{tikzpicture}[->,-Latex,>=stealth,font=\small,node distance=40mm,el/.style = {inner sep=2pt, align=left, sloped}]
\node[state] (1) {$s$};
\node[state, right of=1,double] (4) {$e$};
\draw  
  (1) edge[loop,above] node[el,above]{$(a,\epsilon, r)$} (1)
  (1) edge[ left, below] node[el,above,pos=0.5]{$(a,1-\epsilon,r)$} (4)
  (4) edge[loop,above] node[el,above]{$(a,1,0)$} (4);
\end{tikzpicture}
 \caption{Left: a discounted MDP, Right: a transient MDP}
 \label{fig:discounted-transient-mdp}
\end{figure}

Consider an example in \cref{fig:discounted-transient-mdp}. There is one non-sink state $s$ and one action $a$. A triple tuple represents an action, a transition probability, and a reward separately. For the discounted MDP, the discount factor is $\gamma$. For the transient MDP, $e$ is the sink state, and there is a positive probability $1-\epsilon$ that it will transit from state $s$ to state $e$. Once the agent reaches the state $e$, it stays in $e$.

\paragraph{Monetary risk measures}
Monetary risk measures aim to generalize the expectation operator to account for the spread of the random variable. 
\emph{Entropic risk measure}~(ERM) is a popular risk measure, defined for any risk level $\beta > 0$ and $\tilde{x}\in \mathbb{X}$ as~\cite{Follmer2016stochastic}
\begin{align} \label{eq:defn_ent_risk}
  \erm{\beta}{\tilde{x}} \;=\; - \beta^{-1} \cdot \log  \E \exp{-\beta\cdot  \tilde{x}} .
\end{align}
and extended to $\beta \in [0, \infty]$ as $\ermo_0[\tilde{x}] = \lim_{\beta \to 0^{+}} \erm{\beta}{\tilde{x}} = \E[\tilde{x}] $ and $\ermo_{\infty}[\tilde{x}] = \lim_{\beta\to \infty} \erm{\beta}{\tilde{x}} = \operatorname{ess} \inf[\tilde{x}]$.
ERM plays a unique role in sequential decision-making because it is the only law-invariant risk measure that satisfies the tower property shown in \cref{pro:erm-tower-property}, which is essential in constructing dynamic programs~\cite{hau2023entropic}. Unfortunately, two significant limitations of ERM hinder its practical applications. First, ERM is not positively homogenous and, therefore, the risk value depends on the scale of the rewards, and ERM is not coherent~\cite{Follmer2016stochastic, hau2023entropic, Ahmadi-Javid2012}. Second, the risk parameter $\beta$ is challenging to interpret and does not relate well to other standard risk measures, like VaR or CVaR.

For these reasons, we focus on the \emph{Entropic Value at Risk}~(EVaR), defined as, for a given $\alpha \in (0,1)$,
\begin{equation}\label{eq:evar-def-app}
  \begin{aligned}
    \evar{\alpha}{\tilde{x}}
   & \;=\;
    \sup_{\beta>0}  -\beta^{-1} \log \left(\alpha^{-1} \E \exp{ -\beta \tilde{x} } \right) \\
   & \;=\;
    \sup_{\beta>0}  \erm{\beta}{\tilde{x}} + \beta^{-1} \log  \alpha,
  \end{aligned}
\end{equation}
and extended to $\evar{0}{\tilde{x}} = \ess \inf [\tilde{x}]$ and $\evar{1}{\tilde{x}} = \Ex{\tilde{x}}$~\cite{Ahmadi-Javid2012}. It is important to note that the supremum in~\eqref{eq:evar-def-app} may not be attained even when $\tilde{x}$ is a finite discrete random variable~\cite{ahmadi2017analytical}.

EVaR addresses the limitations of ERM while preserving its benefits. EVaR is coherent and positively homogenous. EVaR is also a good approximation to interpretable quantile-based risk measures, like VaR and CVaR~\cite{Ahmadi-Javid2012,hau2023entropic}.

\begin{proposition}{Tower Property for ERM \cite[theorem~3.1]{hau2023entropic}}

   \label{pro:erm-tower-property} 
   For any two random variables $\tilde{x}_1,\tilde{x}_2 \in \mathbb{X}$, we have
\[
\ermo_{\beta}[\tilde{x}_1] = \ermo_{\beta}[\ermo_{\beta}[\tilde{x}_1 \mid  \tilde{x}_2]],
\]
where the conditional ERM is defined as 
\[
\ermo_{\beta}[\tilde{x}_1 | \tilde{x}_2] = -\beta^{-1} \log(\E[e^{-\beta \tilde{x}_1} \mid \tilde{x}_2])
\]
\end{proposition}

\paragraph{Risk-averse MDPs.}
Risk-averse MDPs, using static VaR and CVaR risk measures, under the discounted criterion received abundant attention~\cite{Hau2023a, bauerle2011markov, bauerle2022markov, pflug2016time, li2022quantile}, showing that these objectives require history-dependent optimal policies. In contrast, nested risk measures under the TRC may admit stationary policies that can be computed using dynamic programming~\cite{ahmadi2021risk,meggendorfer2022risk,de2020risk,gavriel2012risk}. However, the TRC with nested CVaR can be unbounded, as shown in \cref{thm:cvar-counter-example}. Recent work has shown that optimal Markov policies exist for EVaR discounted objectives, and they can be computed via dynamic programming~\cite{hau2023entropic}, building upon similar results established for ERM~\cite{Chung1987}. However, in TRC with ERM, the value functions may also be unbounded~\citep[proposition D.1]{su2024stationary}.

The following proposition contradicts \cite[theorem 1]{ahmadi2021risk}.
\begin{proposition}
\label{thm:cvar-counter-example}
There exists a transient MDP and a risk level $\alpha \in (0,1)$ such that the TRC with nested CVaR or EVaR is unbounded. 
\end{proposition}
\begin{proof}%[Proof of \cref{thm:cvar-counter-example}]
Consider the example of a transient MDP in \cref{fig:discounted-transient-mdp} with $r=-1$, a risk level $\alpha \in (0,1)$, and some $\epsilon \in (0,1)$ such that $\epsilon \ge \alpha$. This MDP admits only a single policy because it only has one action $a$, which we omit in the notation below. 

Recall that the nested CVaR objective for some $s \neq e$ is
\begin{equation*}
  v_t(s)
  \; =\;
  \cvaro^{s}_{\alpha}\left[r(\tilde{s}_0, a)
+\cvaro_{\alpha}^{\tilde{s}_1}\left[r(\tilde{s}_1, a) + \cdots 
\cvaro^{\tilde{s}_{t-1}}_{\alpha}\left[r(\tilde{s}_{t-1}, a) \right] \right] \right],
\end{equation*}

 The value function for the non-terminal state can be computed using a dynamic program as~\citep[theorem~1]{ahmadi2021risk}:
\begin{align*}
  v_t(s)
  &= \cvaro^{s}_{\alpha}[r(s,a) + v_{t-1}(\tilde{s}_1)] 
  = -1 + \min \left\{ q_1 \cdot v_{t-1}(s) \mid q\in \Delta^2,\, 
  q_1 \le \frac{\epsilon}{\alpha}, q_2 \le \frac{1-\epsilon}{\alpha} \right\} \\
  &\le -1 - v_{t-1}(s).
\end{align*}
Here, we used the dual representation of CVaR. Then, by induction, $v_t(s) \le -t$ and, therefore, $\lim_{t\to \infty} v_t(s) = -\infty$.
\end{proof}

\section{Solving ERM Total Reward Criterion}
\label{sec:ssp-with-erm}

This section shows that an optimal stationary policy exists for ERM-TRC and that the value function satisfies the dynamic programming equations. We then outline algorithms for computing it.

Our objective in this section is to maximize the ERM-TRC objective for some given $\beta > 0$ defined as
\begin{equation} \label{eq:sup-erm}
  \sup_{\pi\in \PiHR}  \liminf_{t\to \infty}
  \ermp{\beta}{\pi,\bm{\mu}}
   {\sum_{k=0}^{t-1} r(\tilde{s}_k,\tilde{a}_k,\tilde{s}_{k+1})}.
 \end{equation}
% Analogously to~\eqref{eq:objective-expected}, the superscripts indicate the policy and initial state distribution that govern $\tilde{s}_k$ and $\tilde{a}_k$. 
The definition employs limit inferior because the limit may not exist for non-stationary policies. 
Return functions $g_t\colon \PiHR \times \Real_{++} \to  \Real$ and $g_t\opt \colon \Real_{++} \to \Real$ for a horizon $t \in \mathbb{N}$ and the infinite-horizon versions $g_t\colon \PiHR \times \Real_{++} \to  \bar{\Real}$ and $g_t\opt \colon \Real_{++} \to \bar{\Real}$ are defined 
\begin{equation}
  \label{eq:g-definitions}
  \begin{aligned}
  g_t(\pi, \beta)  &:= \ermp{\beta}{\pi,\bm{\mu}}
  {\sum_{k=0}^{t-1} r(\tilde{s}_k,\tilde{a}_k,\tilde{s}_{k+1})}, \\
    g_t\opt(\beta) &:= \sup_{\pi\in \PiHR} g_t(\pi, \beta), \\
  g_{\infty}(\pi, \beta) &:= \liminf_{t\to \infty} g_t(\pi, \beta), \\
   g_{\infty}\opt(\beta) &:= \liminf_{t\to \infty} g_t\opt (\beta). 
  \end{aligned}
\end{equation}
Note that the functions $g_{\infty}$ and $g_{\infty}\opt$ can return infinite values and that~\eqref{eq:sup-erm} differs from  $g\opt_{\infty}$ in the order of the limit and the supremum. Finally,  when $\beta = 0$, we assume that all $g$ functions are defined as the expectation.
In the remainder of the section, we assume that the risk level $\beta > 0$ is fixed and omit it in notations when its value is unambiguous from the context. 

The following lemmas show some properties of the function $g$ defined in ~\eqref{eq:g-definitions}. 
\begin{lemma} \label{lem:erm-convergence-0}
  The functions $g$ defined in~\eqref{eq:g-definitions} satisfy for each $\pi\in \PiSD$ and $t \in \mathbb{N}$ that
  \begin{align*}
  \lim_{\beta\to 0} g_t(\pi, \beta) &= g_t(\pi, 0), &
  \lim_{\beta\to 0} g_t\opt (\beta) &= g_t\opt(0), \\
  \lim_{\beta\to 0} g_{\infty}(\pi, \beta) &= g_{\infty}(\pi, 0), &
  \lim_{\beta\to 0} g_{\infty}\opt (\beta) &= g_{\infty}\opt(0).
  \end{align*}
\end{lemma}
\begin{proof}
The result for $g_t$ holds from the continuity of ERM as $\beta \to 0^+$; e.g. ~\cite[remark~2.8]{ahmadi2017analytical}. The result for $g_t\opt$ then follows from the fact that $\PiSD$ is finite. Then, \cite[lemma~1]{patek2001terminating} shows that there exists $\beta_0 > 0$ such that $g_t \to g_{\infty}$ uniformly on $(0, \beta_0)$ and the limits can be exchanged. Therefore, the result also holds for $g_{\infty}$ and $g_{\infty}\opt$.
\end{proof}

\begin{lemma} \label{lem:erm-decreasing}
  The functions $g$ defined in~\eqref{eq:g-definitions} are monotonically non-increasing in the parameter $\beta \ge 0$.
\end{lemma}
\begin{proof}
This well-known result follows, for example, from the dual representation of ERM~\cite{hau2023entropic}.
\end{proof}

\subsection{Finite Horizon } 

We commence the analysis with definitions and basic properties for the finite horizon criterion. To the best of our knowledge, this analysis is original in the context of the ERM but builds on similar approaches employed in the study of exponential utility functions.

Finite-horizon functions $\bm{v}^t( \pi ) \in \Real^{S}$ and $\bm{v}^{t,\star}\in \Real^{S}$ are defined for each horizon $t \in \mathbb{N}$ and policy $\pi\in \PiMD$, $s\in \mathcal{S}$ as 
\begin{equation}
\label{eq:erm-value-function}
\begin{aligned}
      v^t_s(\pi)
 & \;:=\; \ermp{\beta}{\pi,s}{\sum_{k=0}^{t-1} r(\tilde{s}_k,\tilde{a}_k,\tilde{s}_{k+1})},\\
  v^{t,\star}_s & \;:=\; \max_{\pi\in \PiMD} v^t_s(\pi),
\end{aligned}
\end{equation}
and $v_{e}^t(\pi) := 0$.

Because the nonlinearity of ERM complicates the analysis, it will be convenient to instead rely on \emph{exponential value function} $\bm{w}^{t}(\pi) \in \mathbb{R}^{S}$ for $\pi\in \PiMD$, $t \in \mathbb{N}$, and $s\in \states$ that satisfy
\begin{align} \label{eq:exp-value-function}
w_s^t(\pi)
&:= - \exp {  -\beta \cdot  v_s^t(\pi)  }, \\
v^t_s(\pi)
&= - \beta^{-1} \log (-  w^t_s(\pi)).
\end{align}
The optimal $\bm{w}^{t,\star} \in \mathbb{R}^{S}$ is defined analogously from $\bm{v}^{t,\star}$. Note that $\bm{w}^t < \bm{0}$ (componentwise) and $\bm{w}^0(\pi) = \bm{w}^{0,\star} = -\bm{1}$ for any $\pi \in \PiMD$. Similar exponential value functions have been used previously in exponential utility function objectives~\cite{denardo1979optimal,patek2001terminating}, in the analysis of robust MDPs, and even in regularized MDPs (see \cite{Grand-Clement2022} and references therein).

One can define a corresponding \emph{exponential Bellman operator} for any $\bm{w}\in \Real^S$ as
\begin{equation} \label{eq:exp-bellman-definition}
  \begin{aligned}
  \BellIE^{\bm{d}} \bm{w} &\;:=\;
  \bm{B}^{\bm{d}} \bm{w} - \bm{b}^{\bm{d}},
  \\
  \BellIE\opt  \bm{w} &\;:=\;
  \max_{\bm{d}\in \mathcal{D}} L^{\bm{d}} \bm{w} = \max_{\bm{d}\in \ext \mathcal{D}} L^{\bm{d}} \bm{w},
  \end{aligned}
\end{equation}
where $\ext \mathcal{D}$ is the set of extreme points of $\mathcal{D}$ corresponding to deterministic decision rules and $\bm{B}^{\bm{d}} \in \Real_+^{S \times   S}$ and $\bm{b}^{\bm{d}} \in  \Real_+^{S}$ are defined for $s, s'\in \mathcal{S}$ and $\bm{d}\in \mathcal{D}$ as
\begin{subequations} \label{eq:exponential-definitions}
\begin{align}
B_{s,s'}^{\bm{d}}
& :=
\sum_{a\in \mathcal{A}}  p(s, a, s') \cdot d_a(s) \cdot e^{-\beta \cdot  r(s,a,s')} ,\\
b_{s}^{\bm{d}} 
&:=
\sum_{a\in \mathcal{A}}  p(s, a, e) \cdot d_a(s) \cdot e^{-\beta \cdot  r(s,a,e)} .
\end{align}
\end{subequations}

We state some properties of the exponential Bellman operator in \cref{lem:exp-bellman-monotone} and \cref{lem:exp-bellman-continuous}.
\begin{lemma}\label{lem:exp-bellman-monotone}
The exponential Bellman operator is monotone. That is, for $\bm{x},\bm{y} \in \Real^S$
\begin{align}
  \label{eq:exp-bellman-monotone}
  \bm{x} \ge \bm{y}
  &\quad \Rightarrow \quad
    L^{\bm{d}} \bm{x} \ge L^{\bm{d}} \bm{y}, \qquad \forall \bm{d}\in \mathcal{D} \\
  \label{eq:exp-bellman-opt-monotone}
  \bm{x} \ge \bm{y}
  &\quad \Rightarrow \quad
  L\opt \bm{x} \ge L\opt \bm{y}.
\end{align}
\end{lemma}
\begin{proof}
The property in~\eqref{eq:exp-bellman-monotone} follows immediately from non-negativity of $\bm{B}^{\bm{d}}$. The property in~\eqref{eq:exp-bellman-opt-monotone} then follows from the monotonicity of the $\max$ operator.
\end{proof}

\begin{lemma}\label{lem:exp-bellman-continuous}
The exponential Bellman operators $L^{\bm{d}},\,  \forall \bm{d}\in \mathcal{D}$ and $L\opt$ are continuous.
\end{lemma}
\begin{proof}
The lemma follows directly from the continuity of linear operators and from the fact that the pointwise maximum of a finite number of continuous functions is continuous. See also \citep[lemma~5]{patek2001terminating}
\end{proof}

The following theorem shows that $L$ can be used to compute $\bm{w}$. We use the shorthand notation $\pi_{1{:}t{-}1} = (\bm{d}_1, \dots , \bm{d}_{t-1}) \in \PiMR^{t-1}$ to denote the tail of $\pi$ that starts with $\bm{d}_1$ instead of $\bm{d}_0$.
\begin{theorem}\label{thm:exponential-bellman}
  For each $t =1, \dots $, and $\pi = (\bm{d}_0, \dots , \bm{d}_{t-1}) \in \PiMR^t$, the exponential values satisfy that
\[
\begin{aligned}
\bm{w}^t(\pi) &= L^{\bm{d}_t} \bm{w}^{t-1}(\pi_{1{:}t{-}1}),
 & \bm{w}^0(\pi) &= - \bm{1},
\\
\bm{w}^{t, \star } &= L\opt \bm{w}^{t-1,\star } = \bm{w}^t(\pi\opt) \ge \bm{w}^t(\pi),
& \bm{w}^{0,\star } &= -\bm{1},
\end{aligned}
\]
for some $\pi\opt \in \PiMD^t$.
\end{theorem}
The proof of \cref{thm:exponential-bellman} is standard and has been established both in the context of ERMs~\cite{hau2023entropic} and exponential utility functions~\cite{patek1997stochastic}.
\begin{proof}[Proof of \cref{thm:exponential-bellman}]
To construct the value function, we can define a Bellman operator $\BellI^{\bm{d}}\colon \Real^S \to  \Real^S$ for any decision rule $\bm{d}\colon  \mathcal{S} \to  \probs{A}$ and the optimal Bellman operator $\BellI\opt \colon \Real^S \to \Real^S$ for a \emph{value vector} $\bm{v}\in \Real^S$ as
\begin{equation} \label{eq:bellman-erm}
\begin{aligned}
  (\BellI^{\bm{d}} \bm{v})_s
  &\;:=\; \ermp{\beta}{\bm{d},s}{ r(s,\tilde{a}_0, \tilde{s}_1) + v_{\tilde{s}_1} },  \\
  \BellI\opt \bm{v}
  &\;:=\; \max_{\bm{d}\in \mathcal{D}}   \BellI^{\bm{d}} \bm{v} = \max_{\bm{d}\in \ext \mathcal{D}}   \BellI^{\bm{d}} \bm{v}.
\end{aligned}
\end{equation}
It is easy to see that $\bm{d}$ in \eqref{eq:bellman-erm} can be chosen independently for each state to maximize $\bm{v}$ uniformly across states. The optimality of deterministic decision rules, $\bm{d} \in \ext \mathcal{D}$, follows because ERM is a mixture quasi-convex function~\cite{Delage2019}.

The existence of a value function for a finite-horizon problem under the ERM objective has been analyzed previously~\cite{hau2023entropic}, including in the context of exponential utility functions~\cite{Chung1987}.

To derive the exponential Bellman operator for the exponential value function for $\bm{d}\colon \mathcal{S}\to \probs{A}$, we concatenate the Bellman operator with the transformations to and from the exponential value function: 
\begin{equation} \label{eq:erm-exp-bellman}
\begin{aligned}
 (\BellIE^{\bm{d}} \bm{w})_s 
  &\;=\; - \exp {-\beta \cdot T^{\bm{d}} ( - \beta^{-1} \log (-  \bm{w}) ) }\\
  &\; =\;  -\E^{\bm{d},s}\left[\exp{-\beta \cdot  r(s,\tilde{a}_0,\tilde{s}_1)  + \log (-w_{\tilde{s}_1}) } \right] \\
  &\; =\;  -\E^{\bm{d},s}\left[\exp{-\beta \cdot  r(s,\tilde{a}_0,\tilde{s}_1)}  \cdot  (-w_{\tilde{s}_1})  \right] \\
  &\; =\; \sum_{s'\in \bar{\mathcal{S}}} \sum_{a\in \mathcal{A}}  p(s, a, s') \cdot d_a(s) \cdot \exp{-\beta \cdot  r(s,a,s')}  \cdot  w_{s'} \\
  &\; =\;  \sum_{s'\in \mathcal{S}} \sum_{a\in \mathcal{A}}  p(s, a, s') \cdot d_a(s) \cdot \exp{-\beta \cdot  r(s,a,s')}  \cdot  w_{s'} \\
   & \qquad \qquad-  \sum_{a\in \mathcal{A}}  p(s, a, e) \cdot d_a(s) \cdot \exp{-\beta \cdot  r(s,a,e)}.
\end{aligned}
\end{equation}

The derivation above uses the fact that $w_e = -1$ since $v_e = 0$ by definition. The statement of the theorem then follows by algebraic manipulation of $\bm{B}^{\bm{d}}, \bm{b}^{\bm{d}}$ and by induction on $t$. The base case hold by the definition of $\bm{w}^0(\pi) = \bm{w}^{0,\star } = -\bm{1}$.

The existence of an optimal $\pi\opt$ follows by choosing the maximum in the definition of $L\opt $, which is attained by compactness and continuity of the objective. 
\end{proof}

The following corollary follows directly from \cref{thm:exponential-bellman} by algebraic manipulation and by the monotonicity of exponential value function transformation and the ERM.
\begin{corollary} 
\label{coro:g-pi-beta}
We have that 
\begin{align*} 
g_t(\pi, \beta)  &= \ermo^{\bm{\mu}}_{\beta}\left[ v^t_{\tilde{s}_0}(\pi) \right] ,\\
  g_t\opt(\beta) &= \ermo^{\bm{\mu}}_{\beta}\left[ v^{t,\star}_{\tilde{s}_0} \right]
                   = \max_{\pi\in \PiMD} \ermo^{\bm{\mu}}_{\beta}\left[ v^t_{\tilde{s}_0} (\pi) \right].
\end{align*}
\end{corollary}

In a discounted MDP, the discount factor ensures that the ERM value function is always bounded. However, for the total-reward criterion, the ERM value function can be unbounded, as shown in \cref{prop:erm-unbounded}.

\begin{proposition} \label{prop:erm-unbounded}
There exists a transient MDP and a risk level $\beta > 0$ such that $g_{\infty}\opt(\beta) = -\infty$.
\end{proposition}
\begin{proof}
We use the transient MDP described in \cref{fig:discounted-transient-mdp} to show this result. Because the returns of this MDP follow a truncated geometric distribution, its risk-averse return for each $\beta>0$ and $\epsilon \in (0,1)$ can be expressed analytically for $t \ge 1$ as
\begin{equation} \label{eq:exp-erm-value-func} 
\begin{aligned}
 \ermp{\beta}{\pi}{\sum_{k=0}^{t-1} r(\tilde{s}_k,\tilde{a}_k,\tilde{s}_{k+1})} 
&= -\frac{1}{\beta}\log\left(  \sum_{k = 0}^{t-1} (1-\epsilon) \epsilon^k \cdot \exp{-\beta \cdot k \cdot  r}  + \epsilon^t \cdot 0 \right) \\
&=  -\frac{1}{\beta}\log\left(  \sum_{k = 0}^{t-1} (1-\epsilon) \epsilon^{k} \cdot \exp{-\beta \cdot  r}^k  \right).
\end{aligned}
\end{equation}
Here, $(1-\epsilon) \epsilon^k$ is the probability that the process terminates after exactly $k$ steps, and $\epsilon^t$ is the probability that the process does not terminate before reaching the horizon. Then, using the fact that a geometric series $\sum_{i=0}^{\infty} a \cdot q^i$ for $a \neq 0$ is bounded if and only if $|q| < 1$ we get that
\begin{equation*}
\lim_{t\to \infty} \ermp{\beta}{\pi}{\sum_{k=0}^{t-1} r(\tilde{s}_k,\tilde{a}_k,\tilde{s}_{k+1})}   > - \infty 
\qquad\Leftrightarrow \qquad
\epsilon \cdot \exp{-\beta \cdot  r}  < 1. 
\end{equation*}

Note that $\epsilon \cdot \exp{-\beta \cdot r} \ge 0$ from its definition. Then, setting $r = -1$ and $\beta > -\log \epsilon$ proves the result. 
\end{proof}

Unlike the risk-neutral setting,  for risk-averse objectives, such as ERM, the value functions of a discounted MDP can differ from those in a transient MDP, as the following proposition shows. 
\begin{proposition}
\label{prop:erm-value-function-diverge}
There exists an MDP $\mathcal{M}$, $\gamma > 0$, $\beta > 0$, $\bar{\mathcal{M}}_{\gamma}$ constructed in \cref{sec:preliminaries}, and $\pi\in \PiSD$ such that
\[
\rho_{\gamma}( \pi, \beta) \; \neq \;
g_{\infty}( \pi, \beta),
\]
where $\rho_{\gamma}(\pi, 0)$ is defined in~\eqref{eq:discounted-objective} and $g_{\infty}( \pi, 0)$ is defined in~\eqref{eq:g-definitions}.
\end{proposition}
\begin{proof}
See the one-state example described in \cref{sec:numerical-eval}.
\end{proof}

\subsection{Infinite Horizon}

We now turn to constructing infinite-horizon optimal policies as a limiting case of the finite-horizon case. An important quantity is the infinite-horizon exponential value function defined for each $\pi\in \PiHR$ as
\[
  \bm{w}^{\infty}(\pi)
  \; :=\;  \liminf_{t \to \infty} \bm{w}^t(\pi),
  \quad
  \bm{w}^{\infty,\star}
  \; :=\;  \liminf_{t \to \infty} \bm{w}^{t,\star}.
\]
Note again that we use the inferior limit because the limit may not be defined for non-stationary policies. The limiting infinite-horizon value functions $\bm{w}^{\infty}(\pi)$ and $\bm{w}^{\infty,\star}$ are defined analogously from $\bm{v}^t(\pi)$ and $\bm{v}^{t,\star}$ using the inferior limit. 
The following theorem is the main result of this section. It shows that for an infinite horizon, the optimal exponential value function is attained by a stationary deterministic policy and is a fixed point of the exponential Bellman operator. 
\begin{theorem} \label{thm:erm-main-convergence}
Whenever $\bm{w}^{\infty,\star } > -\bm{\infty}$ there exists $\pi\opt = (\bm{d}\opt)_{\infty} \in \PiSD$ such that
  \[
    \bm{w}^{\infty, \star }
    \; =\; 
    \bm{w}^{\infty}(\pi\opt)
    \; =\; 
    L^{\bm{d}\opt} \bm{w}^{\infty, \star },
  \]
and $\bm{w}^{\infty, \star }$ is the unique value that satisfies this equation.
\end{theorem}

The main idea of the proof of \cref{thm:erm-main-convergence} is to show that whenever the exponential value functions are bounded, the exponential Bellman operator must be \emph{weighted-norm} contraction with a unique fixed point. To facilitate the analysis, we define $\bm{w}^t\colon \PiSR^t \times \Real^{S} \to \Real^{S}, t \in \mathbb{N}$ for $\bm{z}\in \Real^{S},\,\pi \in \PiSR^t$, as
\begin{equation} \label{eq:exp-value-expression}
\begin{aligned}
      \bm{w}^t(\pi, \bm{z})
    \; =\; & \BellIE^{\bm{d}} \bm{w}(\pi_{1{:}t{-}1}) 
     \; =\; \BellIE^{\bm{d}} \BellIE^{\bm{d}} \dots \BellIE^{\bm{d}} (-\bm{z}) \\
    \; =\;& - (\bm{B}^{\bm{d}})^t \bm{\bm{z}} - \sum_{k=0}^{t-1}  (\bm{B}^{\bm{d}})^k \bm{b}^{\bm{d}} .
\end{aligned}
\end{equation}
The value $\bm{z}$ can be interpreted as the exponential value function at the termination of the process following $\pi$ for $t$ periods. 
Note that \( \bm{w}^t(\pi) = \bm{w}^t(\pi, \bm{1}), \, \forall \pi \in \PiMR, t \in \mathbb{N} . \)

 We now outline the proof of \cref{thm:erm-main-convergence}; see \cite[appendix D.4]{su2024stationary} for details. To establish \cref{thm:erm-main-convergence}, we show that $\bm{w}^{t,\star }$ converges to a fixed point as $t\to \infty$. Standard arguments do not apply to our setting~\cite{puterman205markov,Kallenberg2021markov,patek2001terminating} because the ERM-TRC Bellman operator is not an $L_{\infty}$-contraction, it is not linear, and the values in value iteration do not increase or decrease monotonically. Although the exponential Bellman operator $L^{\bm{d}}$ is linear, it may not be a contraction.

Now we state \cref{lem:radius-contractions} and \cref{thm:optim-mark-polic} that are useful for the formal proof of \cref{thm:erm-main-convergence}.

\begin{lemma} \label{lem:radius-contractions}
Assume some $\pi = (\bm{d})_{\infty} \in \PiSR$ such that $\rho(\bm{B}^{\bm{d}}) < 1$. Then for all $\bm{z}\in \Real^S$
\[
   \bm{w}^{\infty}(\pi)  = \bm{w}^{\infty}(\pi, \bm{z}) = L^{\bm{d}} \bm{w}^{\infty}(\pi) > -\bm{\infty}.
  \]
\end{lemma}
\begin{proof}
The result follows by algebraic manipulation from~\eqref{eq:exp-value-expression} and basic matrix analysis. When $\rho(\bm{B}^{\bm{d}}) < 1$, we get from Neumann series~\citep[problem~5.6.P26]{Horn2013}
\[
 \lim_{t\to \infty} \sum_{k=0}^{t-1}  (\bm{B}^{\bm{d}})^k \bm{b}^{\bm{d}} =  (\bm{I} - \bm{B}^{\bm{d}})^{-1}\bm{b}^{\bm{d}},
\]
and a consequence of Gelfand's formula~\citep[theorem~4.5]{Kallenberg2021markov}
\[
  \lim_{k\to \infty} (\bm{B}^{\bm{d}})^k \bm{z} = \bm{0}.
\]
\end{proof}

\cref{thm:optim-mark-polic} can be derived from the optimality of Markov deterministic policies in MDPs with exponential utility functions~\cite{Chung1987, patek2001terminating}. 

\begin{theorem} \label{thm:optim-mark-polic}
For each $\beta > 0$, there exists an optimal deterministic Markov policy $\pi^{t,\star} \in \PiMD$ for each horizon $t \in \mathbb{N} $: 
\begin{equation*}
\begin{split}
 \max_{\pi\in \PiMD}  \ermp{\beta}{\pi,s}{\sum_{k=0}^{t-1} r(\tilde{s}_k,\tilde{a}_k,\tilde{s}_{k+1})}
  \;=\;   \max_{\pi \in \PiHR} \ermp{\beta}{\pi,s}{\sum_{k=0}^{t-1} r(\tilde{s}_k,\tilde{a}_k,\tilde{s}_{k+1})}.
  \end{split}
\end{equation*}
\end{theorem}
See \cite[Corollary~4.2]{hau2023entropic} for a proof.

We are ready to prove \cref{thm:erm-main-convergence}.
\begin{proof}[Proof of \cref{thm:erm-main-convergence}]
%For the remainder of the proof,
Let $\pi_{\mathrm{M}}\opt \in \arg\max_{\pi\in \PiMR} \bm{\mu}\tr \bm{v}^{\infty}(\pi)$ and suppose that $\bm{\mu}\tr \bm{v}^{\infty}(\pi\opt_{\mathrm{M}}) > -\infty$. Then, the exponential value function $\bm{w}^{t,\star}  = \bm{w}^t(\pi_{\mathrm{M}}\opt) \in \Real^S$ of $\pi_{\mathrm{M}}\opt $ satisfies by \cref{thm:exponential-bellman} that
\[
\bm{w}^{0,\star} = -\bm{1},
\qquad
\bm{w}^t(\pi\opt_{\mathrm{M}})  = L\opt \bm{w}^{t-1}(\pi\opt_{\mathrm{M}}), \quad t = 1, \dots .
\]

We show that $\lim_{t\to \infty} \bm{w}^{t,\star}$ exists and that a stationary policy attains it. We construct a sequence $\bm{w}_{\mathrm{u}}^t \in \Real^S, t \in \mathbb{N} $ as
\[
\bm{w}_{\mathrm{u}}^0 = \bm{0},
\qquad
\bm{w}_{\textrm{u}}^t  = L\opt \bm{w}_{\textrm{u}}^{t-1}, \quad t = 1, \dots .
\]
First, we show by induction that
\begin{equation} \label{eq:w-hat-upper-bound}
  \bm{w}_{\textrm{u}}^t \quad\ge\quad \bm{w}^{t,\star}, \qquad t \in \mathbb{N}.
\end{equation}
The base case $t=0$ follows immediately from the definitions of $\bm{w}_{\textrm{u}}^0$ and $\bm{w}^{0,\star }$. Next, suppose that~\eqref{eq:w-hat-upper-bound} holds for some $t > 0$, then it also holds for $t+1$:
\[
 \bm{w}_{\textrm{u}}^{t+1} \; =\;  L\opt \bm{w}_{\textrm{u}}^t \; \ge\;  L\opt \bm{w}^{t,\star} \; =\;   \bm{w}^{t+1,\star}, 
\]
where the inequality follows from the inductive assumption and from \cref{lem:exp-bellman-monotone}.
Second, we show by induction that
\begin{equation}\label{eq:w-hat-decreasing}
  \bm{w}_{\textrm{u}}^{t+1} \quad\le\quad \bm{w}_{\textrm{u}}^{t}, \qquad t \in \mathbb{N}.
\end{equation}
The base case for $t=0$ holds as
\[
  \bm{w}_{\textrm{u}}^1 \; =\;  L\opt \bm{w}_{\textrm{u}}^0 \; =\;  L\opt \bm{0} \; =\;  \max_{\bm{d}\in \mathcal{D}} - \bm{b}^{\bm{d}} \; \le\;  \bm{0} \; =\; \bm{w}_{\textrm{u}}^0, 
\]
where the inequality holds because $\bm{b}^{\bm{d}} \ge 0$ from its construction. To prove the inductive step, assume that~\eqref{eq:w-hat-decreasing} holds for $t>0$ and prove it for $t+1$:
\[
  \bm{w}_{\textrm{u}}^{t+1} \; =\;  L\opt \bm{w}_{\textrm{u}}^t \; \le\;  L\opt \bm{w}_{\textrm{u}}^{t-1} \; =\;  \bm{w}_{\textrm{u}}^t,
\]
where the inequality follows from the inductive assumption and from \cref{lem:exp-bellman-monotone}.

Then, using the Monotone Convergence Theorem~\cite[theorem~16.2]{Johnsonbaugh1981}, finite $\mathcal{S}$, and $\inf_{t \in \mathbb{N}}\bm{w}_{\textrm{u}}^t \ge \inf_{t \in \mathbb{N} } \bm{w}^{t, \star} > -\infty$, we get that there exists $\bm{w}_{\mathrm{u}}\opt  \in \Real^S$ such that
\[
\bm{w}_{\textrm{u}}\opt = \lim_{t\to \infty} \bm{w}_{\textrm{u}}^t,  
\]
and the limit exists. Then, taking the limit of both sides of $\bm{w}_{\textrm{u}}^t  = L\opt \bm{w}_{\textrm{u}}^{t-1}$, we have that
\begin{align*}
  \lim_{t\to \infty} \bm{w}_{\textrm{u}}^t  &= \lim_{t\to \infty} L\opt \bm{w}_{\textrm{u}}^{t-1} \\
  \bm{w}_{\textrm{u}}\opt   &= \lim_{t\to \infty} L\opt \bm{w}_{\textrm{u}}^{t-1} \\
  \bm{w}_{\textrm{u}}\opt   &=  L\opt \lim_{t\to \infty}\bm{w}_{\textrm{u}}^{t-1}  \\
  \bm{w}_{\mathrm{u}}\opt &= L^{\bm{d}\opt} \bm{w}_{\mathrm{u}}\opt ,
\end{align*}
where $\bm{d}\opt = \argmax_{\bm{d}\in \mathcal{D}} L^{\bm{d}} \bm{w}_{\mathrm{u}}$. Above, we can exchange the operators $L\opt $ and $\lim$ by the continuity of $L\opt $~(\cref{lem:exp-bellman-continuous}).

Now, define $\bm{w}_{\mathrm{l}}^t \in \Real^S, t \in \mathbb{N}$ as
\[
\bm{w}_{\mathrm{l}}^0 = -\bm{1},
\qquad
\bm{w}_{\textrm{l}}^t  = L^{\bm{d}\opt} \bm{w}_{\textrm{l}}^{t-1}, \quad t = 1, \dots .
\]
From the definition of $L\opt$ and by induction on $t$ we have that.
\[
 \bm{w}_{\mathrm{l}}^t  \le \bm{w}^{t,\star}.
\]
By \cref{lem:bounded-contraction-fixedpoint} for $\bm{z} = \bm{0}$, we have that $\rho(\bm{B}^{\bm{d}\opt}) < 1$ and therefore from \cref{lem:radius-contractions}
\[
\lim_{t\to \infty} \bm{w}_{\mathrm{l}}^{t}  = \lim_{t\to \infty}  \bm{w}_{\mathrm{u}}^t = \bm{w}_{\mathrm{u}}\opt.
\]
In addition, because
\[
 \bm{w}_{\mathrm{u}}^t\; \ge \; \bm{w}^{t,\star }\; \ge \; \bm{w}_{\mathrm{l}}^t, \qquad t \in \mathbb{N},  
\]
the Squeeze Theorem~\cite[theorem~14.3]{Johnsonbaugh1981} shows that
\[
 \lim_{t \to \infty} \bm{w}^{t, \star } = \bm{w}_{\mathrm{u}}\opt , 
\]
and $\bm{d}\opt $ is a stationary policy that attains the return of $\pi\opt_{\mathrm{M}}$.
\end{proof}

\begin{corollary} \label{cor:erm-optimal-stationary}
Assuming the hypothesis of \cref{thm:erm-main-convergence}, we have that \(   \bm{v}^{\infty, \star} =  \bm{v}^{\infty}(\pi\opt) \) and
  \[
  g_{\infty}\opt(\beta) = \ermo^{\bm{\mu}}_{\beta}\left[ v^{\infty,\star}_{\tilde{s}_0} \right]
                   = \max_{\pi\in \PiSD} \ermo^{\bm{\mu}}_{\beta}\left[ v^{\infty}_{\tilde{s}_0} (\pi) \right].
  \]
\end{corollary}

\label{proof:cor-erm-optimal-stationary}
\begin{proof}[Proof of \cref{cor:erm-optimal-stationary}]
  From the existence of an optimal stationary policy $\pi\opt \in \PiSD$ from for a sufficiently large horizon $t$ from \cref{thm:erm-main-convergence} and \cref{sec:optim-mark-polic}, we get that
\begin{equation*}
\begin{aligned} 
\bm{\mu}\tr\bm{v}^{\infty}(\pi\opt)
\le
\sup_{\pi\in \PiHR} \liminf_{t \to \infty} \bm{\mu}\tr \bm{v}^t(\pi) 
\le 
\liminf_{t \to \infty} \sup_{\pi\in \PiHR^t} \bm{\mu}\tr \bm{v}^t(\pi) 
\le
\bm{\mu}\tr \bm{v}^{\infty}(\pi\opt), 
\end{aligned}
\end{equation*}
which implies that all inequalities above hold with equality.
\end{proof}

An important technical result we show is that the only way a \emph{stationary} policy's return can be bounded is if the policy's matrix has a spectral radius strictly less than $1$.
\begin{lemma} \label{lem:bounded-contraction-fixedpoint}
For each $\pi = (\bm{d})_{\infty} \in \PiSR$ and $\bm{z} \ge \bm{0}$:
\[
  \bm{w}^{\infty}(\pi, \bm{z}) > -\bm{\infty}
  \quad \Leftrightarrow \quad
  \rho(\bm{B}^{\bm{d}}) < 1.
\]
\end{lemma}

\cref{lem:bounded-contraction-fixedpoint} uses the transience property to show that the Perron vector (with the maximum absolute eigenvalue) $\bm{f}$ of $\bm{B}^{\bm{d}}$ satisfies that $\bm{f}\tr \bm{b}^{\bm{d}}> 0$. Therefore, $\rho(\bm{B}^{\bm{d}}) < 1$ is necessary for the series in~\eqref{eq:exp-value-expression} to be bounded.

We state \cref{lem:termination-probability}, \cref{lem:exp-prob-monotone}, \cref{lem:pro-goal-state-zero}, and \cref{lem:exponential-eigen-non-zero} that are useful in the proof of \cref{lem:bounded-contraction-fixedpoint}. We use $\bm{p}^{\bm{d}}\in \Real_+^S$ to represent the probability of terminating from any state for each $\bm{d}\in \mathcal{D}$:
\[
 p^{\bm{d}}_s = \sum_{a\in \mathcal{A}} d_a(s) \cdot \bar{p}(s, a, e), \quad \forall s\in \states.
\]
The following lemma establishes a convenient representation of the termination probabilities. 
\begin{lemma} \label{lem:termination-probability}
Assume a policy $\pi = (\bm{d})_{\infty}\in \PiSR$. Then, the probability of terminating in $t \in \mathbb{N}, t > 0$ or fewer steps is
\[
  \sum_{k=0}^{t-1} \bm{\mu}\tr(\bm{P}^{\bm{d}})^k \bm{p}^{\bm{d}}
  \;=\;
  \bm{\mu}\tr (\bm{I} - (\bm{P}^{\bm{d}})^t) \bm{1},
\]
where $\bm{P}^{\bm{d}}$ is a $|\states| \times|\states|$ matrix, and each element $P^{\bm{d}}_{s,s'} =\sum_{a \in \mathcal{A}} d_a(s) \cdot  p(s,a,s') , \forall s,s' \in \states$.
\end{lemma}
\begin{proof}
We have by algebraic manipulation that
\[
\bm{p}^{\bm{d}} = (\bm{I} - \bm{P}^{\bm{d}}) \bm{1}. 
\]
The probability of terminating in step $t > 0$ exactly is
\[
\bm{\mu}\tr(\bm{P}^{\bm{d}})^{t-1} \bm{p}^{\bm{d}}.
\]
Using algebraic manipulation and recognizing a telescopic sum, we have that the probability of terminating in $k \le t$ steps is
\begin{equation*}
\sum_{k=0}^{t-1} \bm{\mu}\tr(\bm{P}^{\bm{d}})^k \bm{p}^{\bm{d}} 
=  \sum_{k=0}^{t-1} \bm{\mu}\tr(\bm{P}^{\bm{d}})^k (\bm{I} - \bm{P}^{\bm{d}}) \bm{1} 
=  \bm{\mu}\tr(\bm{I} - (\bm{P}^{\bm{d}})^t) \bm{1}.
\end{equation*}
\end{proof}

\begin{lemma} \label{lem:exp-prob-monotone}
  For any $\bm{d}\in \mathcal{D}$, the exponential transition matrix is monotone:
  \[
    \bm{x} \ge \bm{y}
    \quad \Rightarrow \quad
    \bm{B}^{\bm{d}} \bm{x} \ge \bm{B}^{\bm{d}} \bm{y},
    \qquad \forall \bm{x}, \bm{y} \in \Real^S.
  \]
\end{lemma}
\begin{proof}
The result follows immediately from the fact that $\bm{B}^{\bm{d}}$ is a non-nonnegative matrix.
\end{proof}

\begin{lemma} \label{lem:pro-goal-state-zero}
For each $t \in \mathbb{N}$ and each policy $\pi = (\bm{d})_{\infty} \in \PiSR$ and each $\bm{\eta}\in \probs{S}$:
\begin{equation}\label{eq:exponential-zero-equivalence}
  \bm{\eta}\tr (\bm{B}^{\bm{d}})^t \bm{b}^{\bm{d} }= 0
  \qquad \Leftrightarrow \qquad
  \bm{\eta}\tr (\bm{P}^{\bm{d}})^t \bm{p}^{\bm{d}} = 0.
\end{equation}
\end{lemma}
\begin{proof}
Algebraic manipulation from the definition in~\eqref{eq:exponential-definitions} shows that
 \begin{equation} \label{eq:goal-squeeze}
   \begin{array}{rcl}
   c_{\mathrm{l}} \cdot  \bm{p}^{\bm{d}}
   \; \le \;&  
   \bm{b}^{\bm{d}}
   &\; \le \;  
   c_{\mathrm{u}} \cdot  \bm{p}^{\bm{d}}, \\
   c_{\mathrm{l}} \cdot  \bm{P}^{\bm{d}} \bm{x}
   \; \le\;& 
   \bm{B}^{\bm{d}} \bm{x}
   &\;\le\; 
   c_{\mathrm{u}} \cdot  \bm{P}^{\bm{d}} \bm{x},
   \qquad
   \forall\bm{x} \in \Real^S , 
   \end{array}
\end{equation}
where
\begin{equation*}
    \begin{aligned}
     & c_{\mathrm{l}} := \min_{s,s'\in \bar{\mathcal{S}}, a\in \mathcal{A}} \exp{ - \beta \cdot  r(s, a, s')} , \\
    & c_{\mathrm{u}} := \max_{s,s'\in \bar{\mathcal{S}}, a\in \mathcal{A}} \exp{ - \beta \cdot  r(s, a, s') }.  
    \end{aligned}
\end{equation*}

Note that $\infty > c_{\mathrm{u}} > c_{\mathrm{l}} > 0$.

We now extend the inequalities in~\eqref{eq:goal-squeeze} to multiple time steps. Suppose that $\bm{y}_{\mathrm{l}} \le \bm{y} \le \bm{y}_{\mathrm{u}}$, then, for $t = \mathbb{N}$:
\begin{equation} \label{eq:goal-squeeze-t}
  c_{\mathrm{l}}^t \cdot (\bm{P}^{\bm{d}})^t \bm{y}_{\mathrm{l}}
  \; \le\;  (\bm{B}^{\bm{d}})^t \bm{y}
  \; \le\;
  c_{\mathrm{u}}^t \cdot (\bm{P}^{\bm{d}})^t \bm{y}_{\mathrm{u}}.
\end{equation}
For the left inequality in~\eqref{eq:goal-squeeze-t}, the induction proceeds as follows. The base case $t = 0$ holds immediately. For the inductive step, suppose that the left inequality in~\eqref{eq:goal-squeeze-t} property holds for $t \in \mathbb{N}$  then it also holds for $t+1$ for each $\bm{y}\in \Real^S$ as
\begin{equation*}
\begin{aligned}
         (\bm{B}^{\bm{d}})^{t+1} \bm{y} 
  \; =\; 
    \bm{B}^{\bm{d}}(\bm{B}^{\bm{d}})^t \bm{y} 
  \; \ge\;   c_{\mathrm{l}}^t\bm{B}^{\bm{d}}(\bm{P}^{\bm{d}})^t \bm{y}_{\mathrm{l}} 
  \; \ge\; 
 c_{\mathrm{l}}^{t+1}\bm{P}^{\bm{d}}(\bm{P}^{\bm{d}})^t \bm{y}_{\mathrm{l}} 
  \; =\; 
 c_{\mathrm{l}}^{t+1}(\bm{P}^{\bm{d}})^{t+1} \bm{y}_{\mathrm{l}}.
 \end{aligned}
\end{equation*}

Above, the first inequality follows from \cref{lem:exp-prob-monotone} and the inductive assumption, and the second inequality follows from~\eqref{eq:goal-squeeze-t} by setting $\bm{x} = \bm{P}^{\bm{d}}\bm{y}$. The right inequality in~\eqref{eq:goal-squeeze-t} follows analogously.

Exploiting the fact that $\bm{\eta} \ge \bm{0}$ and substituting $\bm{y} = \bm{b}^{\bm{d}}$, $\bm{y}_{\mathrm{l}} = c_{\mathrm{l}} \cdot \bm{p}^{\bm{d}}$, $\bm{y}_{\mathrm{u}} = c_{\mathrm{u}}\cdot \bm{p}^{\bm{d}}$ into~\eqref{eq:goal-squeeze-t} and using the bounds in~\eqref{eq:goal-squeeze}, we get that
\begin{equation*} 
\begin{aligned}
  0
  \;  \le\; 
  c_{\mathrm{l}}^{t+1} \cdot \bm{\eta}\tr (\bm{P}^{\bm{d}})^t \bm{p}^{\bm{d}} 
  \; \le\;  \bm{\eta}\tr (\bm{B}^{\bm{d}})^t \bm{b}^{\bm{d}} 
  \; \le\;
  c_{\mathrm{u}}^{t+1} \cdot \bm{\eta}\tr (\bm{P}^{\bm{d}})^t \bm{p}^{\bm{d}}, 
\end{aligned}
\end{equation*}
where the terms are non-negative because all constants, matrices, and vectors are non-negative. Therefore,
\[
 \bm{\eta}\tr (\bm{B}^{\bm{d}})^t \bm{b}^{\bm{d}} = 0
 \quad\Leftrightarrow \quad
  \bm{\eta}\tr (\bm{P}^{\bm{d}})^t \bm{p}^{\bm{d}} = 0.
\]
\end{proof}

\begin{lemma} \label{lem:exponential-eigen-non-zero}
  For each $\pi = (\bm{d})_{\infty} \in \PiSR$, there exists an $\bm{f}\in \Real^S$ such that
  \[
    \bm{f}\tr \bm{B}^{\bm{d}} = \rho(\bm{B}^{\bm{d}}) \cdot \bm{f}\tr,
    \quad \bm{f} \ge \bm{0},
    \quad \bm{f} \neq \bm{0},
   \quad \bm{f}\tr \bm{b}^{\bm{d}}  > 0.
  \]
\end{lemma}
\begin{proof}
Because $\bm{B}^{\bm{d}}$ is non-negative, there exists an $\bm{f} \in \Real^S$ from the Perron-Frobenius theorem~\cite[Theorem~8.3.1]{Horn2013}, such that
  \[
    \bm{f}\tr \bm{B}^{\bm{d}} = \rho(\bm{B}^{\bm{d}}) \cdot \bm{f}\tr,
    \qquad \bm{f} \ge \bm{0},
    \qquad \bm{f} \neq \bm{0}.
  \]
  Furthermore, because  $\bm{b}^{\bm{d}} \ge  \bm{0}$, we also have that
  \[
    \bm{f}\tr \bm{b}^{\bm{d}} \ge \bm{0}.
  \]
To prove the theorem, it remains to show that $\bm{f}\tr \bm{b}^{\bm{d}} \neq 0$, which we do by deriving a contradiction. Note that the $\bm{f}$ constructed from the Perron-Frobenius theorem is scale-independent. Therefore, without loss of generality, assume that $\bm{1}\tr \bm{f} = 1$ and suppose that $\bm{f}\tr \bm{b}^{\bm{d}} = 0$. Then:
\begin{align*}
\bm{f}\tr \bm{b}^{\bm{d}} &= 0 \\
   \bm{f}\tr (\bm{B}^{\bm{d}})^k \bm{b}^{\bm{d}} &= 0, \, \forall k \in \mathbb{N} &&  \text{[ from } \bm{f}\tr \bm{B}^{\bm{d}} = \rho(\bm{B}^{\bm{d}}) \cdot \bm{f}\tr \text{]} \\
   \bm{f}\tr (\bm{P}^{\bm{d}})^k \bm{p}^{\bm{d}}&= 0, \, \forall k \in \mathbb{N} &&  \text{[ from \cref{lem:pro-goal-state-zero} ]} \\
   \sum_{k=0}^{t-1}  \bm{f}\tr (\bm{P}^{\bm{d}})^k \bm{p}^{\bm{d}} &= 0, \, \forall t \in \mathbb{N}, t > 0 && \text{[ by summing elements ]}   \\
    \bm{f}\tr (\bm{I} - (\bm{P}^{\bm{d}})^t)\bm{1} &= 0, \, \forall t \in \mathbb{N} && \text{[ from \cref{lem:termination-probability} ]} \\
   \lim_{t\to \infty} \bm{f}\tr (\bm{I} - (\bm{P}^{\bm{d}})^t)\bm{1} &= 0,  &&  \text{[ limit ]}\\
    \bm{f}\tr \bm{1} &= 0,  && \text{[ from \cref{lem:transient-spectral-radius} ]} 
\end{align*}
which is a contradiction with $\bm{1}\tr \bm{f} \neq  0$. The last step in the derivation follows from $\rho(\bm{P}^{\bm{d}}) < 1$ and therefore $\lim_{t\to \infty} (\bm{P}^{\bm{d}})^t = \bm{0}$~\citep[Theorem~4.5]{Kallenberg2021markov}.
\end{proof}

\begin{proof}[Proof of \cref{lem:bounded-contraction-fixedpoint}]
From \cref{lem:exponential-eigen-non-zero}, there exists an $\bm{f} \in \Real_+^{S}$ that $\bm{f}\tr \bm{B}^{\bm{d}} = \rho(\bm{B}^{\bm{d}}) \cdot \bm{f}\tr$ and $\bm{f} \ge \bm{0}, \bm{f} \neq \bm{0}$. Then from~\eqref{eq:exp-value-expression}:
\begin{equation*}
\begin{aligned}
    -\infty &<  \bm{\eta}\tr  \bm{w}^t(\pi, \bm{z}) \\
    &=  - \bm{\eta}\tr  (\bm{B}^{\bm{d}})^t \bm{z} - \bm{\eta}\tr \sum_{k=0}^{t-1}  (\bm{B}^{\bm{d}})^k \bm{b}^{\bm{d}} \\
    &\le  - \sum_{k=0}^{t-1}  \rho(\bm{B}^{\bm{d}})^k \bm{\eta}\tr \bm{b}^{\bm{d}}.
\end{aligned}
\end{equation*}
The second inequality follows because $\bm{z} \ge \bm{0}$ and $\bm{B}^{\bm{d}}$ is non-negative. Since $\bm{\eta}\tr \bm{b} > 0$ from \cref{lem:exponential-eigen-non-zero}, we can cancel it from the inequality getting that
\[
  \sum_{k=0}^{t-1}  \rho(\bm{B}^{\bm{d}})^k  < \infty.
\]
Then $\rho(\bm{B}^{\bm{d}}) < 1$ because $\rho(\bm{B}^{\bm{d}})\ge 0$ and the geometric series above is bounded.
\end{proof}

The limitation of \cref{lem:bounded-contraction-fixedpoint} is that it only applies to stationary policies. The lemma does not preclude the possibility that all stationary policies have unbounded returns, but a Markov policy with a bounded return exists. We construct an upper bound on $\bm{w}^{t,\star}$ that decreases monotonically in $t$ and converges to show this is impossible. The proof then concludes by squeezing $\bm{w}^{t,\star}$ between the lower and the upper bound with the same limits. This technique allows us to relax the limiting assumptions from prior work~\cite{patek2001terminating,de2020risk}. Finally, our results imply that an optimal stationary policy exists whenever the planning horizon $T$ is sufficiently large. Because the set $\PiSD$ is finite, one policy must be optimal for a sufficiently large $T$. This property suggests behavior similar to \emph{turnpikes} in discounted MDPs~\cite{puterman205markov}.

\subsection{Algorithms}
We now briefly describe the algorithms we use to compute the optimal ERM-TRC policies. Surprisingly, the main algorithms for discounted MDPs, including value iteration, policy iteration, and linear programming, can be adapted to this risk-averse setting with only minor modifications.

\emph{Value iteration} is the most direct method for computing the optimal value function~\cite{puterman205markov}. The value iteration computes a sequence of $\bm{w}^k, k = 0, \dots $ such that
\[
 \bm{w}^{k+1} = L\opt \bm{w}^k, \quad \bm{w}^0 = \bm{0}. 
\]
The initialization of $\bm{w}^0 = \bm{0}$ is essential and guarantees convergence directly from the monotonicity argument used to prove \cref{thm:erm-main-convergence}.

\emph{Policy iteration} (PI) starts by initializing with a stationary policy $\pi_0 = (\bm{d}^0)_{\infty} \in \PiSD$. Then, for each iteration $k = 0, \dots $, PI alternates between the policy evaluation step and the policy improvement step:
\begin{equation*}
\bm{w}^k = - (\bm{I} - \bm{B}^{\bm{d}^k})^{-1} \bm{b}^{\bm{d}^k}, \;
  \bm{d}^{k+1} \in \argmax_{\bm{d}\in \mathcal{D}} \bm{B}^{\bm{d}} \bm{w}^k - \bm{b}^{\bm{d}}. 
\end{equation*}
PI converges because it monotonically improves the value functions when initialized with a policy $\bm{d}^0$ that has a bounded return~\cite{patek2001terminating}. However, we lack a practical approach to finding such an initial policy.

Finally, \emph{linear programming} is a fast and convenient method for computing optimal exponential value functions:
\begin{equation} \label{eq:erm-lp}
  \min \left\{  \bm{1}\tr \bm{w} \mid \bm{w}\in \Real^{S}, \bm{w} \ge  - \bm{b}^a  + \bm{B}^{a} \bm{w} ,    \,\forall a\in \actions \right\}.
\end{equation}
Here, $\bm{B}_{s,\cdot }^{a}  =(B_{s,s_1 }^{a}, \cdots, B_{s,s_{S}}^{a})$, $B_{s,s'}^{a}$ and $b_{s}^a$ are constructed as in~\eqref{eq:exponential-definitions}. We use the shorthand $\bm{B}^a = \bm{B}^{\bm{d}}$ and $\bm{b}^a = \bm{b}^{\bm{d}}$ where  $d_{a'}(s) = 1$ if $a = a'$ for each $s\in \mathcal{S}, a'\in \mathcal{A}$.

It is important to note that the value functions, as well as the coefficients of $\bm{B}^{\bm{d}}$ may be irrational. It is, therefore, essential to study the sensitivity of the algorithms to errors in the input. However, this question is beyond the scope of the present paper, and we leave it for future work. 

\section{Solving EVaR Total Reward Criterion}
\label{sec:ssp-with-evar}

This section shows that the EVaR-TRC objective can be reduced to a sequence of ERM-TRC problems, similarly to the discounted case~\cite{hau2023entropic}. As a result, an optimal stationary EVaR-TRC policy exists and can be computed using the methods described in \cref{sec:ssp-with-erm}.

Formally, we aim to compute a policy that maximizes the EVaR of the random return at some given fixed risk level $\alpha \in (0,1)$ defined as
\begin{equation} \label{eq:risk_object}
\sup_{\pi \in \PiHR} \liminf_{t\to \infty}  
\evaro_{\alpha}^{\pi, \bm{\mu}}
\left[ \sum_{k=0}^{t-1} r(\tilde{s}_k,\tilde{a}_k,\tilde{s}_{k+1}) \right].
\end{equation}
In contrast with \cite{ahmadi2021constrained}, the objective in~\eqref{eq:risk_object} optimizes EVaR rather than Nested EVaR.

\subsection{Reduction to ERM-TRC}
To solve~\eqref{eq:risk_object}, we exploit that EVaR can be defined in terms of ERM as shown in~\eqref{eq:evar-def-app}. To that end, define a function $h_t\colon \PiHR \times \Real \rightarrow \bar{\Real}$ for $t \in \mathbb{N} $ as
\begin{equation} \label{eq:h}
h_t(\pi, \beta) \;:=\;   g_t(\pi, \beta) + \beta^{-1} \log(\alpha),
\end{equation}
where $g_t$ is the ERM value of the policy defined in~\eqref{eq:g-definitions}. Also, $h_t\opt$, $h_{\infty}$, $h_{\infty}\opt$ are defined analogously in terms of $g_t\opt$, $g_{\infty}$, and $g_{\infty}\opt$ respectively. The functions $h$ are useful, because by~\eqref{eq:evar-def-app}:
\begin{equation} \label{eq:h-function-utility}
\evaro_{\alpha}^{\pi, \bm{\mu}}
\left[ \sum_{k=0}^{t-1} r(\tilde{s}_k,\tilde{a}_k,\tilde{s}_{k+1}) \right]
\; =\; 
\sup_{\beta > 0} h_t(\pi, \beta),
\end{equation}
for each $\pi\in \PiHR$ and $t \in \mathbb{N} $. However, note that the limit in the definition of $\sup_{\beta > 0} h_{\infty}\opt(\beta)$ is inside the supremum, unlike in the objective in~\eqref{eq:risk_object}.

There are two challenges with solving~\eqref{eq:risk_object} by reducing it to~\eqref{eq:h-function-utility}. First, the supremum in the definition of EVaR in~\eqref{eq:evar-def-app} may not be attained, as mentioned previously. Second, the functions $g_t\opt$ and $h_t\opt$ may not converge \emph{uniformly} to $g_{\infty}\opt$ and $h_{\infty}\opt$. Note that \cref{thm:erm-main-convergence} only shows \emph{pointwise} convergence when the functions are bounded.

To circumvent the challenges described above, we replace the supremum in~\eqref{eq:h-function-utility} with a maximum over a \emph{finite} set $\mathcal{B}(\beta_0, \delta)$ of discretized $\beta$ values:
\begin{subequations} \label{eq:b-set-defs}
\begin{equation} \label{eq:b-set-definition}
  \mathcal{B}(\beta_0, \delta) \;:=\;  \left\{ \beta_0, \beta_1, \dots , \beta_K \right\},
\end{equation}
where $\delta > 0$,  $0 < \beta_0 < \beta_1 < \dots  < \beta_K$, and
\begin{equation}
\label{eq:beta-construction}
\beta_{k+1} \; :=\;  \frac{\beta_k \log \frac{1}{\alpha}}{\log\frac{1}{ \alpha } -\beta_k \delta} ,
\qquad
\beta_K \; \ge\;  \frac{\log \frac{1}{\alpha}}{\delta},
\end{equation}
\end{subequations}
for an appropriately chosen value $K$ for each $\beta_0$ and $\delta$. We assume that the denominator in the expression for $\beta_{k+1}$ in \cref{eq:beta-construction} is positive; otherwise $\beta_{k+1} = \infty$ and $\beta_k$ is sufficiently large.

The construction in~\eqref{eq:b-set-defs} resembles equations (19) and (20) in~\cite{hau2023entropic} but differs in the choice of $\beta_0$ because Hoeffding's lemma does not readily bound the TRC criterion.

The following auxiliary  \cref{lem:case1}, \cref{lem:case2}, \cref{lem:case3} and \cref{lem:beta-b-approximation} will be used to bound the approximation error of $h$.
\begin{lemma} \label{lem:case1}
Suppose that $f\colon \Real_{+} \to \bar{\Real}$ is non-increasing and $0 < \beta_0$. Then: 
\begin{equation*}
\begin{aligned}
 \sup_{\beta \in [0,\beta_0)} (f(\beta) + \beta^{-1} \log \alpha)  -
(f(\beta_0) + \beta_0^{-1} \log \alpha)  
& \;\le\;  f(0) - f(\beta_0) .
\end{aligned}
\end{equation*}

\end{lemma}

\begin{proof}
  Follows immediately from the fact that $f(\beta) \le f(0), \forall \beta \ge 0$ and $\beta^{-1} \log \alpha < 0, \forall \beta > 0$. 
\end{proof}

\begin{lemma} \label{lem:case2}
Suppose that $f\colon \Real_{+} \to \bar{\Real}$ is non-increasing and $0 < \beta_k < \beta_{k+1}$. Then: 
\begin{equation*}
\begin{aligned}
\sup_{\beta \in [\beta_k,\beta_{k+1})} (f(\beta) + \beta^{-1} \log \alpha) -
(f(\beta_k) + \beta_k^{-1} \log \alpha) 
& \;\le\;  (\beta_{k+1}^{-1} - \beta_k^{-1}) \cdot \log \alpha. 
\end{aligned}
\end{equation*}
\end{lemma}

\begin{proof}
  The proof uses the same technique as Lemma D.5 from~\cite{hau2023entropic}. We restate the proof here for completeness.

  Using the fact that $f$ is non-increasing and $\beta \mapsto \beta^{-1} \log \alpha$ is increasing we get that:
\begin{gather*}
  \sup_{\beta\in [\beta_k,\beta_{k+1})}  \left(  f(\beta) + \beta^{-1}\cdot \log (\alpha) \right) - 
      \left(  f(\beta_k) + \beta_k^{-1}\cdot \log \alpha \right)  \\
    \le \sup_{\beta\in [\beta_k,\beta_{k+1})}  \left(  f(\beta) + \beta_{k+1}^{-1}\cdot \log \alpha \right) - \left( f(\beta_k) + \beta_k^{-1}\cdot \log \alpha \right) \\
    \le \sup_{\beta\in [\beta_k,\beta_{k+1})}  \left(  f(\beta)  - f(\beta_k) \right) +  \left(\beta_{k+1}^{-1}\cdot \log (\alpha)  - \beta_k^{-1}\cdot \log \alpha \right) \\
\le  \beta_{k+1}^{-1}\cdot \log \alpha  - \beta_k^{-1}\cdot \log \alpha .
\end{gather*}
\end{proof}

\begin{lemma} \label{lem:case3}
Suppose that $f\colon \Real_{+} \to \bar{\Real}$ is non-increasing and $0 < \beta_K$. Then:
\begin{equation*}
    \begin{aligned}
       & \sup_{\beta\in [\beta_K, \infty)} (f(\beta) + \beta^{-1} \log  \alpha ) -
  (f(\beta_K) + \beta_K^{-1} \log \alpha) \; \le\;
\frac{-\log (\alpha)}{\beta_K}. 
    \end{aligned}
\end{equation*}
\end{lemma}

\begin{proof}
The proof uses the same technique as Lemma D.6 from~\cite{hau2023entropic}. We restate the proof here for completeness.
  
Because $f$ non-increasing and  $\beta^{-1}\cdot \log \alpha \le 0, \forall \beta > 0$ we get that:
\begin{align*}
 \sup_{\beta\in [\beta_K, \infty)} (f(\beta) + \beta^{-1} \log  \alpha ) -
  (f(\beta_K) + \beta_K^{-1} \log \alpha) 
\le \sup_{\beta\in [\beta_K, \infty)} (f(\beta)  ) -
  (f(\beta_K) + \beta_K^{-1} \log \alpha) 
\le
\frac{-\log (\alpha)}{\beta_K}.
\end{align*}
\end{proof}

\begin{lemma} \label{lem:beta-b-approximation}
Suppose that $f\colon \Real_{+} \to \bar{\Real}$ is non-increasing and $\mathcal{B}(\beta, \delta)$ defined in~\eqref{eq:b-set-defs}. Then for each $\beta_0 > 0$ and $\delta > 0$:
\begin{equation*}
\begin{aligned}
\max_{\beta\in \mathcal{B}(\beta_0, \delta)}  f(\beta) + \beta^{-1} \log \alpha 
& \; \le\; 
\sup_{\beta > 0 } f(\beta) + \beta^{-1} \log  \alpha \\
&  \; \le\; 
\max_{\beta\in \mathcal{B}(\beta_0, \delta)}  f(\beta) + \beta^{-1} \log \alpha +
\max \left\{ f(0) - f(\beta_0) , \delta \right\} .
\end{aligned}
\end{equation*}
\end{lemma}
\begin{proof}
The result follows by algebraic manipulation from \cref{lem:case1,lem:case2,lem:case3}.
\end{proof}

The following proposition upper-bounds the value of $K$; see~\citep[theorem 4.3]{hau2023entropic} for a proof that $K$ is polynomial in $\delta$. 
\begin{proposition} \label{prop:choice-K}
  Assume a given $\beta_0 > 0$ and $\delta \in (0,1)$ such that $\beta_0 \delta < \log \frac{1}{\alpha}$. Then, to satisfy the condition in~\eqref{eq:beta-construction}, it is sufficient to choose $K$ as
  \begin{equation} \label{eq:K-inequality}
   K := \frac{\log  z}{\log (1-z)}, \quad\text{where}\quad z := \frac{\beta_0 \delta }{\log \frac{1}{\alpha}}. 
\end{equation}
\end{proposition}
\begin{proof}[Proof of \cref{prop:choice-K}]
  First, the iterates $\beta_k, k\ge 0$ can be bounded as
  \begin{align*}
    \beta_{k+1}
    &= \frac{\beta_k \log \frac{1}{\alpha}}{\log\frac{1}{ \alpha } -\beta_k \delta}
              \ge \beta_k \frac{\log \frac{1}{\alpha}}{\log \frac{1}{\alpha} - \beta_0 \delta} \\
    &=  \beta_0 \left(\frac{\log \frac{1}{\alpha}}{\log \frac{1}{\alpha} - \beta_0 \delta} \right)^{k+1}.
\end{align*}
Using the lower bound on $\beta_k$, to show that
\begin{equation*}
\beta_K \ge  \frac{\log \frac{1}{\alpha}}{\delta},
\end{equation*}
it is sufficient to show that
\[
\beta_0 \left(\frac{\log \frac{1}{\alpha}}{\log \frac{1}{\alpha} - \beta_0 \delta} \right)^K \ge 
\frac{\log \frac{1}{\alpha}}{\delta}.
\]
Using the variable $z$, the sufficient condition translates to 
\begin{align*}
  \left( \frac{1}{1-z} \right)^K &\ge \frac{1}{z} \\
  K &\ge \frac{\log  z}{\log (1-z)},
\end{align*}
since $z \in (0,1)$ by the assumption $\beta_0 \delta < \log \frac{1}{\alpha}$.
\end{proof}

The following theorem shows that one can obtain an optimal ERM policy for an appropriately chosen $\beta$ that approximates an optimal EVaR policy arbitrarily closely. 
\begin{theorem} \label{thm:optimal-evar-erm}
For any $\delta > 0$, let
\[
  (\pi\opt,\beta\opt)  \in
  \argmax_{(\pi,\beta)\in \PiSD \times \mathcal{B}(\beta_0, \delta)}
  h_{\infty}(\pi,\beta),
\]
where $\beta_0 > 0$ is chosen such that $g_{\infty}\opt(0) \le g_{\infty}\opt(\beta_0) - \delta$. Then the limits below exist and satisfy:
\begin{equation} \label{eq:evar-guarantee}
\lim_{t\to \infty}  
\evaro_{\alpha}^{\pi\opt, \bm{\mu}}
\left[ \sum_{k=0}^{t-1} r(\tilde{s}_k,\tilde{a}_k,\tilde{s}_{k+1}) \right]
  \ge
  \sup_{\pi\in \PiHR} \lim_{t\to \infty} \sup_{\beta>0} h_t(\pi,\beta)
  -\delta.
\end{equation}
\end{theorem}
Note that the right-hand side in~\eqref{eq:evar-guarantee} is the $\delta$-optimal objective in~\eqref{eq:risk_object}. 
\begin{proof}[Proof of \cref{thm:optimal-evar-erm}]
From \cref{lem:erm-convergence-0}, we can choose a $\beta_0 > 0$ be such that $g_{\infty}\opt(0) - g\opt_{\infty} (\beta_0) \le \delta$.

Then we can upper bound the objective as:
\begin{align*}
\sup_{\pi\in \PiHR} \lim_{t\to \infty }  \sup_{\beta > 0} h_t(\pi ,\beta )
&\stackrel{\text{(a)}}{\le} \lim_{t\to \infty }  \sup_{\beta > 0}\sup_{\pi\in \PiHR}  h_t(\pi ,\beta )  \\
& \stackrel{\text{(b)}}{=} 
  \lim_{t\to \infty }  \sup_{\beta > 0}  h_t\opt(\beta ) \\
&\stackrel{\text{(c)}}{\le} 
  \lim_{t\to \infty }  \max_{\beta \in \mathcal{B}(\beta_0, \delta)}  h_t\opt(\beta ) + \delta \\
& \stackrel{\text{(d)}}{\le} 
    \max_{\beta \in \mathcal{B}(\beta_0, \delta)}  \lim_{t\to \infty } h_t\opt(\beta ) + \delta\\
&\le \max_{\beta \in \mathcal{B}(\beta_0, \delta)}  h_{\infty }\opt(\beta ) + \delta.
\end{align*}
Here, (a) follows because the limit of an upper bound on a sequence is an upper bound on the limit, (b) follows from the definition of $h_t\opt $, (c) from \cref{lem:beta-b-approximation}, (d) from the continuity of $\max$ over finite sets.

Let $\pi\opt \in \PiSD, \beta\opt \in \mathcal{B}(\beta_0, \delta)$ be the maximizers that attain the objective above:
\[
  h_{\infty }(\pi\opt, \beta\opt ) = \max_{\beta \in \mathcal{B}(\beta_0, \delta)}  h_{\infty }\opt(\beta ).
\]
Then, continuing the upper bound on the objective:
\begin{align*}
 \max_{\beta \in \mathcal{B}(\beta_0, \delta)}  h_{\infty }\opt(\beta ) + \delta 
  &= h_{\infty }(\pi\opt, \beta\opt )  + \delta  \\
  &= \lim_{t\to \infty } h_t(\pi\opt, \beta\opt )  + \delta  \\
  &\le  \lim_{t\to \infty } \sup_{\beta > 0} h_t(\pi\opt, \beta)  + \delta  \\
 & \le \lim_{t\to \infty }  \evaro_{\alpha}^{\pi\opt, \bm{\mu}} \left[ \sum_{k=0}^{t-1} r(\tilde{s}_k,\tilde{a}_k,\tilde{s}_{k+1}) \right] + \delta ,
\end{align*}
which shows that $\pi\opt$ is $\delta$-suboptimal since $\displaystyle \lim_{t\to \infty }  \evaro_{\alpha}^{\pi\opt, \bm{\mu}} \left[ \sum_{k=0}^{t-1} r(\tilde{s}_k,\tilde{a}_k,\tilde{s}_{k+1}) \right]$ is a lower bound on the objective. 
\end{proof}

The first implication of \cref{thm:optimal-evar-erm} is that there exists an optimal stationary deterministic policy. 
\begin{corollary} \label{cor:evar-stationary-policy}
  There exists an optimal stationary deterministic policy $\pi\opt \in \PiSD$ that attains the supremum in~\eqref{eq:risk_object}.
\end{corollary}
\begin{proof}[Proof of \cref{cor:evar-stationary-policy}]
Suppose that there exists a $\hat{\pi}\in \PiHR \setminus \PiSD$ such that it attains an objective value greater than the best $\pi\in \PiSD$ by at least $\epsilon > 0$. Note that a best $\pi\in \PiSD$ exists because $\PiSD$ is finite. Then, one can derive a contradiction by choosing a $\pi\in \PiSD$ that is at most $\epsilon / 2$ suboptimal.   
\end{proof}

The second implication of \cref{thm:optimal-evar-erm} is that it suggests an algorithm for computing the optimal, or near-optimal, stationary policy. We summarize it in \cref{sec:evar-algorithms}.

\subsection{Algorithms}
\label{sec:evar-algorithms}

We now propose a simple algorithm for computing a $\delta$-optimal EVaR policy described in \cref{alg:evar-algo}. The algorithm reduces finding optimal EVaR-TRC policies to solving a sequence of ERM-TRC
problems in~\eqref{eq:sup-erm}. As \cref{thm:optimal-evar-erm} shows, there exists a $\delta$-optimal policy such that it is ERM-TRC optimal for some $\beta \in \mathcal{B}(\beta_0, \delta)$. It is, therefore, sufficient to compute an ERM-TRC  optimal policy for one of those $\beta$ values. 
\begin{algorithm}
\textbf{Input}:MDP and desired precision $\delta > 0$ \\
\textbf{Output}:$\delta$-optimal policy $\pi\opt \in \PiSD$
\begin{algorithmic}[1]
\WHILE{$g_{\infty}\opt(0) - g_{\infty}\opt(\beta_0) > \delta$}
\STATE $\beta_0 \gets \beta_0 / 2$ 
\ENDWHILE
\STATE Construct $\mathcal{B}(\beta_0, \delta)$ as described in~\eqref{eq:b-set-definition} \;
\STATE Compute $\pi\opt  \in
\argmax_{\pi\in \PiSD} \max_{\beta\in \mathcal{B}(\beta_0, \delta)} h_{\infty}(\pi,\beta)$ by solving a linear program in~\eqref{eq:erm-lp} \;
\caption{Simple EVaR algorithm}
\label{alg:evar-algo}
\end{algorithmic}
\end{algorithm}

The analysis above shows that \cref{alg:evar-algo} is correct.
\begin{corollary} \label{prop:evar-algo-ok}
\Cref{alg:evar-algo} computes the $\delta$-optimal policy $\pi\opt \in \PiSD$ that satifies the condition~\eqref{eq:evar-guarantee}.
\end{corollary}
\Cref{prop:evar-algo-ok} follows directly from \cref{thm:optimal-evar-erm} and from the existence of a sufficiently small $\beta_0$ from the continuity of $g_{\infty}\opt(\beta)$ for positive $\beta$ around $0$.

\Cref{alg:evar-algo} prioritizes simplicity over computational complexity and could be accelerated significantly. Evaluating each $h_{\infty}\opt(\beta)$ requires computing an optimal ERM-TRC solution, which involves solving a linear program. One could reduce the number of evaluations of $h_{\infty}\opt$ needed by employing a branch-and-bound strategy that takes advantage of the monotonicity of $g_{\infty}\opt$.

An additional advantage of \cref{alg:evar-algo} is that the overhead of computing optimal solutions for multiple risk levels $\alpha$ can be small if one selects an appropriate set $\mathcal{B}$.

\section{Numerical Evaluation}
\label{sec:numerical-eval}

In this section, we illustrate our algorithms and formulations on tabular MDPs that include positive and negative rewards. 

\begin{figure}
\centering
\includegraphics[width=0.65\linewidth]{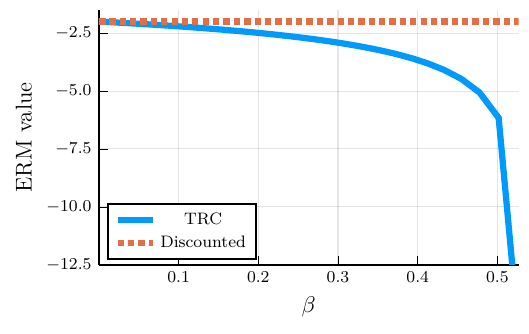} 
\caption{ERM values with TRC and discounted criteria.}
\label{fig:unbounded-erm}
\end{figure}

The ERM returns for the discounted and transient MDPs in \cref{fig:discounted-transient-mdp} with parameters $r = -0.2,\, \gamma = 0.9,\, \epsilon = 0.9$ are shown in \cref{fig:unbounded-erm}. The figure shows that, as expected, the returns are identical in the risk-neutral objective (when $\beta = 0$). However, for $\beta > 0$, the discounted and TRC returns differ significantly. The discounted return is unaffected by $\beta$ while the ERM-TRC return decreases with an increasing $\beta$. Please see \cite[appendix B]{su2024stationary} for more details.
 
To evaluate the effect of risk-aversion on the structure of the optimal policy, we use the \emph{gambler's ruin} problem~\cite{hau2023entropic,bauerle2011markov}. In this problem, a gambler starts with a given amount of capital and seeks to increase it up to a cap $K$. In each turn, the gambler decides how much capital to bet. The bet doubles or is lost with a probability $q$ and $1-q$, respectively. The gambler can quit and retain their current wealth; the game also ends when the gambler goes broke or reaches the cap $K$. The reward equals the final capital, except it is -1 when the gambler is broke. In the formulation, we use $q = 0.68$, and a cap is $K = 7$. The state space is $\bar{\states} = \{0,1,\cdots,7,e \}$ and the action space is $\actions = \{0,1,\cdots,6 \}$. Action $0$ represents quitting the game, and the other actions represent the bet size. The initial state is chosen randomly according to a uniform distribution over states $1, \dots, 7$. In state $0$, the gambler has capital $0$, takes the only available action $0$, receives the reward $-1$, and transits to the sink state $e$. That is: $\bar{p}(0,0,e)=1 $ and $
\bar{r}(0,0,e)=-1$. If state $7$, the gambler has capital 7, reaches the target wealth level, takes the only available action $0$, receives the reward $7$, and transits to the sink state $e$. That is: $ \bar{p}(7,0,e)=1$ and $ \bar{r}(7,0,e)=7$. For the sink state $e$, the only available action is $0$, and the transition probabilities and rewards are
$\bar{p}(e,0,e)=1 $ and $
  \bar{r}(e,0,e)=0$. For other states $s\in \left\{ 1, \dots , 6 \right\}$, the gambler's capital is $s$, can take any action $a \in \{ 1,\cdots,s \}$ to continue the game or an action $0$ to quit the game.
If the game is won, the gambler's bet doubles. If the game is lost, the gambler's bet is lost. That is, the transition probabilities and rewards for $a\in \left\{ 1, \dots , s \right\}$ are
\[
  \bar{p}(s,a,s') =
  \begin{cases}
      q &\text{if } s' = \min \left\{  s + a, 7 \right\}, \\
    1-q &\text{if } s' = s - a, \\
      0 &\text{otherwise},
  \end{cases}
  \qquad
  \bar{r}(s,a,s') = 0,
\]
for each $s' \in  \left\{ 0, \dots, 7, e \right\}$ where $q \in [0,1]$ is the win probability. If $s\in \left\{ 1, \dots , 6 \right\}$ and the action is $0$, the gambler quits the game and collects the reward. That is:
$ \bar{p}(s,0,e) = 1, $ and $\bar{r}(s,0,e)=s $. The algorithm was implemented in Julia 1.10, and is available at \url{https://github.com/suxh2019/ERMLP}. Please see \cite[appendix F]{su2024stationary} for more details. 

\begin{figure}
\centering
\includegraphics[width=0.65\linewidth]{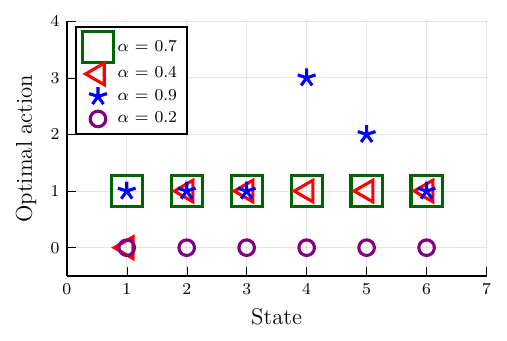}
\caption{The optimal EVaR-TRC policies.}
\label{fig:opt-policy}
\end{figure}

\Cref{fig:opt-policy} shows optimal policies for four different EVaR risk levels $\alpha$ computed by  \cref{alg:evar-algo}. The state represents the amount of capital the gambler holds. The optimal action indicates the amount of capital invested. The action zero means quitting the game. Note that there is only one action when the capital is $0$ and $7$ for all policies so that action is neglected in \Cref{fig:opt-policy}. Since the optimal policy is stationary, we can interpret and analyze it effectively. The policies become notably less risk-averse as $\alpha$ increases. For example, when $\alpha = 0.2$, the gambler is very risk-averse and always quits with the current capital. When $\alpha = 0.4$, the gambler invests $1$ when capital is greater than $1$ and quits otherwise to avoid losing it all. When $\alpha = 0.9$, the gambler makes bigger bets, increasing the probability of reaching the cap and losing all capital. 
 
\begin{figure}
\centering
\includegraphics[width=0.65\linewidth]{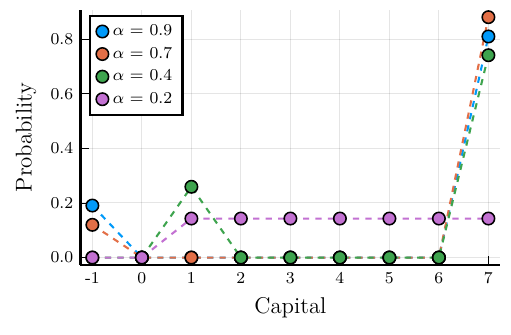}
\caption{Distribution of the final capital for EVaR optimal policies. }
\label{fig:capital-p0.65}
\end{figure}

To understand the impact of risk-aversion on the distribution of returns, we simulate the resulting policies over 7,000 episodes and show the distribution of capitals in \cref{fig:capital-p0.65}.
When $\alpha = 0.2$, the return follows a uniform distribution on [1, 7]. When $\alpha = 0.4$, the returns are $1$ and $7$.  When $\alpha = 0.7$ or $0.9$, the returns are $-1$ and $7$. Overall, the figure shows that for lower values of $\alpha$, the gambler sacrifices some probability of reaching the cap in exchange for a lower probability of losing all their capital.

\section{Conclusion and Future Work}

We analyze transient MDPs with two risk measures: ERM and EVaR. We establish the existence of stationary deterministic optimal policies without any assumptions on the sign of the rewards, a significant departure from past work. Our results also provide algorithms based on value iteration, policy iteration, and linear programming for computing optimal policies.

Future directions include extensions to infinite-state TRC problems, risk-averse MDPs with average rewards, and partial-state observations.

 \chapter{Risk-Averse Total-Reward Reinforcement Learning}
\label{chp:model-free}

This chapter focuses on mitigating the effects of aleatoric uncertainty and risk due to the inherent stochasticity of the environment, without knowledge of the transition functions. The proposed Q-learning algorithms in this chapter belong to model-free RL approaches.

Risk-averse total-reward Markov Decision Processes (MDPs) offer a promising framework for modeling and solving undiscounted infinite-horizon objectives. Existing model-based algorithms for risk measures like the entropic risk measure (ERM) and entropic value-at-risk (EVaR) are effective in small problems, but require full access to transition probabilities. We propose a Q-learning algorithm to compute the optimal stationary policy for total-reward ERM and EVaR objectives with strong convergence and performance guarantees. The algorithm and its optimality are made possible by ERM's dynamic consistency and elicitability. Our numerical results on tabular domains demonstrate quick and reliable convergence of the proposed Q-learning algorithm to the optimal risk-averse value function.

This work has been accepted by the 39th Annual Conference on Neural Information Processing Systems (NeurIPS 2025). Xihong Su,  Jia Lin Hau, Gersi Doko, Kishan Panaganti, Marek Petrik. Risk-Averse Total-Reward Reinforcement Learning (accepted). This work relates to Jia Lin Hau's work~\cite{hau2024q}, and he helped me better understand his paper.
Jia and Gersi wrote the basic code to implement ERM and EVaR Q-learning algorithms. Kishan attended some meetings. Xihong wrote most of the code for this work and the draft. Marek and Xihong edited and finalized the paper. 

\section{Introduction}

Risk-averse reinforcement learning~(RL) is an essential framework for practical and high-stakes applications, such as self-driving, robotic surgery, healthcare, and finance~\cite{greenberg2022efficient,fei2021exponential,urpi2021risk,hau2024q,suevar,zhang2021mean,hau2023entropic, Kastner2023, Marthe2023a, lam2022risk, li2022quantile, bauerle2022markov,pan2019risk,mazouchi2023risk}. A risk-averse policy prefers actions with more certainty even if it means a lower expected return~\cite{shen2014risk, mazouchi2023risk}. The goal of Risk-averse RL is to compute risk-averse policies. To this end, RARL specifies the objective using monetary risk measures, such as value-at-risk~(VaR), conditional-value-at-risk~(CVaR), entropic risk measure~(ERM), or entropic value at risk~(EVaR). These risk measures penalize the variability of returns interpretably and yield policies with stronger guarantees on the probability of catastrophic losses~\cite{Follmer2016stochastic}.

A major challenge in deriving practical RL algorithms is that the model of the environment is often unknown. This challenge is even more salient in risk-averse RL because computing the risk of random return involves evaluating the full distribution of the return rather than just its expectation~\cite{hau2024q}. 
% \todo{do you mean methods of rarl?} yes, it is harder to derive risk-averse Q-learning algorithm compared to the standard Q-leanring algorithm
Traditional definitions of risk measures assume a known  discounted~\cite{hau2023entropic,Kastner2023, Marthe2023a, lam2022risk, li2022quantile, bauerle2022markov, Hau2023a}  or transient~\cite{su2025risk} MDP model and use learning to approximate a global value or policy function~\cite{moerland2023model}. Recent works have presented model-free methods to find optimal CVaR policies~\cite{keramati2020being,lim2022distributional} or optimal VaR policies~\cite{hau2024q}, where the optimal policy may be Markovian or history-dependent. Another recent work derives a risk-sensitive Q-learning algorithm, which applies a nonlinear utility function to the temporal difference~(TD) residual~\cite{shen2014risk}.

In RL domains formulated as Markov Decision Processes (MDPs), future rewards are handled differently based on whether the model is discounted or undiscounted. Most RL formulations assume discounted infinite-horizon objectives in which future rewards are assigned a lower value than present rewards. Discounting in financial domains is typically justified by interest rates and inflation, which make earlier rewards inherently more valuable~\cite{puterman205markov,Huang2024}. However, many RL tasks, such as robotics or games, have no natural justification for discounting. Instead, the domains have absorbing terminal states, which may represent desirable or undesirable outcomes~\cite{mahadevan1996average,andrew2018reinforcement,gao2022partial}. 

Total reward criterion~(TRC) generalizes stochastic shortest and longest path problems and has gained more attention~\cite{su2024stationary,su2025risk,Kallenberg2021markov,fei2021exponential, fei2021risk, ahmadi2021risk, cohen2021minimax, meggendorfer2022risk}. TRC has absorbing goal states and does not discount future rewards. A common assumption is that the MDP is transient and guarantees that any policy eventually terminates with positive probability. While this assumption is sufficient to guarantee finite risk-neutral expected returns, it is insufficient to guarantee finite returns with risk-averse objectives. For example, given an ERM objective, the risk level factor must also be sufficiently small to ensure a finite return~\cite{patek2001terminating,su2025risk}.

This paper derives risk-averse TRC model-free Q-learning algorithms for ERM-TRC  and EVaR-TRC objectives. There are three main challenges to deriving these risk-averse TRC Q-learning algorithms. First, generalizing the standard Q-learning algorithm to risk-averse objectives requires computing the full return distribution rather than just its expectation. Second, the risk-averse TRC Bellman operator may not be a contraction. The absence of a contraction property precludes the direct application of model-free methods in an undiscounted MDP. Third, instead of relying on the contraction property, we need to utilize other properties of the Bellman operator and an additional bounded condition to prove the convergence of risk-averse TRC Q-learning algorithms.

As our main contribution,in this chapter, we prove that the ERM-TRC Q-learning algorithm and the EVaR-TRC Q-learning algorithm converge to the optimal risk-averse value functions. We also show that the proposed Q-learning algorithms compute the optimal stationary policies for the ERM and EVaR objectives, and the optimal state-action value function $q$ can be computed by the stochastic gradient descent along the derivative of the exponential loss function.

The rest of the paper is organized as follows. We define our research setting and preliminary concepts in \cref{sec:preliminary}. In \cref{sec:erm}, we leverage the elicitability of ERM, define a new ERM Bellman operator, and propose Q-learning algorithms for the ERM and EVaR objectives. In \cref{sec:convergence-analysis}, we give a rigorous convergence proof of the proposed ERM-TRC Q-learning and EVaR-TRC Q-learning algorithms. Finally, the numerical results presented in \cref{sec:results} illustrate the effectiveness of our algorithms.

\section{Preliminaries for Risk-aversion in Markov Decision Processes }\label{sec:preliminary}

We first introduce our notations and overview relevant properties for monetary risk measures. We then formalize the MDP framework with the risk-averse objective and summarize the standard Q-learning algorithm. 

\paragraph{Notation} We denote by $\Real$ and $\Nats$ the sets of real and natural (including $0$) numbers, and $\bar{\Real} := \Real \cup \left\{ -\infty, \infty \right\}$ denotes the extended real line. We use $\Real_+$ and $\Real_{++}$ to denote non-negative and positive real numbers, respectively. We use a tilde to indicate a random variable, such as $\tilde{x}\colon \Omega \to \Real$ for the sample space $\Omega$. The set of all real-valued random variables is denoted as $\mathbb{X}:= \Real^{\Omega}$. Sets are denoted with calligraphic letters. 

We defined monetary risk measures in previous chapters. Below we restate monetary risk measures for the convenience of readers.

\paragraph{Monetary risk measures}
Monetary risk measures generalize the expectation operator to account for the uncertainty of the random variable. In this chapter, we focus on two risk measures. The first one is the \emph{entropic risk measure}~(ERM) defined for any risk level $\beta > 0$ and $\tilde{x}\in \mathbb{X}$ as~\cite{Follmer2016stochastic}
\begin{align} \label{eq:defn_ent_risk}
  \erm{\beta}{\tilde{x}} \;:=\; - \beta^{-1} \cdot \log  \E \exp{-\beta\cdot  \tilde{x}} ,
\end{align}
and can be extended to $\beta \in [0, \infty]$ as $\ermo_0[\tilde{x}] = \lim_{\beta \to 0^{+}} \erm{\beta}{\tilde{x}} = \E[\tilde{x}] $ and $\ermo_{\infty}[\tilde{x}] = \lim_{\beta\to \infty} \erm{\beta}{\tilde{x}} = \operatorname{ess} \inf[\tilde{x}]$. ERM is popular because of its simplicity and its favorable properties in multi-stage optimization formulations~\cite{su2025risk,hau2023entropic,Hau2023a,hau2024q}. In particular, dynamic decision-making with ERM allows for the existence of dynamic programming equations and Markov or stationary optimal policies.  In addition, in this work we leverage the fact that ERM is \emph{elicitable} which means that it can be estimated by solving a linear regression problem~\cite{bellini2015elicitable,embrechts2021bayes}.

The second risk measure we consider is \emph{entropic value at risk}~(EVaR), which is defined for a given risk level $\alpha \in (0,1)$ and $\tilde{x} \in \mathbb{X}$ as
\begin{equation}\label{eq:evar-def-app}
  \begin{aligned}
    \evar{\alpha}{\tilde{x}}
   & \;:=\;
    \sup_{\beta>0}  -\beta^{-1} \log \left(\alpha^{-1} \E \exp{ -\beta \tilde{x} } \right) 
    \;=\;
    \sup_{\beta>0}  \erm{\beta}{\tilde{x}} + \beta^{-1} \log  \alpha,
  \end{aligned}
\end{equation}
% \mm{Check the meaning of $\alpha$ and that it is consistent with what we say here}
and is extended to $\evar{0}{\tilde{x}} = \ess \inf [\tilde{x}]$ and $\evar{1}{\tilde{x}} = \E[\tilde{x}]$~\cite{Ahmadi-Javid2012}. It is important to note that the supremum in~\eqref{eq:evar-def-app} may not be attained even when $\tilde{x}$ is a finite discrete random variable~\cite{ahmadi2017analytical}. EVaR addresses several important shortcomings of ERM~\cite{su2025risk,hau2023entropic,Hau2023a,hau2024q}. In particular, EVaR is coherent and closely approximates popular and interpretable quantile-based risk measures, like VaR and CVaR~\cite{Ahmadi-Javid2012,hau2023entropic}.

\paragraph{Risk-Averse Markov Decision Processes}
We formulate the decision process as a \emph{Markov Decision Process}~(MDP) $(\states, \actions, p, r, \bm{\mu})$~\cite{puterman205markov}. The set $\states = \{1,2, \dots , S, e\}$ is the finite set of states and $e$ represents a \emph{sink} state. The set $\actions=\{1,2,\ldots, A\}$ is the finite set of actions. The transition function $p\colon \states \times \actions \to \probs{\states}$ represents the probability $p(s, a, s')$ of transitioning to $s'\in \mathcal{S}$ after taking $a\in \mathcal{A}$ in $s\in \mathcal{S}$. The function $r \colon \states \times  \actions \times \states \to \Real$ represents the reward $r(s,a,s') \in \Real$ associated with transitioning from $s\in \states$ and $a\in \mathcal{A}$ to $s'\in \states$. The vector $\bm{\mu} \in \probs{\states}$ is the initial state distribution.

This chapter has the same objective as that in \cref{chp:model-based} and focus on computing \emph{stationary deterministic policies} $\Pi := \actions^{\states}$ that maximize the \emph{total reward criterion}~(TRC):
\begin{equation} \label{eq:trm-objective}
 \max_{\pi\in \Pi} \lim_{t \to \infty}  \operatorname{Risk}^{\pi,\bm{\mu}} \left[ \sum_{k=0}^{t-1} r(\tilde{s}_k, \tilde{a}_k, \tilde{s}_{k+1})  \right],
\end{equation}
where Risk represents either ERM or EVaR risk measures. The limit in \eqref{eq:trm-objective} exists for stationary policies $\pi$ but may be infinite~\cite{su2025risk}. The superscript $\pi$ indicates the policy that guides the actions' probability, and the $\bm{\mu}$ indicates the distribution over the initial states.  

Recent work has shown that an optimal policy in~\eqref{eq:trm-objective} exists, is stationary, and has bounded return for a sufficiently small $\beta$ as long as the following assumptions hold~\cite{su2025risk}. The sink state $e$ satisfies that $p(e,a,e) = 1$ and $r(e,a,e) = 0$ for each $a\in \mathcal{A}$, and $\mu_e = 0$~\cite{su2025risk}. It is crucial to assume that the MDP is \emph{transient} for any $\pi\in \Pi$: 
\begin{equation} \label{eq:transient-condition}
\sum_{t = 0}^{\infty} \P^{\pi,s}\left[\tilde{s}_t = s'\right] \;<\;  \infty, \qquad
\forall s,s'\in \mathcal{S} \setminus  \left\{  e\right\} .
\end{equation}
Intuitively, this assumption states that any policy eventually reaches the sink state and effectively terminates. We adopt these assumptions in the remainder of the paper.

With a risk-neutral objective, transient MDPs guarantee that the return is bounded for each policy. However, that is no longer the case with the ERM objective. In particular, for some $\beta$ it is possible that the return is unbounded.  The return is bounded for a sufficiently small $\beta$ and there exists an optimal stationary policy~\cite{su2025risk}. In contrast, the return of EVaR is always bounded, and an optimal stationary policy always exists regardless of the risk level $\alpha$~\cite{su2025risk}.

\paragraph{Standard Q-learning}
To help with the exposition of our new algorithms, we now informally summarize the Q-learning algorithm for a risk-neutral setting~\cite{andrew2018reinforcement}. Q-learning is an essential component of most model-free reinforcement learning algorithms, including DQN and many actor-critic methods~\cite{la2013actor,dabney2018implicit,yoo2024risk}. Its simplicity and scalability make it especially appealing.

Q-learning iteratively refines an estimate of the optimal state-action value function $\tilde{q}_i\colon \mathcal{S} \times \mathcal{A} \to \Real, i \in \Nats$ such that it satisfies
\begin{equation} \label{eq:standard-qlearning}
\tilde{q}_{i+1}(\tilde{s}_i,\tilde{a}_i) = \tilde{q}_i(\tilde{s}_i,\tilde{a}_i) - \tilde{\eta}_i \tilde{z}_i,
\quad
\tilde{z}_i = r(\tilde{s}_i,\tilde{a}_i,\tilde{s}'_i) + \max_{a'\in\mathcal{A}} \tilde{q}_i(\tilde{s}_i,a') - \tilde{q}_i(\tilde{s}_i,\tilde{a}_i).
\end{equation}
Here, $\tilde{z}_i$ is also known as the TD residual. The algorithm assumes a stream of samples $(\tilde{s}_i, \tilde{a}_i, \tilde{s}'_i)_{i = 1, \dots }$ sampled from the transition probabilities and appropriately chosen step sizes $\tilde{\eta}_i$ to converge to the optimal state-action value function. Note that $\tilde{q}_i$ is random because it is a function of a random variable. It is well-known that the standard Q-learning algorithm can be seen as a stochastic gradient descent on the quadratic loss function~\cite{hau2024q,Asadi2024,shen2014risk}. We leverage this property to build our algorithms.

\section{Q-learning Algorithms: ERM and EVaR }\label{sec:erm}

In this section, we derive new Q-learning algorithms for the ERM and EVaR objectives. First, we propose the Q-learning algorithm for ERM in \cref{subsec:bellman-operator}, which requires us to introduce a new ERM Bellman operator based on the elicitability of the ERM risk measure. Then, we use this algorithm in \cref{sec:evar-Q-learning} to propose an EVaR Q-learning algorithm. 

\subsection{ERM Q-learning Algorithm}
\label{subsec:bellman-operator}

The algorithm we propose in this section computes the state-action function for multiple values of the risk level $\beta \in \mathcal{B}$ for some given non-empty \emph{finite} set $\mathcal{B} \subseteq \Real_{++}$ of positive reals. The set $\mathcal{B}$ may be a singleton or may include multiple values, which is necessary to optimize EVaR in the next section. 

Before describing the Q-learning algorithm, we define the ERM risk-averse state-action value function and describe the Bellman operator that can be used to compute it. We define the ERM risk-averse state-action value function $q_{\beta}\colon \mathcal{S} \times  \mathcal{A} \times \mathcal{B} \to \bar{\Real}$ for each $\beta \in \mathcal{B}$ as
\begin{equation} \label{eq:erm-q}
q^{\pi}(s,a, \beta)
:=
\lim_{t\to \infty} \ermo_{\beta}^{\pi,\langle s, a \rangle} \left[\sum_{k=0}^{t-1}r(\tilde{s}_k,\tilde{a}_k, \tilde{s}_{k+1})  \right],
\quad
q\opt(s,a,\beta)
:=
\max_{\pi \in \Pi} q^{\pi}(s,a,\beta).
\end{equation}
for each $s\in \mathcal{S}, a\in \mathcal{A}, \beta\in \mathcal{B}$. The limit in this equation exists by~\cite[lemma D.5]{su2024stationary}.

The superscript $ \left< s,a \right>$ in~\eqref{eq:erm-q} indicates the initial state and action are $\tilde{s}_0 = s,\tilde{a}_0 = a$, and the policy $\pi$ determines the actions henceforth. We use $\mathcal{Q} := \Real^{\mathcal{S} \times  \mathcal{A}}$ to denote the set of possible state-action value functions. 

To facilitate the computation of the value functions, we define the following ERM Bellman operator $B_{\beta}\colon \Real^{\mathcal{S} \times  \mathcal{A}} \to \Real^{\mathcal{S} \times  \mathcal{A}}$ as
% \textcolor{blue}{Notation inconsistency: $q(s,a)$ or $q(s,a,\beta)$? $r(s,a)$ or $r(s,a,s')$? }
\begin{equation} \label{eq:bellman-operator-model}
  (B_{\beta} q)(s,a) :=  \ermo^{a,s}_{\beta} \left[ r(s,a, \tilde{s}_1) + \max_{a' \in \actions}q(\tilde{s}_1,a') \right],
  \qquad
  \forall s\in \mathcal{S}, a\in \mathcal{A}, \beta \in \mathcal{B}, q\in \mathcal{Q}.
\end{equation}
Note that the Bellman operator $B_{\beta}$ applies to the value functions $q(\cdot, \cdot , \beta)$ for a fixed value of $\beta$. We abbreviate $B_{\beta}$ to $B$ when the risk level $\beta$ is clear from the context.

The following result, which follows from~\cite[theorem 3.3]{su2025risk}, shows that the ERM state-action value function can be computed as the fixed point of the Bellman operator.  
\begin{theorem} \label{thm:bellman-update-bar}
Assume some $\beta\in \mathcal{B}$ and suppose that $q_{\beta}\opt(s,a) > -\infty, \forall s\in \mathcal{S}, a\in \mathcal{A}$. Then $q_{\beta}\opt$ in~\eqref{eq:erm-q} is the unique solution to 
\begin{equation} \label{eq:bellman-update-bar}
  q_{\beta}\opt
  \;=\;
  B_{\beta} q_{\beta}\opt,
  \quad\text{where}\quad
  q\opt_{\beta}= q\opt(\cdot , \cdot , \beta).
\end{equation}
\end{theorem}
\begin{proof}
The exponential value function $w\opt(s)$ is the unique solution to the exponential Bellman equation~\cite[theorem 3.3]{su2025risk}. Therefore, for a state $s \in \mathcal{S}$ and the optimal policy $\pi$, state value function $v^{\pi}(s)$, and the exponential state value function $w^{\pi}(s)$, $\beta \in \mathcal{B}$, we have that
\begin{equation}
\label{eq:w-v}
    v^{\pi}(s) = -\beta^{-1} \log(-w^{\pi}(s))
\end{equation}
Because of the bijection between $w\opt(s)$ and $v\opt(s)$, we can conclude that $v\opt(s)$ is the unique solution to the regular Bellman equation.

The optimal state value function can be rewritten in terms of the optimal state-action value function in~\eqref{eq:v-q}
\begin{equation}
\label{eq:v-q}
    v\opt(s) = \max_{a' \in \mathcal{A}} q\opt(s,a,\beta)
\end{equation}
Then $ q\opt(s,a,\beta)$ exists. The value $q\opt$ is unique directly from the uniqueness of $v\opt$.
\end{proof}

Although the Bellman operator defined in~\eqref{eq:bellman-update-bar} can be used to compute the value function when the transition probabilities are known, it is inconvenient in the model-free reinforcement learning setting where the state-action value function must be estimated directly from samples.

We now turn to an alternative definition of the Bellman operator that can be used to estimate the value function directly from samples. To develop our ERM Q-learning algorithm, we need to define a Bellman operator that is amenable to compute its fixed point using stochastic gradient descent. For this purpose, we use the \emph{elicitability} property of risk measures~\cite{bellini2015elicitable}. ERM is known to be elicitable using the following \emph{loss} function $\ell_{\beta}\colon \Real \to \Real$: 
\begin{equation} \label{eq:erm-loss-function}
\ell_{\beta}(z) := \beta^{-1}(\exp{-\beta \cdot z}-1) + z, 
\quad
 \ell_{\beta}'(z) =
%\frac{\partial \beta^{-1}(e^{-\beta \cdot z}-1) + z }{\partial z}=
 1-\exp{-\beta \cdot z},
 \quad
 \ell_{\beta}'' (z)
=\beta \cdot \exp{-\beta \cdot  z}. 
\end{equation}
The functions $\ell'$ and $\ell''$ are the first and second derivatives which are be important in constructing and analyzing the Q-learning algorithm. 

The following proposition summarizes the elicitability property of ERM, which we need to develop our Q-learning algorithm. Elicitable risk measures can be estimated using regression from samples in a model-free way. 
\begin{proposition} \label{lemma:erm-argmin}
For each $\tilde{x} \in \mathbb{X}$ and  $\beta > 0$:
\begin{equation}
\label{eq:erm=argmin}
\erm{\beta}{\tilde{x}}
=
\argmin_{y \in \mathbb{R}}\E[\ell_{\beta}(\tilde{x}-y )] ,
\end{equation}
where the minimum is unique because $\ell_{\beta}$ is strictly convex. 
\end{proposition}
\begin{proof}
The proof is broken into two parts. First, we show that $\ell_\beta(\tilde{x}-y)$ is strictly convex, and a minimum value of $\ell_\beta$ exists. Second, we show that the minimizer of $\mathbb{E}[\ell_{\beta}(\tilde{x}-y )]$ is equal to $\erm{\beta}{\tilde{x}}$.

Take the first derivative of $\ell_\beta(\tilde{x}-y)$ with respect to $y$.
\begin{equation}
    \begin{aligned}
        \ell_{\beta}(\tilde{x}-y)
        & = \beta^{-1}(e^{-\beta \cdot (\tilde{x}-y)}-1) + (\tilde{x}-y) \\
       \ell_{\beta}'(y)
        & = \beta^{-1}(e^{-\beta \cdot (\tilde{x}-y)}) \cdot \beta - 1 \\
         \ell_{\beta}'(y)
         & = e^{-\beta \cdot (\tilde{x}-y)} - 1
    \end{aligned}
\end{equation}
Take the second derivative of $\ell_\beta(\tilde{x}-y)$  with respect to $y$
\begin{equation}
    \begin{aligned}
        \frac{\partial^2 \ell_{\beta}}{\partial y^2 } & =\frac{\partial (e^{-\beta \cdot (\tilde{x}-y)} -1) }{\partial y} \\
            \ell_{\beta}''(y) & = e^{-\beta \cdot (\tilde{x}-y)} \cdot \beta\\
             \ell_{\beta}''(y) & = \beta e^{-\beta \cdot (\tilde{x}-y)} > 0 \\
    \end{aligned}
\end{equation}
Therefore, $\ell_{\beta}$ is strongly convex with respect to $y$, and the minimum value of $\ell_{\beta}$ exists.

Second, take the first derivative of $\mathbb{E}[\ell_{\beta}(\tilde{x}-y )] $ with respective to $y$
    \begin{align*}
        \frac{\partial \E \bigl[ \beta^{-1}(e^{-\beta \cdot ( \tilde{x} -y)} -1) + (\tilde{x} -y)  \bigr]}{\partial y} & = 0 \\
        \E \bigl[ e^{-\beta \cdot (\tilde{x} -y)} -1  \bigr] & = 0\\
         \E \bigl[ e^{-\beta \cdot (\tilde{x} -y)}  \bigr] & = 1\\
          \E \bigl[ e^{-\beta \cdot \tilde{x}}  \bigr] \cdot  \E \bigl[ e^{\beta \cdot y}  \bigr]  & = 1 \\
           \E \bigl[ e^{-\beta \cdot \tilde{x}}  \bigr] \cdot e^{\beta \cdot y}    & = 1 \\
             e^{-\beta \cdot y}     & =   \E \bigl[ e^{-\beta \cdot \tilde{x}}  \bigr]  \\
             y &= -\beta^{-1} \log(\E \bigl[ e^{-\beta \cdot \tilde{x}}  \bigr])
    \end{align*}
    Therefore, the minimizer of $\mathbb{E}[\ell_{\beta}(\tilde{x}-y )]$ is equal to $\erm{\beta}{\tilde{x}}$.
\end{proof}

Using the elicitability property, we define the Bellman operator $\hat{B}_{\beta}\colon \mathcal{Q} \to \mathcal{Q}$ for $\beta\in \mathcal{B}$ as 
\begin{equation}
\label{eq:bellman-operator}
(\hat{B}_{\beta} q)(s,a)
:=
\argmin_{y \in \mathbb{R}} \E^{a,s} \left[ \ell_{\beta} \left(r(s,a,\tilde{s}_1) + \max_{a' \in \actions}q(\tilde{s}_1,a')-y \right)\right],
\;
\forall s\in \mathcal{S}, a\in \mathcal{A}.
\end{equation}

Following \cref{lemma:erm-argmin}, we get the following equivalence of the Bellman operator, which implies that their fixed points coincide.
\begin{theorem}
\label{thm:q-optimal-q-hat}
For each $\beta > 0$ and $q\in \mathcal{Q}$, we have that
\[
B_{\beta} q = \hat{B}_{\beta} q.
\]
\end{theorem} 

We can introduce the ERM Q-learning algorithm in \cref{alg:erm-Q-learning-algorithm} using the results above. This algorithm adapts the standard Q-learning approach to the risk-averse setting. Intuitively, it works as follows. Each iteration $i$ processes a single transition sample to update the optimal state-action value function estimate $\tilde{q}_i$. The value function estimates are random because the samples are random. The value function estimates are updated following a stochastic gradient descent along the derivative of the loss function $\ell_{\beta}$ defined in~\eqref{eq:erm-loss-function}. Each sample is used to simultaneously update the state-action value function for multiple values of $\beta\in \mathcal{B}$.

\begin{algorithm}
%\SetAlgoLined 

\textbf{Input}{Risk levels $\mathcal{B} \subseteq \Real$, samples: $(\tilde{s}_i,\tilde{a}_i, \tilde{s}_i')$, step sizes $\tilde{\eta}_i, i\in \Nats$, bounds $z_{\min}, z_{\max}$}\\
\textbf{Output}{Estimate state-action value function $\tilde{q}_i$} 
\begin{algorithmic}[1]
 \STATE  $ \tilde{q}_0(s,a,\beta) \gets 0, \quad \forall  s \in \mathcal{S}, a \in \mathcal{A} $ \;
\FOR{$i \in \Nats, s\in \mathcal{S}, a\in \mathcal{A}, \beta \in \mathcal{B}$}
  \IF{$s = \tilde{s}_i \wedge a = \tilde{a}_i$}
    \STATE $\tilde{z}_i(\beta) \gets r(s,a,\tilde{s}'_i) + \max_{a'\in\mathcal{A}} \tilde{q}_i(\tilde{s}'_i,a',\beta) - \tilde{q}_i(s,a,\beta)$ \;
     \IF{$\neg (z_{\min} \le \tilde{z}_i(\beta) \le z_{\max})$}
        \STATE   \textbf{return} {$\tilde{q} = -\infty$}
     \ENDIF
   \STATE $\tilde{q}_{i+1}(s,a,\beta) \gets \tilde{q}_i(s,a,\beta) - \tilde{\eta}_i \cdot ( \exp {-\beta \cdot \tilde{z}_i(\beta)} - 1)$ \;
  \ELSE
    \STATE $\tilde{q}_{i+1}(s,a,\beta) \gets \tilde{q}_i(s,a,\beta)$ 
  \ENDIF
\ENDFOR
 \caption{ERM-TRC Q-learning algorithm }
 \label{alg:erm-Q-learning-algorithm}
 \end{algorithmic}
\end{algorithm}

There are three main differences between \cref{alg:erm-Q-learning-algorithm} and the standard Q-learning algorithm described in \cref{sec:preliminary}.  
First, standard Q-learning aims to maximize the expectation objective, and the proposed algorithm maximizes the ERM-TRC objective. Second, the state-action value function update
 % \mm{TODO} 
follows a sequence of stochastic gradient steps, but it replaces the quadratic loss function with the exponential loss function $\ell_{\beta}$ in~\eqref{eq:erm-loss-function}. Third, $q$ values in the proposed algorithm
need an additional bounded condition that the TD residual $\tilde{z}_i(\beta)$ is in $[z_{\min}, z_{\max}]$.

\begin{remark}
The $q$ value can be unbounded for two main reasons. First, as the value of $\beta$ increases, the ERM value function does not increase~\cite[Lemma A.7]{hau2023entropic} and can reach $-\infty$~\cite[Theorem 3.3]{su2025risk}. Second, the agent has some probability of repeatedly receiving a reward for the same action, leading to an uncontrolled increase in the $q$ value for that action. 
 \end{remark}
Note that if $\tilde{z}_i(\beta)$ is outside the $[z_{\min}, z_{\max}]$ range, it indicates that the value of the risk level $\beta$ is so large that $\tilde{q}$ is unbounded. This aligns with the conclusion that the value function may be unbounded for a large risk factor~\cite{su2025risk,patek2001terminating}. Detecting unbounded values of $q$ caused by random chance is useful but is challenging and beyond the scope of this work.

\begin{remark} \label{lemma:estimating-z-bounds}
We now discuss how to estimate $z_{\min}$ and $z_{\max}$ in \cref {alg:erm-Q-learning-algorithm}.
Assume that the sequences $\{\tilde{\eta}_i\}_{i=0}^{\infty}$ and $\{(\tilde{s}_i, \tilde{a}_i,\tilde{s}'_i) \}_{i=0}^{\infty}$ used in \cref {alg:erm-Q-learning-algorithm}, $\beta > 0$, we have
\[
z_{\min}(\beta) = - 2 \max\{|c - \beta \cdot d|, |c|\} -\|r\|_{\infty}, \quad z_{\max}(\beta) = 2 \max\{|c - \beta \cdot d|, |c|\} +\|r\|_{\infty}.
\]
Where $\|r\|_{\infty}=\max_{s,s' \in \mathcal{S}, a \in \mathcal{A}} |r(s,a,s')|$. The constants $c$ and $d$ can be estimated by \cref{alg:z-bounds}. 
\end{remark}

We derive the $z$ bounds as follows. From line 4 of \cref{alg:erm-Q-learning-algorithm}, we have 
\[
 \tilde{z}_i(\beta) \gets r(s,a,\tilde{s}'_i) + \max_{a'\in\mathcal{A}} \tilde{q}_i(\tilde{s}'_i,a',\beta) - \tilde{q}_i(s,a,\beta), s \in \mathcal{S}, a \in \mathcal{A}, \beta \in \mathcal{B}, i \in \mathbb{N} 
\]
Do some algebraic manipulation,
\begin{align*}
    \mid \tilde{z}_i(\beta) \mid  & \le \|r\|_{\infty} + |q|_{\max} + |q|_{\max} \\
      \mid \tilde{z}_i(\beta) \mid  & \le \|r\|_{\infty} + 2|q|_{\max}
\end{align*}
where $\displaystyle \|r\|_{\infty} = \max_{s,s' \in \mathcal{S}, a \in \mathcal{A}} |r(s,a,s')|$ and $\displaystyle |q|_{\max} = \max_{s \in \mathcal{S}, a \in \mathcal{A}} |q_i(s,a,\beta)|, i \in \mathbb{N}$.

Let us estimate $|q|_{\max}$. For any random variable $\tilde{x}$, for any risk level $\beta \in \mathcal{B}$, $\erm{\beta}{\tilde{x}}$ is non-increasing and satisfies monotonicity~\cite[Lemma A.7 and Lemma A.8]{hau2023entropic},
\[
 \E[\tilde{x}] - \beta (x_{\max}-x_{\min})^2 /8 \le \erm{\beta}{\tilde{x}} \le \E[\tilde{x}]
\]
 That is, 
\[
 \E[\tilde{x}] - \beta (x_{\max}-x_{\min})^2 /8 \le \tilde{q}_{\infty}(s,a,\beta) \le \E[\tilde{x}], s \in \mathcal{S}, a \in \mathcal{A}
\]
Then, for any $\beta \in \mathcal{B}$,
\begin{align*}
     |q|_{\max} & \approx \max_{s \in \mathcal{S}, a \in \mathcal{A}} |\tilde{q}_{\infty}(s,a,\beta)|  \\
     & = \max\{ |\E[\tilde{x}] - \beta (x_{\max}-x_{\min})^2 /8|, |\E[\tilde{x}]|\}
\end{align*}
As mentioned in \cref{sec:preliminary}, 
\[
\ermo_0[\tilde{x}] = \lim_{\beta \to 0^{+}} \erm{\beta}{\tilde{x}} = \E[\tilde{x}] 
\]
Then 
\[
\E[\tilde{x}]  \approx \max_{s \in \mathcal{S}, a \in \mathcal{A}} \tilde{q}_{\infty}(s,a,0^{+})
\]
We estimate $\E[\tilde{x}]$, $x_{\max}$, $x_{\min}$ by setting $\beta_c = 1^{-10}$ in \cref{alg:z-bounds}. Therefore, for any $\beta \in \mathcal{B}$,  
\[
z_{\min}(\beta) = -||r||_{\infty} -2\max\{|c - \beta \cdot d|, |c|\}, \quad z_{\max}(\beta) = ||r||_{\infty} +2\max\{|c - \beta \cdot d|, |c|\}
\]
where $c= \max \tilde{q}_{\infty}(s,a, \beta_c), s \in \mathcal{S}, a \in \mathcal{A}$ and $d =(x_{\max}-x_{\min})^2/8 $. 

\begin{algorithm}
\textbf{Input: }{Risk levels $\beta_{c} =1^{-10} $, samples: $(\tilde{s}_i,\tilde{a}_i, \tilde{s}_i')$, step sizes $\tilde{\eta}_i, i\in \Nats$}

\textbf{Output: }{Estimated expectation value $c$, $x_{\min}$, and $x_{\max}$} 
\begin{algorithmic}[1]
  \STATE $ \tilde{q}_0(s,a,\beta_c) \gets 0,\tilde{x}(s,a,\beta_{c}) \gets 0, \quad \forall  s \in \mathcal{S}, a \in \mathcal{A} $ \;
\FOR{$i \in \Nats, s\in \mathcal{S}, a\in \mathcal{A}$}
  \IF{$s = \tilde{s}_i \wedge a = \tilde{a}_i$}
   \STATE $\tilde{z}_i(\beta_{c}) \gets r(s,a,\tilde{s}'_i) + \max_{a'\in\mathcal{A}} \tilde{q}_i(\tilde{s}'_i,a',\beta_{c}) - \tilde{q}_i(s,a,\beta_{c})$ \;
   \STATE $\tilde{q}_{i+1}(s,a,\beta_{c}) \gets \tilde{q}_i(s,a,\beta_{c}) - \tilde{\eta}_i \cdot ( \exp {-\beta \cdot \tilde{z}_i(\beta_{c})} - 1)$ \;
   \STATE $\tilde{x}(s,a,\beta_{c}) \gets \tilde{x}(s,a,\beta_{c}) + r(s,a)  $
  \ELSE
    \STATE $\tilde{q}_{i+1}(s,a,\beta_{c}) \gets \tilde{q}_i(s,a,\beta_{c})$ 
  \ENDIF
\ENDFOR
 \STATE $c \gets \max \tilde{q}_{\infty}(s,a,\beta_{c}), s \in \mathcal{S}, a \in \mathcal{A}$ \;
 \STATE $x_{\min} \gets \min\tilde{x}(s,a,\beta_{c}), s \in \mathcal{S}, a \in \mathcal{A} $ \;
\STATE $x_{\max} \gets \max\tilde{x}(s,a,\beta_{c}), s \in \mathcal{S}, a \in \mathcal{A} $ \;
 \caption{A heuristic algorithm for computing $z$ bounds}
 \label{alg:z-bounds}
 \end{algorithmic}
\end{algorithm}

\subsection{EVaR Q-learning Algorithm}
\label{sec:evar-Q-learning}
In this section, we adapt our ERM Q-learning algorithm to compute the optimal policy for the static EVaR-TRC objective in~\eqref{eq:trm-objective}. As mentioned in \cref{sec:preliminary}, EVaR is preferable to ERM because it is coherent and closely approximates VaR and CVaR. The construction of an optimal EVaR policy follows the standard methodology proposed in prior work~\cite{su2025risk,hau2023entropic}. We only briefly summarize it here.

The main challenge in solving~\eqref{eq:trm-objective} is that EVaR is not dynamically consistent, and the supremum in~\eqref{eq:evar-def-app} may not be attained. We show that the EVaR-TRC problem can be reduced to a sequence of ERM-TRC problems, similarly to the discounted case~\cite{hau2023entropic} and the undiscounted case~\cite{su2025risk}. We define the objective function $h\colon \Pi \times \Real_{++} \to  \bar{\Real}$:
\begin{equation} \label{eq:h-definition}
  \begin{aligned}
  h(\pi, \beta)
  &\;:=\; 
  \ermo_{\beta}^{\pi, \mu }\left[q^{\pi}(\tilde{s}_0, \tilde{a}_0, \beta)\right] + \beta^{-1} \log(\alpha) \\
  &\; =\; 
 \lim_{t\to \infty} \ermp{\beta}{\pi,\mu}{\sum_{k=0}^{t-1}r(\tilde{s}_k,\tilde{a}_k,\tilde{s}_{k+1})} + \beta^{-1} \log(\alpha).
  \end{aligned}
\end{equation}
Recall that $\tilde{s}_0$ is distributed according to the initial distribution $\mu$ and $\tilde{a}_0$ is the action distributed according to the policy $\pi$. The equality in the equation above follows directly from the definition of the value function in~\eqref{eq:erm-q}.

We now compute a $\delta$-optimal EVaR-TRC policy for any $\delta > 0$ by solving a sequence of ERM-TRC problems. Following prior work~\cite{su2025risk}, we replace the supremum over continuous $\beta$ in the definition of EVaR in \eqref{eq:evar-def-app} with a maximum over a \emph{finite} set $\mathcal{B}(\beta_0, \delta)$ of discretized $\beta$ values chosen as
\begin{subequations} \label{eq:b-set-defs}
\begin{equation} \label{eq:b-set-definition}
  \mathcal{B}(\beta_0, \delta) \;:=\;  \left\{ \beta_0, \beta_1, \dots , \beta_K \right\},
\end{equation}
where  $0 < \beta_0 < \beta_1 < \dots  < \beta_K$, and
\begin{equation}
\label{eq:beta-construction}
 % \beta_0  \; :=\; \frac{8  \delta}{(z_{\max}  -z_{\min})^2}, 
\beta_{k+1} \;:=\;  \frac{\beta_k \log \frac{1}{\alpha}}{\log\frac{1}{ \alpha } -\beta_k \delta} ,
\quad
\beta_K \; \ge\;  \frac{\log \frac{1}{\alpha}}{\delta},
\quad
K \; \ge\;  \frac{\log(\theta)}{\log (1-\theta)},
\end{equation}
\end{subequations}
for $\theta := -\nicefrac{\beta_0 \cdot \delta}{\log(\alpha)}$, and the minimal $K$ that satisfies the inequality. Note that the construction in~\eqref{eq:b-set-defs}  differs in the choice of $\beta_0$ from $\mathcal{B}(\beta_0, \delta)$ in~\cite{hau2023entropic}. 

Let us derive $\beta_0$. For a desired precision $\delta > 0$, $\beta_0$ is chosen such that
\begin{equation}
\label{eq:difference-expectation-erm}
      \E^{\pi\opt,\bm{\mu}}[\tilde{x}]  -  \ermp{\beta_0}{\pi\opt,\bm{\mu}}{\tilde{x}}  \le \delta 
\end{equation}
For a random variable $\tilde{x}$ and $\beta > 0$~\cite[Lemma A.8]{hau2023entropic}  
\[
 \E[\tilde{x}] - \erm{\beta}{\tilde{x}} \le \beta \cdot (x_{\max} - x_{\min})^2/8
\]
Then we have 
\begin{equation}
\label{eq:erm-bound}
    \E^{\pi\opt,\bm{\mu}}[\tilde{x}]  -  \ermp{\beta_0}{\pi\opt,\bm{\mu}}{\tilde{x}}   \le \beta_0 \cdot (x_{\max} - x_{\min})^2/8
\end{equation}
When the equality conditions in~\eqref{eq:difference-expectation-erm} and~\eqref{eq:erm-bound} hold, we have  
\begin{align*}
    \beta_0 \cdot (x_{\max}  -x_{\min})^2/8 & = \delta \\
               \beta_0 & = \frac{8  \delta}{(x_{\max}  -x_{\min})^2}
\end{align*}
where $x_{\max}$ and $x_{\min}$ are estimated in \cref{alg:z-bounds}.

The following theorem summarizes the fact that the EVaR-TRC problem reduces to a sequence of ERM-TRC optimization problems. 
\begin{theorem}[Theorem 4.2 in \cite{su2024stationary}] \label{thm:optimal-evar-erm}
For any $\delta > 0$ and a sufficiently small $\beta_0 > 0$, let
\[
  (\pi\opt,\, \beta\opt)  \in
  \argmax_{(\pi,\beta)\in \Pi \times \mathcal{B}(\beta_0, \delta)}
  h(\pi,\beta).
\]
Then, the limits below exist and satisfy that
\begin{equation} \label{eq:evar-guarantee}
\lim_{t\to \infty}  
\evaro_{\alpha}^{\pi\opt, \bm{\mu}}
\left[ \sum_{k=0}^{t-1} r(\tilde{s}_k,\tilde{a}_k,\tilde{s}_{k+1}) \right]
\; \ge \;
  \sup_{\pi\in \Pi} \lim_{t\to \infty} \sup_{\beta>0} h(\pi,\beta)
  -\delta.
\end{equation}
\end{theorem}

\begin{algorithm}
\textbf{Input}{desired precision $\delta > 0$, risk level $\alpha \in (0,1)$, initial $\beta_0 > 0$, samples: $(\tilde{s}_i,\tilde{a}_i, \tilde{s}_i')$, $i\in \Nats$ }
\textbf{Output}{$\delta$-optimal policy $\pi\opt \in \Pi$}
\begin{algorithmic}[1]
\STATE Construct $\mathcal{B}(\beta_0, \delta)$ as described in~\eqref{eq:b-set-defs} \;
\STATE Compute $(\pi\opt_{\beta},\, h\opt(\beta))$ by \cref{alg:erm-Q-learning-algorithm} for each $\beta  \in \mathcal{B}(\beta_0, \delta)$ where $h\opt(\beta) = \max_{\pi \in \Pi } h(\pi, \beta)$\;
\STATE Let $\beta\opt \in
\argmax_{\beta \in \mathcal{B}(\beta_0, \delta)} h\opt(\beta)$\;
 \STATE \textbf{return } $\pi\opt_{\beta \opt}$
\caption{EVaR-TRC Q-learning algorithm}
\label{alg:evar-algorithm}
\end{algorithmic}
\end{algorithm}

\Cref{alg:evar-algorithm} summarizes the procedure for computing a $\delta$-optimal EVaR policy. Note that the value of $h(\pi,\beta)$ may be $-\infty$, indicating either the ERM-TRC objective is unbounded for $\beta$, or the Q-learning algorithm diverged by random chance. 

\begin{remark}
Note that \cref{alg:evar-algorithm} assumes a small $\beta_0$ as its input. 
% We provide the procedure for computing the value $\beta_0$.
It is unclear whether it is possible to obtain a prior bound on $\beta_0$ without knowing the model. We, therefore, employ the heuristic outlined in \cref{alg:z-bounds} that estimates a lower bound $x_{\min}$ and an upper bound $x_{\max}$ on the random variable of returns. Then, we set \(  \beta_0  := \nicefrac{8  \delta}{(x_{\max}  -x_{\min})^2}. \)
\end{remark}

\section{Convergence Analysis}
\label{sec:convergence-analysis}
This section presents our main convergence guarantees for the ERM-TRC Q-learning and EVaR-TRC Q-learning algorithms. 

We require the following standard assumption to prove the convergence of the proposed Q-learning algorithms. The intuition of \cref{assump:transition-prop} is that each state-action pair must be visited infinitely often. 

\begin{assumption}
\label{assump:transition-prop}
The input to \cref{alg:erm-Q-learning-algorithm} and \cref{alg:evar-algorithm} satisfies that
\[
\mathbb{P}[\tilde{s}'_i = s' \mid \mathcal{G}_{i-1}, \tilde{s}_i, \tilde{a}_i,\tilde{\eta}_i] = p(\tilde{s}_i,\tilde{a}_i,s'),
\qquad
    \forall s' \in \mathcal{S},\, \forall i \in \mathbb{N},
\]
almost surely, where $\mathcal{G}_{i-1}:= (\tilde{\eta}_l, (\tilde{s_l},\tilde{a}_l,\tilde{s}'_l))_{l=0}^{i-1}$.
\end{assumption}

The following theorem shows that the proposed ERM-TRC Q-learning algorithm enjoys convergence guarantees comparable to standard Q-learning. As \cref{lemma:estimating-z-bounds} states, $z_{\min}$ and $z_{\max}$ are chosen to ensure that $q$ values are bounded. 
\begin{theorem}
\label{theorem:erm-q-converge}
For $\beta \in \mathcal{B}$, assume that the sequence  $\left(\tilde{\eta}_i\right)_{i=0}^{\infty}$ and $\left((\tilde{s}_i, \tilde{a}_i,\tilde{s}'_i) \right)_{i=0}^{\infty}$ used in \cref {alg:erm-Q-learning-algorithm} satisfies \cref{assump:transition-prop} and step size condition 
\[
    \sum_{i=0 }^{\infty} \tilde{\eta}_i= \infty , \quad \sum_{i =0}^{\infty}\tilde{\eta}_i^2 < \infty,
\] 
where $i \in \{ i \in \mathbb{N} \mid (\tilde{s}_i, \tilde{a}_i) = (s,a)\}$, 
if $\tilde{z}_i \in  [z_{\min}, z_{\max}]$ almost surely, then the sequence $(\tilde{q}_i)_{i=0}^{\infty}$ produced by \cref{alg:erm-Q-learning-algorithm} convergences almost surely to $q_{\infty}$ such that $q_{\infty} = \hat{B}_{\beta} q_{\infty}$.
\end{theorem}

The proof of \cref{theorem:erm-q-converge} follows an approach similar to that in the proofs of standard Q-learning~\cite{bertsekas1996neuro} with three main differences. First, the algorithm converges for an undiscounted MDP, and $\hat{B}$ is not a contraction captured by~\cite[assumption 4.4]{bertsekas1996neuro}, which is restated here as \cref{assump:monotonicity}. Instead, we show that $\hat{B}$ satisfies monotonicity in \cref{lemma:bellman-operator-monotonicity}. Second, \cref{assump:monotonicity} leads to a convergence result somewhat weaker than the results for a contraction. Then, a separate boundedness condition\cite[proposition 4.6]{bertsekas1996neuro} is imposed. Third, using exponential loss function $\ell_{\beta}$ requires a more careful choice of step-size than the standard analysis. The proof of \cref{theorem:erm-q-converge} is shown in \cref{proof:theorem-erm-q-converge}.

\begin{lemma}
\label{lemma:bellman-operator-monotonicity}
For $\beta \in \mathcal{B}$, let $\bm{1} \in \mathbb{R}^n$ represent the vector of all ones. Then the ERM Bellman operator defined in~\eqref{eq:bellman-operator} satisfies the monotonicity property if for some  $g >0$ and $q\opt\colon \Real^{\mathcal{S} \times \mathcal{A}} \to \Real^{\mathcal{S} \times \mathcal{A}}$ and for all $q\colon \Real^{\mathcal{S} \times \mathcal{A}} \to \Real^{\mathcal{S} \times \mathcal{A}}$ it satisfies that
\begin{align}
  \tag{a}
  x \le y &\quad\Rightarrow \quad \hat{B}_{\beta} x \le \hat{B}_{\beta} y \\
  \tag{b}
  \hat{B}_{\beta}q\opt &\quad=\quad q\opt \\
  \tag{c}
  \hat{B}_{\beta}q -g\cdot \bm{1} \quad\le \quad \hat{B}_{\beta}(q - g\cdot \bm{1}) &\quad\le \quad \hat{B}_{\beta}(q + g\cdot \bm{1}) \quad\le \quad \hat{B}_{\beta}q + g\cdot \bm{1}
\end{align}
Here, the relations hold element-wise.
\end{lemma}
\begin{proof}

  First,  from \cref{thm:q-optimal-q-hat}, we have 
\begin{align*}
    B_{\beta}q &= \hat{B}_{\beta}q  \\
       &\Downarrow \\
\ermo^{a,s}_{\beta} \Bigl[ r(s,a,\tilde{s}_1) + \max_{a' \in \actions}q(\tilde{s}_1,a',\beta) \Bigr] &= \argmin_{y \in \mathbb{R}} \E^{a,s} \Bigl[ \ell_{\beta} \bigl(r(s,a,\tilde{s}_1) + \max_{a' \in \actions}q(\tilde{s}_1,a',\beta)-y \bigr)\Bigr]
\end{align*}
 
For the part $(a)$, ERM is monotone~\cite{hau2023entropic}. That is, for some fixed $\beta \in \mathcal{B}$ value, if $x \le y$, then $ \hat{B}_{\beta}x \le \hat{B}_{\beta}y$.

For the part $(b)$,  $\ell_{\beta}$ is a strongly convex function because $\ell''_{\beta}(y) = \beta \cdot e^{-\beta \cdot(\tilde{x}-y)} > 0$. Then there will be a unique $y\opt$ value such that $\E^{a,s} \Bigl[ \ell_{\beta} \bigl(r(s,a,\tilde{s}_1) + \max_{a' \in \actions}q(\tilde{s}_1,a',\beta)-y \bigr)\Bigr]$ attains the minimum value, and the $y\opt$ value is equal to $q\opt(s,a,\beta)$, which follows from \cref{lemma:erm-argmin}. So $\hat{B}_{\beta}q\opt = q\opt$

For the part $(c)$, ERM is monotone and satisfies the law-invariance property~\cite{hau2023entropic}.
\begin{align*}
   \hat{B}_{\beta}(q)  - 1\cdot g 
  \stackrel{\text{(1)}}{\le}  \hat{B}_{\beta}(q  - 1\cdot g)
  \stackrel{\text{(2)}}{\le} \hat{B}_{\beta}(q  + 1\cdot g) 
   \stackrel{\text{(3)}}{\le}  \hat{B}_{\beta}(q)  + 1\cdot g 
\end{align*}
Steps $(1)$ and $(3)$ follow from the law-invariance property of ERM. Step $(2)$ follows from the mononotone property of ERM.
\end{proof}

The following lemma shows that the loss function $l_{\beta}$ is strongly convex and has a Lipschitz-continuous gradient. These properties are instrumental in showing the uniqueness of our value functions and analyzing the convergence of our Q-learning algorithms by analyzing them as a form of stochastic gradient descent.
\begin{lemma}
\label{lemma:gradient-lipschitz-bound-z}
The function $\ell_{\beta}\colon [z_{\min}, z_{\max}] \to \Real$ defined in~\eqref{eq:erm-loss-function} is $l$-strongly convex with  $l = \beta \exp{-\beta \cdot z_{\max}}$ and its derivative $\ell_{\beta}'(z)$ is $L$-Lipschitz-continuous with $L = \beta \cdot  \exp{-\beta \cdot z_{\min}}$.
\end{lemma}
\begin{proof}
We use the fact that $\ell_{\beta}$ is twice continuously differentiable and prove the $l$-strong convexity from~\cite[theorem~2.1.11]{Nesterov2018a}:
\[
  l = \inf_{z\in [z_{\min}, z_{\max}]} \ell_{\beta}''(z)
  = \inf_{z\in [z_{\min}, z_{\max}]} \beta \exp{-\beta z} 
  = \beta \exp{-\beta z_{\max}}. 
\]
To prove $L$-Lipschitz continuity of the derivative, using~\cite[Theorem~9.7]{Rockafellar2009}, we have that
\[
  L = \sup_{z\in [z_{\min}, z_{\max}]} |\ell_{\beta}''(z)|
  = \sup_{z\in [z_{\min}, z_{\max}]} |\beta \exp{-\beta z} |
  = \beta \exp{-\beta z_{\min}}.
\]
  
\end{proof}

\cref{theorem:evar-converge}, which follows immediately from \cref{thm:optimal-evar-erm}, shows that \cref{alg:evar-algorithm} enjoys convergence guarantees and converges to a $\delta$-optimal EVaR policy.
\begin{corollary}
 \label{theorem:evar-converge}  
For $\alpha \in (0,1)$ and $\delta > 0$, assume that the sequence  $\{\tilde{\eta}_i\}_{i=0}^{\infty}$ and $\{(\tilde{s}_i, \tilde{a}_i,\tilde{s}'_i) \}_{i=0}^{\infty}$ used in \cref {alg:erm-Q-learning-algorithm} satisfies \cref{assump:transition-prop} and step size condition 
\[
    \sum_{i=0 }^{\infty} \tilde{\eta}_i= \infty , \quad \sum_{i =0}^{\infty}\tilde{\eta}_i^2 < \infty,
\] 
where $i \in \{ i \in \mathbb{N} \mid (\tilde{s}_i, \tilde{a}_i) = (s,a)\}$, if $\tilde{z}_i \in  [z_{\min}, z_{\max}]$. Then \cref{alg:evar-algorithm} converges to the $\delta$-optimal stationary policy $\pi\opt$ for the EVaR-TRC objective in~\eqref{eq:trm-objective} almost surely.
\end{corollary}

The proof of \cref{theorem:evar-converge} follows a similar approach to the proofs of \cref{theorem:erm-q-converge}. However, we also need to show that one can obtain an optimal ERM policy for an appropriately chosen $\beta$ that approximates an optimal EVaR policy arbitrarily closely.
\begin{proof}
    From \cref{theorem:erm-q-converge}, for some $\beta \in \mathcal{B}$, the ERM-Q learning algorithm converges to the optimal ERM value function. \cref{thm:optimal-evar-erm} shows that there exists $\delta-$optimal policy such that it is an ERM-TRC optimal for some $\beta \in \mathcal{B}$. It is sufficient to compute an ERM-TRC optimal policy for one of those $\beta$ values. This analysis shows that \cref{alg:evar-algorithm} converges to its optimal EVaR value function.
\end{proof}

\section{Numerical Evaluation}
\label{sec:results}
In this section, we evaluate our algorithms on two tabular domains:  cliff walking~(CW)~\cite{andrew2018reinforcement} and gambler's ruin~(GR)~\cite{su2025risk}. 

First, to evaluate our EVaR Q-learning algorithm with risk levels $\alpha = 0.2$ and $\alpha = 0.6$, we use the cliff walk problem. In this problem, an agent starts with a random state (cell in the grid world) that is uniformly distributed over all non-sink states and walks toward the goal state labeled by $g$ shown in \cref{fig:optimal-policy-alpha0.2}. At each step, the agent takes one of four actions: up, down, left, or right. The action will be performed with a probability of $0.91$, and the other three actions will also be performed separately with a probability of $0.03$. If the agent hits the wall, it will stay in its place. The reward is zero in all transitions except for the states marked with $c$, $d$, and $g$. The agent receives $2$ in state $g$, $0.004$ in state $d$, and $-0.5,-0.6,-0.7,-0.8$, or $-0.9$ in cliff region marked with $c$ from the left to the right. If an agent steps into the cliff region, it will immediately be returned to the state marked with $b$. Note that the agent has unlimited steps, and the future rewards are not discounted. 

\begin{figure}
% For $\alpha =0.2$
\begin{minipage}{0.45\textwidth}
  \centering
   \begin{tikzpicture}
   \centering
\draw[step=0.5cm,color=gray] (-2,-1.5) grid (1.5,2);
\node at (+1.37,+1.75) {b}; %cell 7
 \node at (+1.25,-1.25) {g}; %cell 49
\node at (+0.80,-1.25) {c}; %cell 48
\node at (+0.30,-1.25) {c}; %cell 47
\node at (-0.20,-1.25) {c}; %cell 46
\node at (-0.70,-1.25) {c}; %cell 45
\node at (-1.20,-1.25) {c}; %cell 44
\node at (-0.65,-0.75) {d}; %cell 38
\node at (-1.75,+1.75) {$\rightarrow$}; %cell 1
\node at (-1.25,+1.75) {$\rightarrow$}; %cell 2
\node at (-0.75,+1.75) {$\rightarrow$};%cell 3
\node at (-0.25,+1.75) {$\rightarrow$}; % cell 4
\node at (+0.25,+1.75) {$\rightarrow$};%cell 5
\node at (+0.75,+1.75) {$\downarrow$};%cell 6
\node at (+1.1,+1.75) {$\downarrow$};%cell 7
\node at (-1.75,+1.25) {$\rightarrow$};%cell 8
\node at (-1.25,+1.25) {$\rightarrow$};%cell 9
\node at (-0.75,+1.25) {$\rightarrow$};%cell 10
\node at (-0.25,+1.25) {$\rightarrow$};%cell 11
\node at (+0.25,+1.25) {$\rightarrow$};%cell 12
\node at (+0.75,+1.25) {$\downarrow$};%cell 13
\node at (+1.25,+1.25) {$\downarrow$};%cell 14
\node at (-1.75,+0.75) {$\rightarrow$};%cell 15
\node at (-1.25,+0.75) {$\rightarrow$};%cell 16
\node at (-0.75,+0.75) {$\rightarrow$};%cell 17
\node at (-0.25,+0.75) {$\rightarrow$};% cell 18
\node at (+0.25,+0.75) {$\rightarrow$};%cell 19
\node at (+0.75,+0.75) {$\rightarrow$};%cell 20
\node at (+1.25,+0.75) {$\downarrow$};%cell 21
\node at (-1.75,+0.25) {$\rightarrow$};%cell 22
\node at (-1.25,+0.25) {$\rightarrow$};%cell 23
\node at (-0.75,+0.25) {$\rightarrow$};%cell 24
\node at (-0.25,+0.25) {$\rightarrow$};%cell 25
\node at (+0.25,+0.25) {$\rightarrow$};%cell 26
\node at (+0.75,+0.25) {$\rightarrow$};%cell 27
\node at (+1.25,+0.25) {$\rightarrow$};%cell 28
\node at (-1.75,-0.25) {$\uparrow$};%cell 29
\node at (-1.25,-0.25) {$\uparrow$};%cell 30
\node at (-0.75,-0.25) {$\uparrow$};%cell 31
\node at (-0.25,-0.25) {$\uparrow$};%cell 32
\node at (+0.25,-0.25) {$\uparrow$};%cell 33
\node at (+0.75,-0.25) {$\rightarrow$};%cell 34
\node at (+1.25,-0.25) {$\uparrow$};%cell 35
\node at (-1.75,-0.75) {$\uparrow$};%cell 36
\node at (-1.25,-0.75) {$\uparrow$};%cell 37
\node at (-0.80,-0.75) {$\uparrow$};%cell 38
\node at (-0.25,-0.75) {$\uparrow$};%cell 39
\node at (+0.25,-0.75) {$\uparrow$};%cell 40
\node at (+0.75,-0.75) {$\uparrow$};%cell 41
\node at (+1.25,-0.75) {$\downarrow$};%cell 42
\node at (-1.75,-1.25) {$\uparrow$};%cell 43
\end{tikzpicture}
\caption{EVaR optimal policy with $\alpha=0.2$}
    \label{fig:optimal-policy-alpha0.2}
\end{minipage}
\hspace{0.5cm}
% For $\alpha =0.6$
\begin{minipage}{0.45\textwidth}
\centering
   \begin{tikzpicture}
\draw[step=0.5cm,color=gray] (-2,-1.5) grid (1.5,2);
\centering
\node at (+1.37,+1.75) {b}; %cell 7
 \node at (+1.25,-1.25) {g}; %cell 49
\node at (+0.80,-1.25) {c}; %cell 48
\node at (+0.30,-1.25) {c}; %cell 47
\node at (-0.20,-1.25) {c}; %cell 46
\node at (-0.70,-1.25) {c}; %cell 45
\node at (-1.20,-1.25) {c}; %cell 44
\node at (-0.65,-0.75) {d}; %cell 38
\node at (-1.75,+1.75) {\textcolor{blue}{$\leftarrow$}}; %cell 1
\node at (-1.25,+1.75) {\textcolor{blue}{$\leftarrow$}}; %cell 2
\node at (-0.75,+1.75) {\textcolor{blue}{$\leftarrow$}};%cell 3
\node at (-0.25,+1.75) {\textcolor{blue}{$\leftarrow$}}; % cell 4
\node at (+0.25,+1.75) {\textcolor{blue}{$\leftarrow$}};%cell 5
\node at (+0.75,+1.75) {\textcolor{blue}{$\rightarrow$}};%cell 6
\node at (+1.1,+1.75) {\textcolor{blue}{$\rightarrow$}};%cell 7
\node at (-1.75,+1.25) {\textcolor{blue}{$\uparrow$}};%cell 8
\node at (-1.25,+1.25) {\textcolor{blue}{$\uparrow$}};%cell 9
\node at (-0.75,+1.25) {\textcolor{blue}{$\uparrow$}};%cell 10
\node at (-0.25,+1.25) {\textcolor{blue}{$\leftarrow$}};%cell 11
\node at (+0.25,+1.25) {\textcolor{blue}{$\uparrow$}};%cell 12
\node at (+0.75,+1.25) {\textcolor{blue}{$\uparrow$}};%cell 13
\node at (+1.25,+1.25) {\textcolor{blue}{$\uparrow$}};%cell 14
\node at (-1.75,+0.75) {\textcolor{blue}{$\uparrow$}};%cell 15
\node at (-1.25,+0.75) {\textcolor{blue}{$\leftarrow$}};%cell 16
\node at (-0.75,+0.75) {\textcolor{blue}{$\uparrow$}};%cell 17
\node at (-0.25,+0.75) {\textcolor{blue}{$\uparrow$}};% cell 18
\node at (+0.25,+0.75) {\textcolor{blue}{$\rightarrow$}};%cell 19
\node at (+0.75,+0.75) {\textcolor{blue}{$\uparrow$}};%cell 20
\node at (+1.25,+0.75) {\textcolor{blue}{$\uparrow$}};%cell 21
\node at (-1.75,+0.25) {\textcolor{blue}{$\uparrow$}};%cell 22
\node at (-1.25,+0.25) {\textcolor{blue}{$\uparrow$}};%cell 23
\node at (-0.75,+0.25) {\textcolor{blue}{$\uparrow$}};%cell 24
\node at (-0.25,+0.25) {\textcolor{blue}{$\uparrow$}};%cell 25
\node at (+0.25,+0.25) {\textcolor{blue}{$\uparrow$}};%cell 26
\node at (+0.75,+0.25) {\textcolor{blue}{$\uparrow$}};%cell 27
\node at (+1.25,+0.25) {\textcolor{blue}{$\uparrow$}};%cell 28
\node at (-1.75,-0.25) {$\uparrow$};%cell 29
\node at (-1.25,-0.25) {\textcolor{blue}{$\leftarrow$}};%cell 30
\node at (-0.75,-0.25) {$\uparrow$};%cell 31
\node at (-0.25,-0.25) {$\uparrow$};%cell 32
\node at (+0.25,-0.25) {$\uparrow$};%cell 33
\node at (+0.75,-0.25) {\textcolor{blue}{$\uparrow$}};%cell 34
\node at (+1.25,-0.25) {$\uparrow$};%cell 35
\node at (-1.75,-0.75) {$\uparrow$};%cell 36
\node at (-1.25,-0.75) {\textcolor{blue}{$\leftarrow$}};%cell 37
\node at (-0.80,-0.75) {$\uparrow$};%cell 38
\node at (-0.25,-0.75) {$\uparrow$};%cell 39
\node at (+0.25,-0.75) {$\uparrow$};%cell 40
\node at (+0.75,-0.75) {$\uparrow$};%cell 41
\node at (+1.25,-0.75) {\textcolor{blue}{$\uparrow$}};%cell 42
\node at (-1.75,-1.25) {$\uparrow$};%cell 43
\end{tikzpicture}
\caption{EVaR optimal policy with $\alpha=0.6$}
    \label{fig:optimal-policy-alpha0.6}
\end{minipage}
\end{figure}

Before computing the optimal EVaR policy, we set $\beta_c = 1^{-10}$ in \cref{alg:z-bounds} to compute $c$ and $d$ in \cref{lemma:estimating-z-bounds}. $c$ and $d$ are used to estimate the bounds of $\tilde{z}_i(\beta)$ in ERM-TRC Q-learning algorithm. \cref{fig:optimal-policy-alpha0.2} shows the optimal policy for EVaR risk level $\alpha = 0.2$ computed by \cref{alg:evar-algorithm}. Each cell represents a state, and the arrow direction indicates the optimal action to take in that state. Note that the optimal actions for states $c$ and $g$ are the same and are omitted in \cref{fig:optimal-policy-alpha0.2} and \cref{fig:optimal-policy-alpha0.6}. Since the optimal policy is stationary, it can be interpreted and analyzed visually. The agent moves right to the last column of the grid world and then moves down to reach the goal. When the agent is close to the cliff region, it moves up to avoid risk. \cref{fig:optimal-policy-alpha0.6} shows the optimal policy for EVaR risk level $\alpha = 0.6$ computed by \cref{alg:evar-algorithm}. Different optimal actions are highlighted in blue. As we can see, for different risk levels $\alpha$, when the agent is near the cliff region, it exhibits consistent behaviors to avoid falling off the cliff. For the remaining states, the agent behaves differently.

\begin{figure}
\begin{minipage}{0.45\textwidth}
\centering
\includegraphics[width=0.95\linewidth]{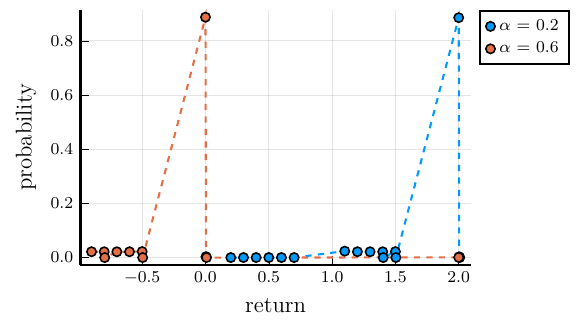}
\caption{Comparison of return distributions of two optimal EVaR policies.}
\label{fig:return-probability}
\end{minipage}%
\hspace{0.5cm}
\begin{minipage}{0.45\textwidth}
\centering
\includegraphics[width=0.85\linewidth]{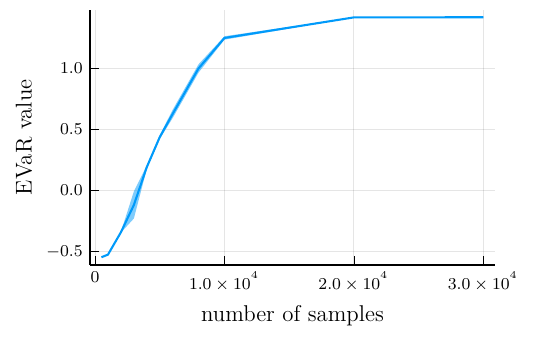}
\caption{Mean and standard deviation of EVaR values with $\alpha = 0.3$ on CW domain}
\label{fig:mean-std}
\end{minipage}%    
\end{figure}

Second, to understand the impact of risk aversion on the structure of returns, we simulate the two optimal EVaR policies over $48,000$ episodes and display the distribution of returns in \cref{fig:return-probability}. The $x$-axis represents the possible return the agent can get for each episode, and the $y$-axis represents the percentage of getting a certain amount of return over $48,000$ episodes. The start state is uniformly distributed over all states except the goal. For each episode, the agent starts with a non-sink state, takes $20,000$ steps, and collects all rewards during the path. When $\alpha = 0.6$, the return is in $(-1,2]$, and its mean value is $-0.074$, and its standard deviation is $0.228$. When $\alpha = 0.2$, the return is in $(0,2]$, its mean value is $1.92$, and its standard deviation is $0.228$. Overall, \cref{fig:return-probability} shows that for the lower value of $\alpha$, the agent has a higher probability of avoiding falling off the cliff.

Third, to assess the stability of our EVaR-TRC Q-learning algorithm, we use six random seeds to generate samples, compute the optimal policies, and calculate the EVaR values on the CW domain. In \cref{fig:mean-std}, the risk level $\alpha $ is $ 0.3$, and the $y$ axis represents the mean value and standard deviation of the EVaR values calculated in samples generated by six random seeds. \cref{fig:mean-std} shows that our Q-learning algorithm converges to similar solutions across different numbers of samples. The standard deviation is $0.021$. Note that our Q-learning algorithm could converge to a different optimal EVaR policy depending on the learning rate and the number of samples.

\begin{figure}
        \begin{minipage}{0.45\textwidth}
     \centering
\includegraphics[width=0.83\linewidth]{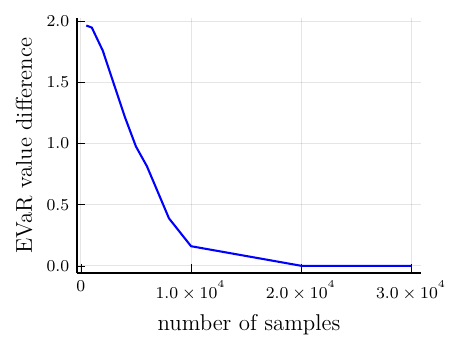}
 \caption{EVaR value converges on CW }
        \label{fig:evar-diff-cw}
    \end{minipage}%  
     \hspace{0.5cm}
     \begin{minipage}{0.45\textwidth}
     \centering
\includegraphics[width=0.9\linewidth]{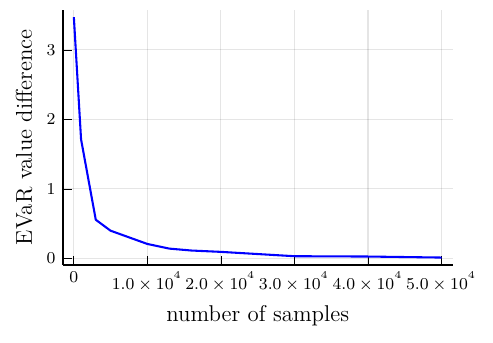}
 \caption{EVaR value converges on GR }
        \label{fig:evar-diff-gr}
    \end{minipage}%
  
\end{figure}

Finally, we evaluate the convergence of \cref{alg:evar-algorithm}, which approximates the EVaR value calculated from the linear programming(LP)~\cite{su2025risk} on CW and GR domains. The CW domain has one absorbing terminal state, and the GR domain has two absorbing terminal states. The $x$ axis represents the number of samples, and the $y$ axis represents the difference in EVaR values computed by LP and Q-learning algorithms. In \cref{fig:evar-diff-cw}, the risk level $\alpha $ is $ 0.4$, and the EVaR value difference decreases to zero when the number of samples is around 20,000 on the CW domain. In \cref{fig:evar-diff-gr}, the risk level $\alpha $ is $ 0.3$, and the EVaR value difference decreases to zero when the number of samples is around 30,000 on the GR domain. We illustrate the convergence result for some $\alpha$ values, but the conclusion applies to any $\alpha$ value. Overall, our Q-learning algorithm converges to the optimal value function.

\section{Conclusion and Limitations}

In this chapter, we proposed two new risk-averse model-free algorithms for risk-averse reinforcement learning objectives. We also proved the convergence of the proposed algorithm under mild assumptions. To the best of our knowledge, these are the first risk-averse model-free algorithms for the TRC criterion. Studying the TRC criterion in risk-averse RL is essential because it exhibits dramatically different properties in risk-averse objectives compared to the discounted criterion. 

%Although we focus on the theoretical foundations of the Q-learning algorithms,
An important limitation of this work is that our algorithms focus on the tabular setting and do not analyze the impact of value function approximation. However, the work lays out solid foundations for building approximately scalable reinforcement learning algorithms. In particular, a strength of Q-learning is that it can be coupled with general value function approximation schemes, such as deep neural networks. Although extending convergence guarantees to such approximate settings is challenging, our convergence results show that our Q-learning algorithm is a sound foundation for scalable, practical algorithms. Future work should include a rigorous numerical evaluation of the deep learning implementation of the Q-learning algorithms. 

Our theoretical analysis has two main limitations that may preclude its use in some practical settings. First, we must choose the limits $z_{\min}, z_{\max}$. If these limits are set too small, the algorithms may fail to compute a good solution, and if they are too large, the algorithms may be excessively slow to detect the divergence of the TRC criterion for large values of $\beta$. The second limitation is the need to select the parameter $\beta_0$ that guarantees the existence of a $\delta$-optimal EVaR solution. A natural choice for $\beta_0$ was proposed in \cite{su2025risk}, but it requires several runs of Q-learning just to establish an appropriate $\beta_0$. Future work needs to develop a better methodology for establishing these parameters. However, it is unclear whether a proper model-free solution is possible without making additional assumptions about the structure of the MDP. 
\chapter{Conclusion}
\label{chp:conclusion}

% \textcolor{red}{This paragraph should take a broader view of justifying the choice of the research questions. Say why these questions are important, that prior work has not answered them, and how the results are significant in their possible future impacts.} 

Researchers are actively working and have achieved some success in mitigating the effects of uncertainty and risk in sequential decision-making applications, such as self-driving cars, robotic surgery, healthcare, and finance. However, designing algorithms that are deployed to effectively mitigate the effects of uncertainty and risk in critical and high-stakes applications is still a challenging task. In this dissertation, we propose several efficient algorithms to mitigate the effects of uncertainty and risk in sequential decision-making processes. The optimal polices computed by the proposed algorithms are easy to interpret and deploy in real-world scenarios.

First, we proposed the CADP algorithm to guarantee that it never performs worse than previous dynamic programming algorithms, such as WSU. WSU has no guarantee of the optimality of its computed policy. CADP takes the coordinate ascent perspective to adjust model weights iteratively to guarantee monotone policy improvements to a local maximum. Our numerical results indicate that CADP substantially outperforms existing methods on several benchmark problems. Note that the objective is to find a Markov policy that maximizes the discounted
return averaged over an MMDP. The discounted criterion is a very common choice in MDPs, especially for infinite-horizon problems. Also, the discount factor ensures that the Bellman equations satisfy the contraction property (used for dynamic programming), and the value functions are always bounded. The limitation of this work is that the state space and action space are finite. Future work should adapt this approach to deep reinforcement learning and use neural networks to approximate weights.

Second, we focus on the total reward criterion under ERM and EVaR risk measures and propose exponential value iteration, policy iteration, and linear programming algorithms to compute the optimal stationary policies for the risk-averse TRC objectives. Unlike the discounted criterion, the Bellman operator in TRC may not be a contraction, and the value functions can be unbounded.
We establish the sufficient and necessary conditions for the Bellman operator to be a contraction. The experiment is based on the tabular Gambler domain. Our results indicate that the total reward criterion may be preferable to the discounted criterion in a broad range of risk-averse reinforcement learning domains.  Future work will evaluate the proposed algorithm on various benchmarks and practical problems. Future directions include extensions to infinite-state TRC problems, risk-averse MDPs with average
rewards, and partial-state observations.

Third, to address the unknown model of the environment, such as critical and high-stakes applications in practice, we proposed Q-learning algorithms to compute the optimal stationary policy for total-reward ERM and EVaR objectives. Because the transition dynamic is unknown, the contraction conditions of the ERM Bellman operator do not apply to this case.  Instead, we 
use the monotonicity of the ERM Bellman operator to prove the convergence of risk-averse Q-learning algorithms. Our numerical results on the Gabmber ruin and cliff walking domains demonstrate quick and reliable convergence
of the proposed Q-learning algorithm to the optimal risk-averse value function computed by linear programming.  This work lays out solid foundations for building scalable approximation reinforcement learning algorithms. In particular, a strength of
Q-learning is that it can be coupled with general value function approximation schemes, such as deep neural networks.  Future work should include rigorous numerical evaluation of deep
learning implementation of the Q-learning algorithms. Additionally, future work can investigate other learning rate methods and regularization techniques to mitigate the impact of the learning rate and the number of samples in Q-learning.

Overall, we investigate the discounted criterion under an expectation objective for an MMDP and the total reward criterion under risk-averse objectives for an MDP. We develop several efficient model-based and model-free algorithms to compute the optimal policies for these objectives. We believe that these approaches open up many opportunities to gain a deeper understanding of the fundamental issues involved in a broad range of risk-averse discounted-reward and total-reward reinforcement learning domains.

% bibliography
\singlespacing
\bibliographystyle{unsrt}

\bibliography{sources}

% appendices
\doublespacing
\titleformat{\chapter}[display]{\normalfont\bfseries}{\centering APPENDIX \thechapter}{0pt}{\centering}
\appendix

\chapter{}

\section{Standard Convergence Results}
Our convergence analysis of Q-learning algorithms is guided by the framework~\cite[section 4]{bertsekas1996neuro}. We summarize the framework in this section. Consider the following iteration for some random sequence $\tilde{r}_i : \Omega \rightarrow \mathbb{R}^{\mathcal{N}}$ where $\mathcal{N} = \{1,2,\cdots, n\}$, then Q-learning is defined as 
\begin{equation}
\label{eq:q-learning-framework}
    \begin{aligned}
        \tilde{r}_{i+1}(b) & = (1-\tilde{\theta}_i(b)) \cdot \tilde{r}_i(b) + \tilde{\theta}_i(b) \cdot ((H\tilde{r}_i)(b) + \tilde{\phi}_i(b)), \quad i=0,1,\cdots  \\
        & = \tilde{r}_i(b) + \tilde{\theta}_i(b) \cdot ((H\tilde{r}_i)(b) + \tilde{\phi}_i(b) - \tilde{r}_i(b)), \quad i =0,1,\cdots
    \end{aligned}
\end{equation}
for all $b \in \mathcal{N}$, where $H: \mathbb{R}^{\mathcal{N}} \rightarrow \mathbb{R}^{\mathcal{N}}$is some possibly non-linear operator, $\tilde{\theta}_i: \Omega \rightarrow \mathbb{R}_{++}$ is a step size, and $\tilde{\phi}_i: \Omega \rightarrow \mathbb{R}^{\mathcal{N}}$ is some random noise sequence. The random history $\mathcal{F}_i$ at iteration $i =1, \cdots$ is denoted by
\[
\mathcal{F}_i = (\tilde{r}_0, \cdots, \tilde{r}_i, \tilde{\phi}_0, \cdots, \tilde{\phi}_{i-1}, \tilde{\theta}_0, \cdots, \tilde{\theta}_i).
\]

The following assumptions will be needed to analyze and prove the convergence of our algorithms.

\begin{assumption}{[assumption 4.3 in \cite{bertsekas1996neuro}]}
\label{assump:step-size}

\begin{enumerate}
\item[(a)]   For every $i$ and $b$, we have 
  \[
  \E[\tilde{\phi}(b) | \mathcal{F}_i] = 0
  \]
\item[(b)]There exists a norm $\|\cdot\|$ on $\mathbb{R}^{\mathcal{N}}$ and constants $A$ and $B$ such that
  \[
  \E[\tilde{\phi}_i(b)^2 | \mathcal{F}_i] \le A + B \|\tilde{r}_i\|^2
  \]
  \end{enumerate}
\end{assumption}

\begin{assumption}{[assumption 4.4 in \cite{bertsekas1996neuro}]}
\label{assump:monotonicity}

\begin{enumerate}
\item[(a)] The mapping $H$ is monotone: that is, if $r \le \bar{r}$, then $Hr \le H\bar{r}$. 
\item[(b)] There exists a unique vector $r\opt$ satisfying $Hr\opt = r\opt$.
\item[(c)] If $e \in \mathbb{R}^n$ is the vector with all components equal to 1, and if $\eta$ is a positive scalar, then 
    \[
    Hr -\eta \cdot e \le H(r-\eta \cdot e) \le H(r+\eta \cdot e) \le Hr + \eta \cdot e
    \]
\end{enumerate}
\end{assumption}

Note that \cref{assump:monotonicity} leads to a convergence result somewhat weaker than the results for weighted maximum norm pseudo-contraction. Then we need a separate boundedness condition~\cite[proposition 4.6]{bertsekas1996neuro}, which is restated here as \cref{prop:boundedness-condition}.

\begin{proposition}[proposition 4.6 in \cite{bertsekas1996neuro}]
    \label{prop:boundedness-condition}
    
    Let $r_t$ be the sequence generated by the iteration shown in \cref{eq:q-learning-framework}. We assume that the $b$th component $\tilde{r}(b)$ of $\tilde{r}$ is updated according to \cref{eq:q-learning-framework}, with the understanding that $\tilde{\theta}_i(b) = 0$ if $\tilde{r}(b)$ is not updated at iteration $i$. $\tilde{\phi}_i(b)$ is a random noise term. Then we assume the following:
  \begin{enumerate}
\item[(a)]The step sizes $\tilde{\theta}_i(b)$ are nonnegative and satisfy
       \[
     \sum_{i=0}^{\infty} \tilde{\theta}_i(b)= \infty , \quad \sum_{i=0}^{\infty}\tilde{\theta}_i^2(b)< \infty,
    \]
\item[(b)]The noise terms $\tilde{\phi}_i(b)$ satisfies \cref{assump:step-size},
\item[(c)]The mapping $H$ satisfies \cref{assump:monotonicity}.
    \end{enumerate}

    If the sequence $\tilde{r}_i$ is bounded with probability 1, then  $\tilde{r}_i$ converges to $r\opt$ with probability 1.
\end{proposition}

\section{Proof of Theorem \ref{theorem:erm-q-converge}}
\label{proof:theorem-erm-q-converge}
To prove \cref{theorem:erm-q-converge}, we need to define some operators and analyze their properties, and prove the properties of random noise.

\paragraph{Operator Definitions}
Let us define some operators $G$ and $H$ for the convergence analysis of \cref{alg:erm-Q-learning-algorithm} and \cref{alg:evar-algorithm}. Let $b = (s,a,\beta), s \in \mathcal{S}, a \in \mathcal{A}$, $\beta \in \mathcal{B}$, and $\xi >0$. $(Gq)(b)$ can be interpreted as the average gradient steps. $(G_{s'}q)(b)$ can be interpreted as the individual gradient step for a specific next state $s'$.

\begin{equation}
\label{eq:gq}
    \begin{aligned}
    (Gq)(b) & := \frac{\partial}{\partial y} \E^{a,s}\Bigl[\ell_{\beta} \Bigl(r(s,a,\tilde{s}_1)+\max_{a'\in\mathcal{A}} q(\tilde{s}_1,a',\beta) - y\Bigr)\Bigr]\mid_{y = q(s,a,\beta)} \\
      & =  \E^{a,s}\Bigl[\partial \ell_{\beta} \Bigl(r(s,a,\tilde{s}_1)+\max_{a'\in\mathcal{A}} q(\tilde{s}_1,a',\beta) - q(s,a,\beta)\Bigr)\Bigr]
    \end{aligned}
\end{equation}
$\tilde{s}_1$ is a random variable representing the next state. When $\tilde{s}_1$ is used in an expectation with a superscript, such as $\E^{a,s}$, then it does not represent a sample $\tilde{s}_i$ with $i=1$. Still, instead it represents the transition from $\tilde{s}_0$ to $\tilde{s}_1$ distributed as $p(s, a,\cdot)$.

\begin{equation}
\label{eq:gsq}
\begin{aligned}
      (G_{s'}q)(b) & := \frac{\partial}{\partial y} \E\Bigl[\ell_{\beta} \Bigl(r(s,a,s')+\max_{a'\in\mathcal{A}} q(s',a',\beta) - y\Bigr)\Bigr]\mid_{y = q(s,a,\beta)} \\
      & =  \E\Bigl[\partial \ell_{\beta} \Bigl(r(s,a,s')+\max_{a'\in\mathcal{A}} q(s',a',\beta) - q(s,a,\beta)\Bigr)\Bigr]
\end{aligned}
\end{equation}

 $(Hq)(b)$ can be interpreted as the $q$ update for a random variable representing the next state. $(H_{s'}q)(b)$ can be interpreted as the $q$ update a specific next state $s'$. 
 
\begin{equation}
\label{eq:hq}
 (Hq)(b) := q(b)-\xi \cdot (Gq)(b) ,
   \qquad (H_{s'}q)(b):= q(b) - \xi \cdot (G_{s'}q)(b)
\end{equation}

Consider a random sequence of inputs $(\tilde{s}_i,\tilde{a}_i)_{i=0}^{\infty}$ in \cref{alg:erm-Q-learning-algorithm}. We can define a real-valued random variable $\tilde{\phi}(s, a)$ for $s \in \mathcal{S}, a \in \mathcal{A}, \beta > 0$ in~\eqref{eq:noise}.
\begin{equation}
    \label{eq:noise}
 \tilde{\phi}_i(s,a,\beta) = \begin{cases}
  (H_{\tilde{s}_i}\tilde{q}_i)(s,a)-(H\tilde{q}_i)(s,a) , & \text{if } (\tilde{s}_i,\tilde{a}_i) = (s,a)\\
    0, & \text{otherwise } \\
  \end{cases}
\end{equation}
$\tilde{\phi}_i$ can be interpreted as the random noise. 

\begin{equation}
    \label{eq:step-size-conversion}
 \tilde{\theta}_i(s,a,\beta) = \begin{cases}
  \tilde{\eta}_i / \xi, & \text{if } (\tilde{s}_i,\tilde{a}_i) = (s,a)\\
    0, & \text{otherwise } \\
  \end{cases}
\end{equation}

We denote by $\mathcal{F}_i$ the history of the algorithm until step $i$, which can be defined as 
\begin{equation}
\label{eq:history}
    \mathcal{F}_i = \{\tilde{q}_0,  \cdots, \tilde{q}_i, {\tilde{\phi}}_0, \cdots, {\tilde{\phi}}_{i-1},\tilde{\theta}_0, \cdots, \tilde{\theta}_i\}
\end{equation}

\begin{lemma}
\label{lemma:sequence-satisfy-hg}
    The random sequences of iterations followed by \cref{alg:erm-Q-learning-algorithm} satisfies
    \[
    \tilde{q}_{i+1}(s,a,\beta) = \tilde{q}_i(s,a,\beta) + \tilde{\theta}_i(s,a,\beta) \cdot (H\tilde{q}_i + \tilde{\phi}_i -\tilde{q}_i)(s,a,\beta), \forall s \in \mathcal{S}, a \in \mathcal{A}, \beta \in \mathcal{B}, i \in \mathbb{N}, a.s.,
    \]
    where the terms are defined in \cref{eq:gq,eq:gsq,eq:hq,eq:noise,eq:step-size-conversion}.
\end{lemma}

\begin{proof}
We prove this claim by induction on $i$. The base case holds immediately from the definition. To prove the inductive case, suppose that $i \in \mathbb{N}$ and we prove the result in the following two cases.

Case 1: suppose that $\tilde{b} = (\tilde{s},\tilde{a},\beta) = (s,a,\beta) = b$, then by algebraic manipulation
\begin{align*}
    \tilde{q}_{i+1}(b) &= \tilde{q}_i(b) + \tilde{\theta}_i(b)((H\tilde{q}_i) (b)+\tilde{\phi}_i(b) -\tilde{q}_i(b)) \\
    & = \tilde{q}_i(b) + \tilde{\theta}_i(b)((H\tilde{q}_i) (b)+ (H_{\tilde{s}'_i}\tilde{q}_i)(b) -(H\tilde{q}_i)(b)
    -\tilde{q}_i(b)) \\
    & = \tilde{q}_i(b) + \tilde{\theta}_i(b)((H_{\tilde{s}'_i}\tilde{q}_i)(b) -\tilde{q}_i(b)) \\
      & = \tilde{q}_i(b) + \tilde{\theta}_i(b)( (\tilde{q}_i -\xi G_{\tilde{s}'_i}\tilde{q}_i)(b) -\tilde{q}_i(b)) \\
       & = \tilde{q}_i(b) -\tilde{\eta}_i( (G_{\tilde{s}'_i}\tilde{q}_i)(b) ) \\
       & = \tilde{q}_i(b) -\tilde{\eta}_i( \partial \ell_{\beta} (r(s,a,\tilde{s}'_i)+\max_{a'\in\mathcal{A}} \tilde{q}_i(\tilde{s}'_i,a',\beta) - \tilde{q}_i(b)) ) 
\end{align*}
Case 2: suppose that $\tilde{b} = (\tilde{s},\tilde{a},\beta) \neq (s,a,\beta) = b$, then by algebraic manipulation, the algorithm does not change the $q$-function:
\begin{align*}
    \tilde{q}_{i+1}(b) &= \tilde{q}_i(b) + \tilde{\theta}_i(b)((H\tilde{q}_i) (b)+\tilde{\phi}_i(b) -\tilde{q}_i(b)) \\
        &= \tilde{q}_i(b) + 0 \cdot ((H\tilde{q}_i) (b)+\tilde{\phi}_i(b) -\tilde{q}_i(b)) \\
         &= \tilde{q}_i(b) 
\end{align*}   
\end{proof}

\paragraph{Monotonicity of Operator $H$}
Let us restate~\cite[C.13]{hau2024q} as \cref{lemma:gradient-optimal-y}, which is useful to prove that the operator $H$ satisfies monotonicity.
\begin{lemma}[C.13 in \cite{hau2024q}]
\label{lemma:gradient-optimal-y}
Suppose that $f: \mathbb{R} \rightarrow \mathbb{R}$ is a differentiable $\mu-$strongly convex function with an $ L-$Lipschitz continuous gradient. Consider $x_i \in \mathbb{R}$ and a gradient update for any step size $\xi \in (0,1/L]:$ 
\[
x_{i+1} := x_i - \xi \cdot f'(x_i)
\]
   Then $\exists l \in [1/L,1/\mu]$ such that $\xi / l \in (0,1]$ and 
   \[
   x_{i+1} = (1-\xi/l) \cdot x_i + \xi/l \cdot x\opt
   \]
   where $x\opt = \argmin_{x \in \mathbb{R}} f(x)$ (\text{unique from strong convexity})
\end{lemma}

\begin{lemma}
    \label{lemma:H-operator-monotonicity}
    Let $b = (s, a, \beta), s \in \mathcal{S}, a \in \mathcal{A}, \beta \in \mathcal{B}$, $\bm{1} \in \mathbb{R}^n$ is the vector with all components equal to 1, and $g >0$, then the $H$ operator defined in~\eqref{eq:hq} satisfies the monotonicity property such that
\begin{align}
\tag{a} 
x(b) \le y(b) & \Rightarrow (H x)(b) \le (H y)(b)\\
\tag{b} 
(Hq\opt)(b) & = q\opt(b) \\
\tag{c} 
(Hq)(b) -\bm{1} \cdot g \le H(q(b) -\bm{1} \cdot g) & \le H(q(b) + \bm{1} \cdot g) \le (Hq)(b) + \bm{1} \cdot g.
\end{align}
\end{lemma}

\begin{proof}

First, we prove part $(a)$: fix $\beta > 0$ and some $b=(s,a,\beta), s \in \mathcal{S}, a \in \mathcal{A}$. Fix $q$ and define 
\begin{equation}
\label{eq:f(y)}
    f(y) = \E^{s,a}\bigl[ \ell_{\beta}(r(s,a,\tilde{s}_1) + \max_{a' \in \mathcal{A}} q(\tilde{s}_1,a',\beta)-y)   \bigr]
\end{equation}
The function $f$ is strongly convex with Lipschitz gradient with parameters $\ell$ and $L$ based on \cref{lemma:gradient-lipschitz-bound-z}. Let $y\opt = \arg\min_{y \in \mathbb{R}}f(y)$ and $\exists l \in [1/L, 1/\ell $]
such that
\begin{equation}
\label{eq:hqb-}
    \begin{aligned}
            (Hq)(b) &= (q-\xi Gq)(b) \\
            & \stackrel{\text{(a)}}{=} q(b) -\xi f'(q(b)) \\
           & \stackrel{\text{(b)}}{=}  (1- \xi /l)q(b) + \xi/l \cdot y\opt \\
            & \stackrel{\text{(c)}}{=} (1- \xi /l)q(b) + \xi/l \cdot (\hat{B}q)(b) \\  
    \end{aligned}
\end{equation}
Step (a) follows the replacement with~\eqref{eq:gq} and~\eqref{eq:f(y)}. Step(b) follows the \cref{lemma:gradient-optimal-y}. Step(c) follows the fact that $\hat{B}$ is the ERM Bellman operator defined in~\eqref{eq:bellman-operator} and $y\opt$ is the unique solution to~\eqref{eq:erm=argmin}.
 
Given $x(b) \le y(b)$, let us prove that $(Hx)(b) \le (Hy)(b)$. Since the ERM Bellman operator $\hat{B}$ is monotone, we have 
\[
x(b) - y(b) \le 0 \Rightarrow (\hat{B}x)(b) - (\hat{B}y)(b) \le 0
\]
Then we have
\begin{align*}
     (Hx)(b) - (Hy)(b) 
    &= (1- \xi /l)x(b) + \xi/l \cdot (\hat{B}x)(b)  -\bigl((1- \xi /l)y(b) + \xi/l \cdot (\hat{B}y)(b)) \\
    &= (1- \xi /l)(x(b) - y(b)) + \xi/l \cdot ((\hat{B}x)(b) - (\hat{B}y)(b)) \\
    &\le  0 
\end{align*}
Second, let us prove part $(b)$: $(Hq)(b)$ can be written as follows.
 \[(Hq)(b) = (1- \xi /l)q(b) + \xi/l \cdot (\hat{B}q)(b) \] 
From~\cite[theorem~3.3]{su2025risk}, we know that $q\opt(b)$ is a fixed point of $\hat{B}$. Then $q\opt(b)$ is also a fixed point of $H$. That is, $(Hq\opt)(b) = q\opt(b)$.

Third, let us prove part $(c)$, we omit $b$ in the $ (Hq)(b)$ and rewrite it as follows.
\[
 Hq =(1- \xi /l)q + \xi/l \cdot \hat{B}q 
\]
Given $e \in \mathbb{R}^n$ is the vector with all components equal to 1 and if $c$ is a positive scalar, we show that 
\[
Hq - c \cdot e = H(q-c \cdot e) \le H(q+c \cdot e) = Hq + c \cdot e
\]

    1) Let us prove $Hq - c\cdot e = H(q-c \cdot e)$
\begin{align*}
    Hq - c \cdot e 
   = & (1-\xi/l)\cdot q + (\xi/l ) \cdot \hat{B}q - c \cdot e \\
   = & (1-\xi/l)\cdot q -(1-\xi/l) \cdot c \cdot e + (\xi/l ) \cdot \hat{B}q - (\xi/l ) \cdot c \cdot e \\
   = &(1-\xi/l) (q- c \cdot e ) + (\xi/l ) (\hat{B}q -c \cdot e ) \\
   \stackrel{\text{(a)}}{=}& (1-\xi/l) (q- c \cdot e ) + (\xi/l ) (\hat{B}(q -c \cdot e) ) \\
   = & H(q- c \cdot e)
\end{align*}
Step $(a)$ follows from the law invariance property of ERM~\cite{hau2023entropic}.

2) Let us prove $H(q-c \cdot e) \le H(q+c \cdot e)$

    Because $q-c \cdot e \le q+c \cdot e$ and the part $(a)$ of the operator $H$, we have
    \[
      H(q-c \cdot e) \le H(q+c \cdot e)
    \]
    
3) Let us prove $H(q+c \cdot e) = H(q) +c \cdot e$
\begin{align*}
    & H(q + c \cdot e) \\
    = &(1- \xi/l)(q + c \cdot e) + (\xi/l) \hat{B}(q+ c \cdot e) \\
    = & (1- \xi/l)(q + c \cdot e) + (\xi/l) (\hat{B}(q)+ c \cdot e) \\
    \stackrel{\text{(a)}}{=}&  (1- \xi/l)(q + c \cdot e) + (\xi/l) \hat{B}(q)+ (\xi/l) \cdot c \cdot e \\
    = & (1- \xi/l)q + (1- \xi/l) \cdot (c \cdot e) + (\xi/l) \cdot c \cdot e + (\xi/l) \hat{B}(q)\\
     = & (1- \xi/l)q +  c \cdot e + (\xi/l) \hat{B}(q)\\
     = & H(q) + c \cdot e
\end{align*}
Step $(a)$ follows from the law invariance of ERM.
Then we have
\[
Hq - c \cdot e \le H(q-c \cdot e) \le H(q+c \cdot e) \le Hq + c \cdot e
\]
\end{proof}

\paragraph{Random Noise Analysis} We restate Lemma C.16 in~\cite{hau2024q} here as \cref{lemma:sample-history}, which is useful for proving properties of the random noise.
\begin{lemma}[Lemma C.16 in \cite{hau2024q}]
\label{lemma:sample-history}

   Under \cref{assump:transition-prop}:
   \[
   \mathbb{P}[\tilde{s}'_i = s' | \mathcal{G}_{i-1}, \tilde{b}_i, \tilde{\xi}_i, \mathcal{F}_i] = p(\tilde{s}_i, \tilde{a}_i, s'), a.s.,
   \]
   for each $s' \in \mathcal{S}$ and $i \in \mathbb{N}$.
\end{lemma}

\begin{lemma}
\label{lemma:noise-expectation}
    The noise $\tilde{\phi}_i$ defined in~\eqref{eq:noise} satisfies 
    \[
    \E[\tilde{\phi}_i(s,a,\beta)| \mathcal{F}_i] = 0, \forall s \in \mathcal{S}, a \in \mathcal{A}, \beta \in \mathcal{B}, i \in \mathbb{N}
    \]
   almost surely, where $\mathcal{F}_i$ is defined in~\eqref{eq:history}.
\end{lemma}
\begin{proof}
Let $b := (s,a,\beta)$, $\tilde{b}_i := (\tilde{s}_i,\tilde{a}_i,\beta)$ and $i \in \mathbb{N}$. We decompose the expectation using the law of total expectation to get 
\begin{equation}
\label{eq:noise-expectation}
  \E[ \tilde{\phi}_i(b) \mid \mathcal{F}_i] = \E[ \tilde{\phi}_i(b) \mid \mathcal{F}_i,\tilde{b}_i \neq b ] \cdot \mathbb{P}[\tilde{b}_i \neq b \mid \mathcal{F}_i] + \E[ \tilde{\phi}_i(b) \mid \mathcal{F}_i,\tilde{b}_i = b ] \cdot \mathbb{P}[\tilde{b}_i = b \mid \mathcal{F}_i]  
\end{equation}
 From the definition in~\eqref{eq:noise}, we have 
 \[
  \E[ \tilde{\phi}_i(b) \mid \mathcal{F}_i,\tilde{b}_i \neq b ] \cdot \mathbb{P}[\tilde{b}_i \neq b \mid \mathcal{F}_i] = 0
 \]
Then 
\begin{align*}
    \E[ \tilde{\phi}_i(b) \mid \mathcal{F}_i,\tilde{b}_i = b ]  
    &=\mathbb{E}[ (H_{\tilde{s}'} \tilde{q}_i)(b) - (H\tilde{q}_i)(b) \mid \mathcal{F}_i, \tilde{b}_i =b   ] \\
      &= \xi \cdot \mathbb{E}[-(G_{\tilde{s}'}\tilde{q}_i)(b) + (G\tilde{q}_i)(b)  \mid \mathcal{F}_i, \tilde{b}_i =b ] \\
      &= \xi\cdot \mathbb{E}[ \mathbb{E} [-(G_{\tilde{s}'}\tilde{q}_i)(b) \mid \mathcal{F}_i, \tilde{b}_i = b, \tilde{\eta}_i, \mathcal{G}_{i-1}] + (G\tilde{q}_i)(b) \mid  \mathcal{F}_i, \tilde{b}_i = b] \\
        &\stackrel{\text{(a)}}{=} \xi \cdot \mathbb{E}[ \mathbb{E}^{a,s} [-(G_{\tilde{s}_1}\tilde{q}_i)(b)] + (G\tilde{q}_i)(b) \mid  \mathcal{F}_i, \tilde{b}_i = b] \\
        &= \xi \cdot \mathbb{E}[ -(G\tilde{q}_i)(b) + (G\tilde{q}_i)(b) \mid  \mathcal{F}_i, \tilde{b}_i = b] \\
        &= 0
\end{align*}
When $\tilde{s}_1$ is used in an expectation with subscript, such as $\E^{a,s}$, then it does not represent a sample $\tilde{s}_i$ with $i=1$, but instead it represents the transition from $\tilde{s}_0 =s $ to $\tilde{s}_1$ distributed as $p(s,a,\cdot)$. Step (a) follows the fact the randomness of $(G_{\tilde{s}'_i}\tilde{q}_i)(b)$ only comes from $\tilde{s}'_i$ when conditioning on $\mathcal{F}_i, \tilde{b}_i =b, \tilde{\eta}_i$, and $\mathcal{G}_{i-1}$.

Then, we have 
\begin{align*}
     \E[ \tilde{\phi}_i(b) \mid \mathcal{F}_i] 
     &= \E[ \tilde{\phi}_i(b) \mid \mathcal{F}_i,\tilde{b}_i \neq b ] \cdot \mathbb{P}[\tilde{b}_i \neq b \mid \mathcal{F}_i] + \E[ \tilde{\phi}_i(b) \mid \mathcal{F}_i,\tilde{b}_i = b ] \cdot \mathbb{P}[\tilde{b}_i = b \mid \mathcal{F}_i] \\
     &= 0 + 0  \cdot \mathbb{P}[\tilde{b}_i = b \mid \mathcal{F}_i]\\
     &= 0
\end{align*}
\end{proof}

\begin{lemma}
    \label{lemma:noise-variance}
      The noise $\tilde{\phi}_i$ defined in~\eqref{eq:noise} satisfies 
    \[
   \E[(\tilde{\phi}_i(s,a,\beta))^2 | \mathcal{F}_i]  \le A + B \|\tilde{q}_i\|^2_{\infty}, \quad \forall s \in \mathcal{S}, a \in \mathcal{A}, \beta \in \mathcal{B}, i \in \mathbb{N}
    \]
    almost surely for some $A,B \in \mathbb{R}_{+}$,  $\mathcal{F}_i$ is defined in~\eqref{eq:history}.
\end{lemma}

\begin{proof}
    Let $b := (s,a,\beta)$, $\tilde{b}_i := (\tilde{s}_i,\tilde{a}_i,\beta)$ and $i \in \mathbb{N}$. We decompose the expectation using the law of total expectation to get 
\begin{align*}
  \mathbb{E}[\tilde{\phi}_i(b)^2 \mid \mathcal{F}_i]
 &= \mathbb{E}[\tilde{\phi}_i(b)^2 \mid \mathcal{F}_i, \tilde{b}_i \neq b] \cdot \mathbb{P}[\tilde{b}_i \neq b \mid \mathcal{F}_i] + \mathbb{E}[\tilde{\phi}_i(b)^2 \mid \mathcal{F}_i, \tilde{b}_i = b] \cdot \mathbb{P}[\tilde{b}_i = b \mid \mathcal{F}_i] \\
 &= \mathbb{E}[\tilde{\phi}_i(b)^2 \mid \mathcal{F}_i, \tilde{b}_i = b] \cdot \mathbb{P}[\tilde{b}_i = b \mid \mathcal{F}_i] 
 \end{align*}
This is because $\mathbb{E}[\tilde{\phi}_i(b)^2 \mid \mathcal{F}_i, \tilde{b}_i \neq b] = 0$. Let us evaluate
$\mathbb{E}[\tilde{\phi}_i(b)^2 \mid \mathcal{F}_i, \tilde{b}_i = b]$.

 \begin{align*}
 \mathbb{E}[\tilde{\phi}_i(b)^2 \mid \mathcal{F}_i, \tilde{b}_i = b] 
   &= \mathbb{E}\Bigl[ \Bigl((H_{\tilde{s}'_i} \tilde{q}_i)(b) - (H\tilde{q}_i)(b) \Bigr)^2\mid \mathcal{F}_i, \tilde{b}_i =b  \Bigr ] \\
     &= \mathbb{E}\Bigl[\Bigl(-\xi \cdot (G_{\tilde{s}'_i}\tilde{q}_i)(b) + \xi \cdot (G\tilde{q}_i)(b) \Bigr)^2 \mid \mathcal{F}_i, \tilde{b}_i =b  \Bigr] \\
       &= \xi^2 \cdot \mathbb{E}\Biggl[ \E \Bigl[\Bigl(-(G_{\tilde{s}'_i}\tilde{q}_i)(b) + (G\tilde{q}_i)(b) \Bigr)^2 \mid \mathcal{F}_i, \tilde{b}_i =b, \mathcal{G}_{i-1}\Bigr]
       \mid \mathcal{F}_i, \tilde{b}_i =b \Biggr] 
\end{align*}
Let us define $\tilde{\delta_i}(s',\beta)$ in~\eqref{eq:delta}.
\begin{equation}
\label{eq:delta}
    \tilde{\delta_i}(s',\beta) = r(s,a,s') + \max_{a' \in \mathcal{A}}\tilde{q}_i(s',a',\beta) - \tilde{q}_i(s,a,\beta)
\end{equation}
\begin{align*}
    & \xi^2 \cdot \mathbb{E}\Biggl[ \E \Bigl[\Bigl(-(G_{\tilde{s}'_i}\tilde{q}_i)(b) + (G\tilde{q}_i)(b) \Bigr)^2 \mid \mathcal{F}_i, \tilde{b}_i =b, \tilde{\eta}_i, \mathcal{G}_{i-1}\Bigr]
       \mid \mathcal{F}_i, \tilde{b}_i =b \Biggr] \\
 \stackrel{\text{(a)}}{=} & \xi^2 \cdot \mathbb{E}\Biggl[ \E^{a,s}\Bigl[\Bigl((G_{\tilde{s}_1}\tilde{q}_i)(b) - (G\tilde{q}_i)(b) \Bigr)^2 \Bigr]
       \mid \mathcal{F}_i, \tilde{b}_i =b \Biggr] \\
\stackrel{\text{(b)}}{=} & \xi^2 \cdot \mathbb{E}\biggl[ \E^{a,s}\Bigl[\Bigl(\E[\partial \ell_{\beta}(\tilde{\delta_i}(\tilde{s}_1,\beta)) \mid \tilde{s}_1] -\E^{a,s}[\partial \ell_{\beta}(\tilde{\delta_i}(\tilde{s}_1,\beta))] \Bigr)^2 \Bigr] \mid \mathcal{F}_i, \tilde{b}_i =b \biggr] \\
\stackrel{\text{(c)}}{=} & \xi^2 \cdot \mathbb{E}\biggl[ \Bigl(\E^{a,s}\Bigl[ \bigl(\E[\partial \ell_{\beta}(\tilde{\delta_i}(\tilde{s}_1,\beta)) \mid \tilde{s}_1] \bigr)^2 \Bigr] - \Bigl(\E^{a,s}[\partial \ell_{\beta}(\tilde{\delta_i}(\tilde{s}_1,\beta))] \Bigr)^2 \Bigr] \Bigr) \mid \mathcal{F}_i, \tilde{b}_i =b \biggr] \\
\le &\xi^2 \cdot \mathbb{E}\biggl[ \E^{a,s}\Bigl[ \bigl(\E[\partial \ell_{\beta}(\tilde{\delta_i}(\tilde{s}_1,\beta)) \mid \tilde{s}_1] \bigr)^2 \Bigr]  \mid \mathcal{F}_i, \tilde{b}_i =b \biggr] \\
\stackrel{\text{(d)}}{\le} & \xi^2 \cdot \mathbb{E}\biggl[ \max_{s' \in \mathcal{S}} \partial \ell_{\beta} (\tilde{\delta_i}(s',\beta))^2 \mid \mathcal{F}_i, \tilde{b}_i =b \biggr] \\
\stackrel{\text{(e)}}{\le} & \xi^2 \cdot \mathbb{E}\biggl[ \max_{s' \in \mathcal{S}} (\mid \partial \ell_{\beta} (  \tilde{\delta_i}(s',\beta))  - \partial \ell_{\beta} (  0) \mid )^2 \mid \mathcal{F}_i, \tilde{b}_i =b \biggr] \\
\stackrel{\text{(f)}}{\le} & \xi^2 \cdot \mathbb{E}\biggl[ \max_{s' \in \mathcal{S}} (\frac{\beta}{e^{\beta z_{\min}} } \cdot \mid \tilde{\delta_i}(s',\beta) \mid )^2 \mid \mathcal{F}_i, \tilde{b}_i =b \biggr] \\
\le & \xi^2 \cdot (\frac{\beta}{e^{\beta z_{\min}} })^2\cdot \mathbb{E}\biggl[ \max_{s' \in \mathcal{S}} ( \tilde{\delta_i}(s',\beta) )^2 \mid \mathcal{F}_i, \tilde{b}_i =b \biggr] \\
\stackrel{\text{(g)}}{\le} & \xi^2 \cdot (\frac{\beta}{e^{\beta z_{\min}} })^2 \cdot (2\cdot \|r\|_{\infty}^2 + 8 \cdot \|\tilde{q}_i\|^2_{\infty}) \\
\end{align*}
Step $(a)$ follows \cref{lemma:sample-history} given that the randomness of $-(G_{\tilde{s}'_i}\tilde{q}_i)(b) + (G\tilde{q}_i)(b) $ only comes from $\tilde{s}'_i$ when conditioning on $\mathcal{F}_i, \tilde{b}_i =b, \tilde{\eta}_i$ and $\mathcal{G}_{i-1}$. Step $(b)$ follows by substituting $(G_{\tilde{s}'_i}\tilde{q}_i)(b)$ with~\eqref{eq:gq}, substituting $(G\tilde{q}_i)(b)$ with~\eqref{eq:gsq}, and replacing by using $\tilde{\delta}_i$ defined in~\eqref{eq:delta}. The equality in step $(c)$ holds because for a random variable $\tilde{x} = \E[\partial \ell_{\beta}(\tilde{\delta_i}(\tilde{s}_1,\beta)) \mid \tilde{s}_1]$, the variance satisfies $\E[(\tilde{x} -\E[\tilde{x}])^2] = \E[\tilde{x}^2] - (\E[\tilde{x}])^2$. Step $(d)$ upper bounds the expectation by a maximum. Step $(e)$ uses $\partial \ell_{\beta}(0)=0 $  from the definition in~\eqref{eq:erm-loss-function}. Step $(f)$ uses \cref{lemma:gradient-lipschitz-bound-z} to bound the derivative difference as a function of the step size. Step $(g)$ derives the final upper bound since
\[
\|r\|_{\infty} = \max_{s,s' \in \mathcal{S}, a \in \mathcal{A}} |r(s,a,s')|, \quad \|\tilde{q}_i\|_{\infty} = \max_{s \in \mathcal{S}, a \in \mathcal{A}}|\tilde{q}_i(s,a,\beta)|
\]
 
\begin{equation}
\label{eq:td-bound}
\begin{aligned}
  \max_{s' \in \mathcal{S}} \tilde{\delta}_i(s', \beta)^2
    &\le  (\|r\|_{\infty} + 2 \|\tilde{q}_i\|_{\infty})^2\\
    &\le  (\|r\|_{\infty} + 2 \|\tilde{q}_i\|_{\infty})^2 +(\|r\|_{\infty} - 2 \|\tilde{q}_i\|_{\infty})^2\\
    &= 2\|r\|_{\infty}^2 + 8 \|\tilde{q}_i\|_{\infty}^2
\end{aligned}
\end{equation}
Then, we have
\begin{align*}
\mathbb{E}[\tilde{\phi}_i(b)^2 \mid \mathcal{F}_i] 
 &= \mathbb{E}[\tilde{\phi}_i(b)^2 \mid \mathcal{F}_i, \tilde{b}_i = b] \cdot \mathbb{P}[\tilde{b}_i = b \mid \mathcal{F}_i] \\
&\le \xi^2 \cdot (\frac{\beta}{e^{\beta z_{\min}} })^2 \cdot (2\cdot \|r\|_{\infty}^2 + 8 \cdot \|\tilde{q}_i\|^2_{\infty})
 \end{align*}

Therefore, $A = \xi^2 \cdot (\frac{\beta}{e^{\beta z_{\min}} })^2 \cdot (2\cdot \|r\|_{\infty}^2 )$ and $B=8\xi^2 \cdot (\frac{\beta}{e^{\beta z_{\min}} })^2 $.
\end{proof}

Now, we are ready to prove \cref{theorem:erm-q-converge}.
\begin{proof}
\label{proof:lemma-erm-q-boundedness} We verify that the sequence of our $q$-learning iterates satisfies the properties in \cref{prop:boundedness-condition}. The step size condition in \cref{theorem:erm-q-converge} guarantees that we satisfy property $(a)$ in \cref{prop:boundedness-condition}. \cref{lemma:noise-expectation} and \cref{lemma:noise-variance} show that we satisfy property (b) in \cref{prop:boundedness-condition}. \cref{lemma:H-operator-monotonicity} shows that we satisfy property $(c)$ in \cref{prop:boundedness-condition}. 
\end{proof}

% add more appendices here using \input{}

\end{document}